\documentclass[10pt,twocolumn,letterpaper]{article}

\usepackage[pagenumbers]{iccv} 

\definecolor{iccvblue}{rgb}{0.21,0.49,0.74}
\usepackage[pagebackref,breaklinks,colorlinks,allcolors=iccvblue]{hyperref}

\usepackage{url}

\usepackage{graphicx}
\usepackage{amsmath}
\usepackage{amssymb}
\usepackage{booktabs}
\usepackage{makecell}
\usepackage{multirow}
\usepackage{overpic}
\usepackage{comment}
\usepackage{float}
\usepackage[export]{adjustbox}
\usepackage{enumitem}
\usepackage{array}
\usepackage{colortbl}%
\usepackage{caption}
\usepackage{wrapfig}
\usepackage{subcaption}
\usepackage{tabularx}
\usepackage{natbib}
\usepackage{mathrsfs}

\definecolor{lightpink}{rgb}{1, 0.8, 0.8}
\definecolor{lightcyan}{rgb}{0.85, 1, 1}
\newcommand\sotaa{\textcolor{red}}
\newcommand\sotab{\textcolor{blue}}

\newcommand{\myrowcolour}{\rowcolor[gray]{0.925}}
\newcommand{\myrowcolourpink}{\rowcolor{pink}}
\newcommand\revise{\textcolor{black}}
\newcommand{\ourmethod}{\textit{Fractal-IR}}

\def\vs{\emph{vs.\ }}
\def\eg{\emph{e.g.,\ }}
\def\ie{\emph{i.e.,\ }}

\def\etc{\emph{etc.\ }}

\def\etal{\emph{et al.\ }}

\def\paperID{515} 
\def\confName{ICCV}
\def\confYear{2025}

\title{\ourmethod: A Unified Framework for Efficient and Scalable Image Restoration}

\author{
\textbf{Yawei Li}$^{1}$\quad
\textbf{Bin Ren}$^{2,3}$\quad
\textbf{Jingyun Liang}$^{1}$\quad
\textbf{Rakesh Ranjan}$^{4}$\quad
\textbf{Mengyuan Liu}$^{5}$\\
\textbf{Nicu Sebe}$^{3}$\quad
\textbf{Ming-Hsuan Yang}$^{6}$\quad 
\textbf{Luca Benini}$^{1}$ \\
$^1$ETH Z\"urich,
$^2$University of Pisa,
$^3$University of Trento,
$^4$Meta Reality Labs,\\
$^5$Peking University,
$^6$University of California, Merced
}

\begin{document}

\twocolumn[{%
\renewcommand\twocolumn[1][]{#1}%
\maketitle

\vspace{-12mm}
\begin{center}
    \includegraphics[width=\linewidth]{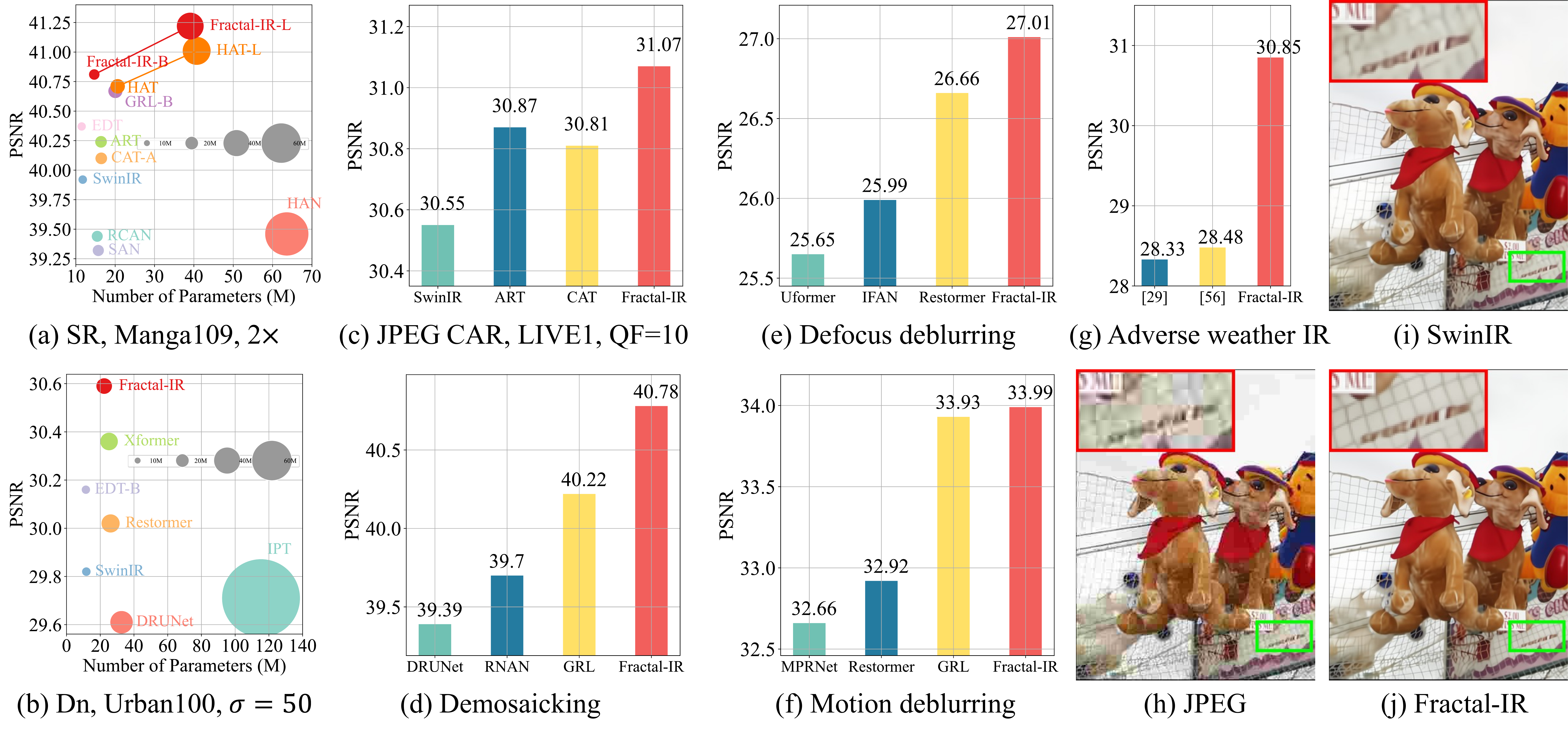}%
    \vspace{-2.3mm}
    \captionof{figure}{The proposed {\ourmethod} is notable for its efficiency and effectiveness (a)-(b), generalizability across seven image restoration tasks (a)-(g), and improvements in the visual quality of restored images (h)-(j).}
    \label{fig:teaser}
\end{center}%
}]

\begin{abstract}

\vspace*{-4mm}
While vision transformers achieve significant breakthroughs in various image restoration (IR) tasks, it is still challenging to efficiently scale them across multiple types of degradations and resolutions. In this paper, we propose {\ourmethod}, a fractal-based design that progressively refines degraded images by repeatedly expanding local information into broader regions. This fractal architecture naturally captures local details at early stages and seamlessly transitions toward global context in deeper fractal stages, removing the need for computationally heavy long-range self-attention mechanisms. Moveover, we observe the challenge in scaling up vision transformers for IR tasks. Through a series of analyses, we identify a holistic set of strategies to effectively guide model scaling. Extensive experimental results show that {\ourmethod} achieves state-of-the-art performance in seven common image restoration tasks, including super-resolution, denoising, JPEG artifact removal, IR in adverse weather conditions, motion deblurring, defocus deblurring, and demosaicking. For $2\times$ SR on Manga109, {\ourmethod} achieves a 0.21 dB PSNR gain. For grayscale image denoising on Urban100, {\ourmethod} surpasses the previous method by 0.2 dB for $\sigma=50$.
\vspace*{-4mm}
\end{abstract}

\section{Introduction}
\label{sec:introduction}

\begin{figure*}[!t]
    \centering
    \includegraphics[width=0.99\linewidth]{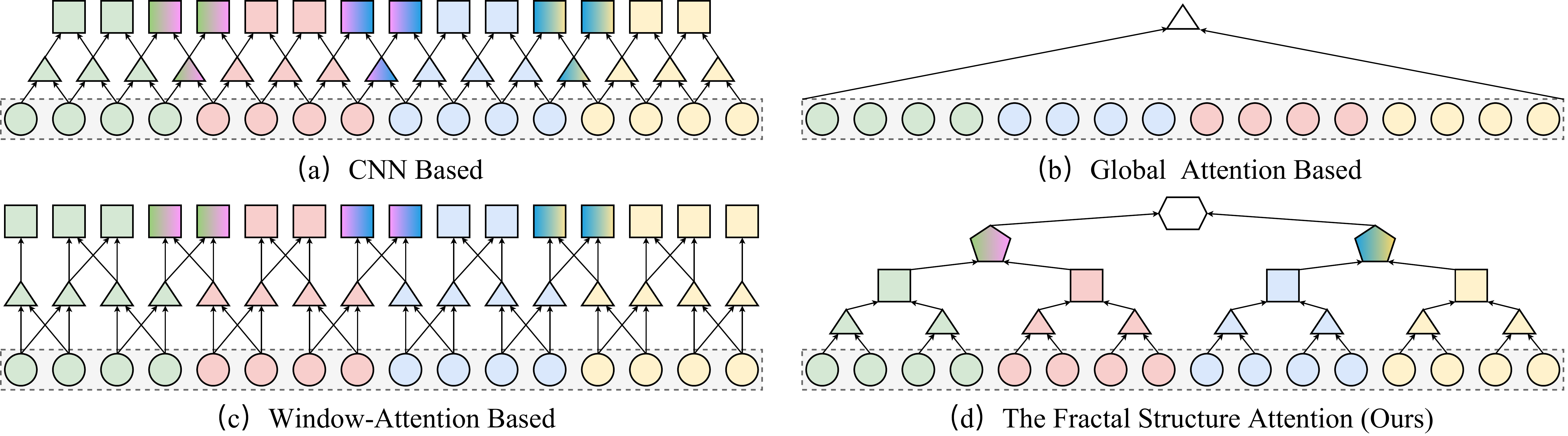}
    \vspace{-2mm}
    \caption{Illustration of information flow principles. \revise{The colors represent local information, with their blending indicating propagation beyond the local region.} (a) The CNN-based. (b) The global attention based. (c) Window attention based. (d) The proposed hierarchical information flow prototype.}
    \label{fig:motivation}
    \vspace{-4mm}
\end{figure*}

Image restoration (IR) aims to improve image quality by recovering high-quality visuals from observations degraded by noise, blur, and downsampling. 
To address these inherently ill-posed problems, numerous methods have been developed primarily for a single degradation, including convolutional neural networks (CNNs)~\citep{dong2014learning,lim2017enhanced}, vision transformers (ViTs)~\citep{chen2021pre,liang2021swinir}, and state space models (Mamba)~\citep{guo2024mambair}.  

Despite the advancements, a key challenge remains: \textit{how to design a model that can efficiently handle multiple IR tasks and scale up to accommodate various image resolutions and degradation complexities.}
Recent transformer-based solutions have focused on improving the efficiency of self-attention, often by partitioning images into smaller windows and refining local details. 
Yet, such design risk isolating spatial regions and may require additional operations (\eg window shifting)~\cite{liang2021swinir,conde2022swin2sr} or large receptive fields to capture global context~\cite{chen2023activating,li2023efficient}. 
As image resolutions grow and degradations vary, purely local or fully global mechanisms can become either insufficiently expressive or computationally prohibitive.

To address the efficiency issue, we propose a fractal-based framework shown in Fig.~\ref{fig:motivation}(d) for IR. 
In our approach, local details are initially processed in smaller fractal cells, then repeatedly expanded and merged, mirroring how fractals capture both micro-level patterns and macro-level structures~\cite{bassett2006adaptive,larsson2016fractalnet,li2025fractal}. 
Each fractal iteration fuses information from earlier stages, ensuring that local details and broad contexts reinforce one another without incurring the high computational costs of naive global self-attention. 

Despite the advantages of the fractal design, scaling the model beyond 50M parameters still remains challenging, as evidenced by the degraded performance of larger models~\cite{lim2017enhanced,chen2023activating}. Further investigation attributes this issue to several factors including parameter initialization, training strategy, and operator design. 
To address these challenges, we propose a holistic set of three strategies for effective model scaling, including replacing heavyweight convolutions with more efficient alternatives, implementing a warm-up phase during training, and using dot-product self-attention. These strategies enable the successful training of larger IR models.

The proposed {\ourmethod} framework has the advantages of efficiency, scalability, and generalization for IR tasks. 
1) By progressively expanding compact representations through fractal stages, {\ourmethod} captures broader structures and semantics without resorting to dense, full-image attention. This leads to notably lower memory usage and computational overhead compared to standard transformer-based approaches.
2) The fractal design inherently supports deeper or wider expansions, enabling the model to scale effectively while maintaining efficiency. With the proposed model scaling strategies, {\ourmethod} can be successfully scaled up to provide large capacity and capture richer patterns. 
3) Empirical results show that the proposed {\ourmethod} framework generalizes well across different IR tasks, consistently achieving strong performance across different types of degradation and restoration challenges.

Our main contributions are summarized as follows:
\begin{itemize}[leftmargin=*]
    \item[$\bullet$] We introduce a novel fractal-based design for image restoration, which facilitates progressive global information exchange and mitigates the curse of dimensionality.

    \item[$\bullet$] We examine the challenge of training convergence for model scaling-up in IR and propose mitigation strategies.

    \item[$\bullet$] Extensive experiments validate the effectiveness of the proposed method. {\ourmethod} consistently outperforms state-of-the-art IR methods for multiple tasks.
\end{itemize}

\section{Related Work}
\label{sec:related-work}
\noindent{\textbf{Image Restoration.}} The focus of IR is to recover high-quality images from their degraded counterparts. 
As a challenging problem, IR has captured substantial interest, leading to practical applications such as denoising~\cite{buades2005non,dabov2007image,gu2014weighted,zhang2017beyond}, deblurring~\cite{levin2009undertanding,shan2008high,pan2014deblurring}, super-resolution (SR)~\cite{farsiu2004fast,protter2008generalizing,jianchao2008image}, demosaicking~\cite{malvar2004high,zhang2005color}, JPEG compression artifacts removal~\cite{chen2016trainable}, \etc
The landscape of IR has shifted with the evolution of deep learning and the increased availability of computational resources. 
%
Many CNN models have been proposed~\citep{anwar2020densely,li2022blueprint,dong2014learning,zhang2017beyond} for different IR tasks.
However, despite their effectiveness, CNNs have been found to struggle in propagating long-range information within degraded input images. 
This challenge is attributed to the limited receptive field of CNNs, which constrains the overall performance of CNN-based methods~\citep{chen2022cross,zhang2022accurate,li2023efficient}. Thus, transformer-based models are proposed to solve this problem~\cite{chen2021pre,liang2021swinir,zamir2022restormer}. Yet, the computational complexity of transformers scales quadratically with the number of tokens. To solve this problem, recent methods utilize state-space models which have linear complexity growth~\cite{gu2023mamba,guo2024mambair}. 

\noindent{\textbf{Hierarchical Models.}} Modeling image hierarchies is essential for IR tasks. Traditional CNNs progressively propagate information to the global range with stacked convolutional layers. However, this has not been efficient. Thus, non-local operators were proposed~\cite{liu2018NLRN,liu2018non,mei2021NLSA,zhang2019residual} to enhance the global modeling capacity of CNNs. With the advent of transformers~\cite{vaswani2017attention,dosovitskiy2020image,chen2021pre}, image hierarchies are modeled via a combination of self-attention and convolution operations. 
Various methods have been proposed to achieve efficient and effective self-attention for hierarchical image modeling. 
SwinIR~\citep{liang2021swinir} conducts multi-head self-attention (MSA) window-wise. 
A shift operation is applied to enable information exchange between windows~\citep{liu2021swin}. 
Uformer~\citep{wang2022uformer} and Restormer~\cite{zamir2022restormer} proposes to propagate information hierarchically with a UNet structure but still with window self-attention. 
Other methods refine self-attention with much more exquisite efforts, including rectangle-window self-attention~\citep{li2021efficient}, sparse self-attention~\cite{huang2021shuffle}, and graph-attention~\citep{ren2024sharing}, spatial shuffle~\citep{huang2021shuffle}, and random spatial shuffle~\cite{xiao2023random}. 
More recently, fractal structures are utilized for image generation~\cite{li2025fractal}, which offers a significant improvement of generation efficiency.
In this paper, we propose a general and efficient IR solution which hierarchically propagates information within a fractal architecture.

\section{Methodology}
\label{sec:methodology}

\begin{figure*}[!t]
    \centering
    \begin{minipage}{0.2\textwidth}
        \centering
        \includegraphics[height=5.5cm]{images/fractal_info_flow.pdf}
        \subcaption{}
    \end{minipage}
    \hspace{6mm}  
    \begin{minipage}{0.15\textwidth}
        \centering
        \includegraphics[height=5.5cm]{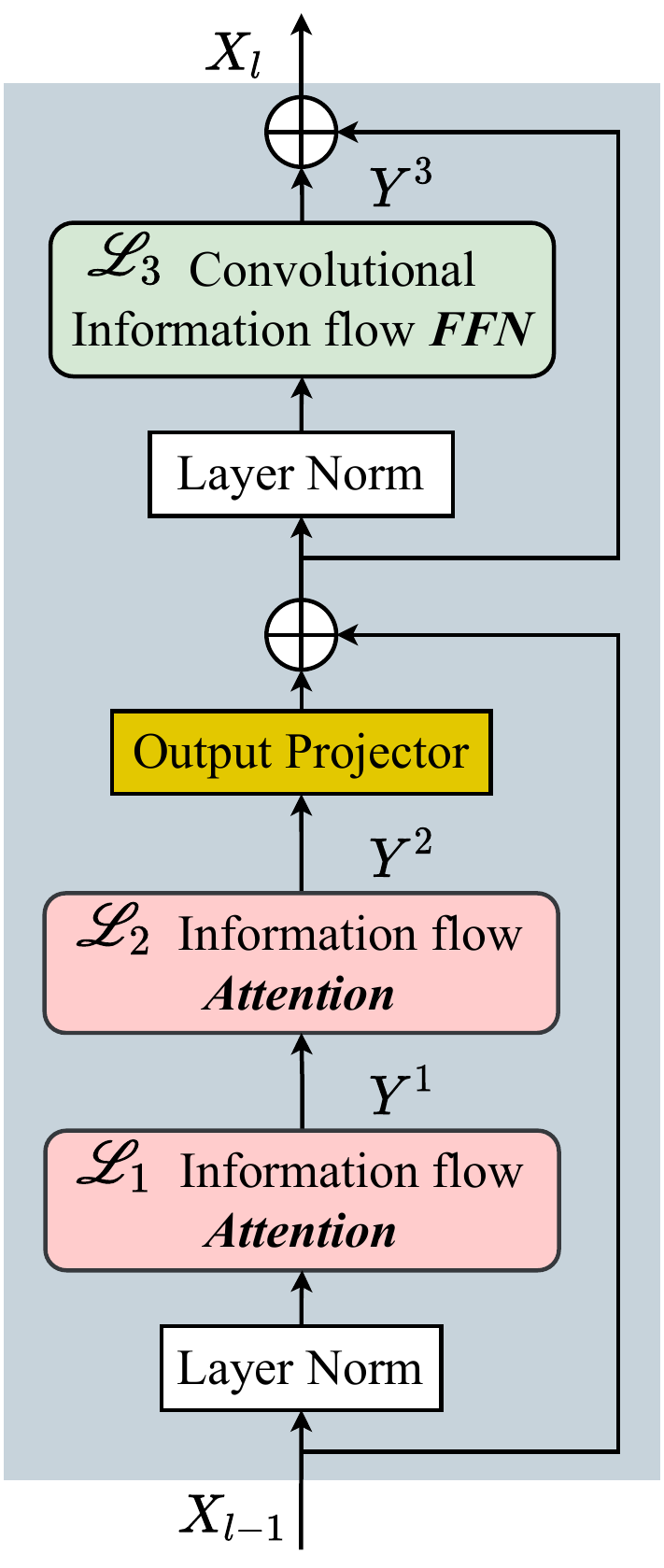}
        \subcaption{}
    \end{minipage}
    \hspace{6mm}
    \begin{minipage}{0.56\textwidth}
        \centering
        \includegraphics[height=5.5cm]{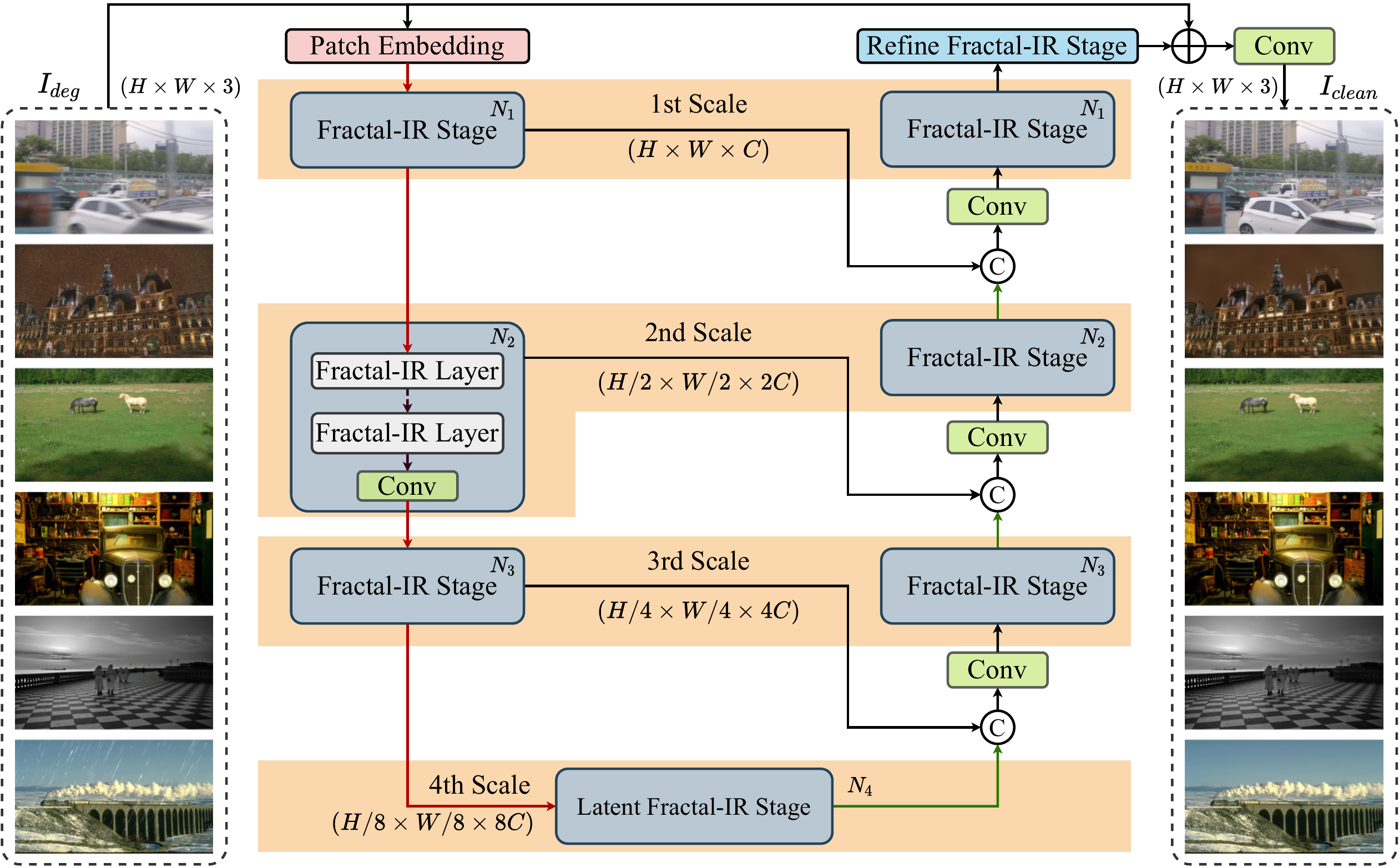}
        \subcaption{}
    \end{minipage}
    \vspace{-2mm}
    \caption{Illustrations of: 
    (a) The proposed fractal information flow.
    (b) The fractal transformer layer.
    (c) And the {\ourmethod} architecture.}
    \label{fig:architecture}
    \vspace{-4mm}
\end{figure*}

\subsection{Motivation}

\begin{table}[!ht]
\vspace{-3mm}
    \centering
    \caption{Removing shifted windows leads to degraded SR performance. PSNR is reported on Urban100 dataset for $4\times$ SR.}
    \vspace{-3mm}
    \label{tab:information_isolation}
    \setlength{\tabcolsep}{3.5pt}
    \setlength{\extrarowheight}{1.0pt}
    \scalebox{0.8}{
    \begin{tabular}{c|cc}
    \toprule[0.1em]
    \multirow{2}{*}{Training Dataset}               & \multicolumn{2}{c}{Window Shift} \\ \cline{2-3}

    & No & Yes \\ \midrule[0.1em]
    
    DF2K~\citep{agustsson2017ntire} &27.45  &27.18(\textcolor{red}{-0.27}) \\
    
    LSDIR~\citep{li2023lsdir}       &27.87	&27.64(\textcolor{red}{-0.23}) \\
    \bottomrule[0.1em]
    \end{tabular}}
    \vspace{-3mm}
\end{table}

\begin{table}[!ht]
    \centering
    \caption{Plateau effect of enlarged window size \revise{reported for $4\times$ SR}. Window size larger than 32 is not investigated due to OOM.}
    \vspace{-3mm}
    \label{tab:plateau}
    \setlength{\extrarowheight}{1.0pt}
    \setlength{\tabcolsep}{3.0pt}
    \scalebox{0.8}{
    \begin{tabular}{c|ccccc}
    \toprule[0.1em]
    Test set & Window size & PSNR & PSNR gain & GPU Mem. & Computation \\ \hline
    \multirow{3}{*}{Urban100} &8	&27.42	&0.00	&14.63GB	& $1\times $\\
    &16	&27.80	&0.38	&17.22GB	& $4\times $\\
    &32	&28.03	&0.22	&27.80GB	& \textcolor{red}{$16\times $}\\ \midrule
    \end{tabular}
}
    \vspace{-2mm}
\end{table}

This paper aims to propose an efficient and scalable framework for generalized IR.
Before presenting technical details, we discuss the motivation behind the fractal design.

\noindent \textbf{Limitation of Localized Computation.}
IR transformers typically conduct localized computation through self-attention in manually partitioned windows, combined with a window shift mechanism~\cite{liang2021swinir,chen2023activating,li2023efficient}. 
When the flow of contextual information between different regions or features within an image is restricted, a model's ability to reconstruct high-quality images from low-quality counterparts is significantly hindered. 
This effect can be observed by isolating the information flow through disabling window shifts in Swin Transformer~\cite{liu2021swin,liu2022swin,liang2021swinir}.
As shown in Tab.~\ref{tab:information_isolation}, removing the window shift mechanism leads to a significant PSNR drop of 0.27 dB for DF2K training and 0.23 dB for LSDIR training. 
The obvious reductions indicate that information isolation degrades the performance of IR models and suggests that algorithms that effectively leverage broader contextual information can yield superior IR results.

\noindent \textbf{Saturation in Computation Scaling.}
On the other hand, we observe that self-attention computation on fully connected graphs is not always necessary or beneficial for improving the performance of IR networks~\citep{chen2021pre,zamir2022restormer}. 
%
As ViTs generate distinct graphs for each token, early attempts to facilitate global information dissemination led to the curse of dimensionality, causing quadratically increasing computational complexity with tokens~\citep{wang2020linformer,liu2021swin}.
%
Subsequent attention mechanisms, building graphs based on windows, achieve better IR results. 
However, the benefits of expanding the window size tend to plateau.
%
%
As shown in Tab.~\ref{tab:plateau}, the PSNR of the SR images improves as the window size grows from 8 to 32. 
Yet, with larger windows, the gains decrease, accompanied by a sharp increase in memory footprint and computational demands. 
This urges a reassessment of the information propagation mechanism on large windows. 
%

\noindent\textbf{Effective Fractal Design.}
The above analysis emphasizes the crucial role of effective information flow in modern architectural designs. 
CNN-based methods propagate information slowly within a small region covered by the filter (Fig.~\ref{fig:motivation}(a)).
A large receptive field has to be achieved by the stack of deep layers. 
Global attention based ViT propagates information directly across the whole sequence with a single step. 
However, the computational complexity grows quadratically with the increase of tokens (Fig.~\ref{fig:motivation}(b)). 
To address this problem, window attention in Fig.~\ref{fig:motivation}(c) propagates information across two levels but still has a limited receptive field even with shift operation.

To facilitate fast and efficient image modelling, we propose a hierarchical design with the fractal architecture shown in Fig.~\ref{fig:motivation}(d). 
In this model, information flows progressively from the local scope, aggregated in several intermediate levels, and disseminated across the whole sequence. 
This new fractal design principle is more efficient in that it enables a global understanding of the input sequence with several operations. 
\revise{The actual implementation of the fractal can be configured to ensure computational efficiency.}
One realization in this work is a three-level structure in Fig.~\ref{fig:architecture}. The space and time complexity in Appx.~2
shows that the proposed fractal design is more efficient in propogating information to the global range under similar space and time complexity of window attention.

\subsection{Fractal Image Restoration}
\label{subsec:fractalir}

The fractal computation mechanism consists of three levels and aims to model both the local and the global information for a given feature $X \in \mathbb{R}^{H \times W \times C}$ efficiently. \revise{We denote the information within $X$ as $\mathscr{L}_0$ level meta-information.}

\noindent\textbf{$\mathscr{L}_1$ Local Fractal Layer.} In the beginning of the network, computation is first done in fractal cells by \revise{applying MSA to the input feature $X$ within a $p\times p$ patch. To facilitate the MSA, the input feature is first partitioned into local patches, leading to $X'\in \mathbb{R}^{\frac{HW}{p^2}\times p^2 \times C}$. Then feature $X'$ is linearly projected into query ($Q^{1}$), key ($K^{1}$), and value ($V^{1}$). Self-attention within the local patches is denoted as 
\begin{equation}
Y_{i}^{1} = \operatorname{SoftMax}\left(
\frac{Q_{i}^{1}(K_{i}^{1})^{\top}}{\sqrt{d}}\right)V_{i}^{1},
\end{equation}
where $i$ index the windows, and $d$ is the head dimension.}
%
This process is shown in Fig.~\ref{fig:architecture}(a). 
\revise{Each node within the $Y^{1}$ grid collects the $\mathscr{L}_0$ level meta-information from its corresponding original window, marked by the same color.}

\noindent\textbf{$\mathscr{L}_2$ Non-Local Fractal Aggregation Layer.} 
%
Upon completion of the computation in the local fractal cells, information must be progressively propagated to higher fractal levels.
As indicated conceptually in Fig.~\ref{fig:motivation}(d), 
the fractal design requires that information to be progressively aggregated beyond local patches while avoiding direct extension to the global scale. This design is different from the previous operations~\citep{xiao2023random,huang2021shuffle} and
is driven by two key considerations: 1) {\textit{The computational complexity of attention in the global image can be quite high}}; 2) {\textit{Not all global image information is relevant to the reconstruction of a specific pixel}}.
Thus, guided by this design principal, 2D $s \times s$ non-overlapping local patches $p \times p$ in $\mathscr{L}_1$ layer should be grouped together to form a broader $P \times P$ region for $\mathscr{L}_2$ layer. 
Then the dispersed pixels need to be grouped together.
The seemingly complex operation is simplified by a reshaping operation and a permutation operation, \ie 
\begin{align}
    \hat{Y} \in \mathbb{R}^{\frac{H}{P} \times s \times p \times \frac{W}{P} \times s \times p \times C} & = \operatorname{Reshape}(Y^1), \\
    Y' \in \mathbb{R}^{(\frac{H}{P} \times \frac{W}{P} \times p^2) \times s^2 \times C} & = \operatorname{Permute}(\hat{Y}).
\end{align}
%
The simple permutation operation facilitates the distribution of $\mathscr{L}_1$ information nodes across a higher $\mathscr{L}_2$ level.

To better integrate the permuted information $Y^{\prime}$, we further project $Y^{\prime}$ to $Q^{2}$, $K^{2}$, and $V^{2}$ and conduct another self-attention to get $Y^2$. 
%
As a result, the larger patch-wise global information denoted as colorful nodes in $Y^{1}$ of Fig.~\ref{fig:architecture} now is well propagated to each triangle node in $Y^{2}$ of Fig.~\ref{fig:architecture}.

\noindent\textbf{$\mathscr{L}_3$ Global Modelling Layer.} Finally, non-local information in the broader $\mathscr{L}_3$ region needs to be aggregated to form a global understanding of the image. To comply with previous designs of transformers, this is achieved by inserting convolutions into the feed-forward network (FFN) of transformers. This layer leads to the final representation $Y^{3}$.
As a result, this design not only aggregates all the channel-wise information more efficiently but also enriches the inductive modeling ability~\citep{chu2022conditional,xu2021vitae} for the proposed mechanism.

\subsection{{\ourmethod} Layer}
\label{subsec:fractalir_layer}
The {\ourmethod} layer is constructed based on the fractal information flow mechanism (FIFM). The detailed structure shown in Fig.~\ref{fig:architecture}(b) serves as the basic component of the proposed IR network. 
For each {\ourmethod} layer, the input feature $X_{l-1}$ first passes through a layer normalization and two consecutive information propagation attentions.
After adding the shortcut, the output $X^{'}_{l}$ is fed into the convolutional FFN with another shortcut connection and outputs $X_{l}$. 
We formulate this process as follows:
\begin{equation}
    \begin{aligned}
    X^{\prime}{ }_l & =\operatorname{FIFM_{Att}}\left(\mathrm{LN}\left(X_{l-1}\right)\right)+X_{l-1}, \\
    X_l & =\operatorname{FIFM_{Conv}}\left(\operatorname{LN}\left(X^{\prime}{ }_l\right)\right)+X^{\prime}{ }_l, 
    \end{aligned}
\end{equation}
where $\operatorname{FIFM_{Att}}$ consists of both the $\mathscr{L}_1$ and $\mathscr{L}_2$ hierarchies,  and $\operatorname{FIFM_{Conv}}$ denotes the $\mathscr{L}_3$ module.

\subsection{Overall architecture}
To comprehensively validate the effectiveness of the proposed method, similar to prior methods~\citep{chen2022simple,li2023efficient,ren2024sharing}, we choose two commonly used basic architectures including the U-shape hierarchical architecture shown in Fig.~\ref{fig:architecture}(c) and the columnar architecture shown in Appx~1.1.
The columnar architecture is used for image SR while the U-shape architecture is used for other IR tasks.
Specifically, the degraded low-quality image $I_{low} \in \mathbb{R}^{H \times W \times 1/3}$ (1 for the grayscale image and 3 for the color image) is first sent to the convolutional feature extractor and outputs the shallow feature $F_{in} \in \mathbb{R}^{H \times W \times C}$ for the following {\ourmethod} stages/layers. 
For the U-shape architecture, $F_{in}$ undergoes representation learning within the U-shape structure. In contrast, for the columnar architecture, $F_{in}$ traverses through $N$ consecutive {\ourmethod} stages.
Both architectures ultimately generate a restored high-quality image $I_{high}$ through their respective image reconstructions. 

\section{Scaling Up {\ourmethod}}
\label{sec:model_scaling_up}

Existing IR models are limited to a model size of 10-20M parameters.
In this paper, we develop models of medium and large sizes.
However, scaling up the model leads to a performance drop as shown in the pink rows of Tab.~\ref{tab:scaling_up_convergence}. 

\begin{table}[!ht]
\vspace{-2mm}
    \centering
    \caption{Model scaling-up exploration with SR.}
    \label{tab:scaling_up_convergence}
    \vspace{-3mm}
    \setlength{\extrarowheight}{0.7pt}
    \setlength{\tabcolsep}{2pt}
    \scalebox{0.65}{
        \begin{tabular}{c|c|cc|ccccc}
        \toprule[0.1em]
        \multirow{2}{*}{\textbf{Scale}}  &  \multirow{2}{*}{\makecell{\textbf{Model} \\ \textbf{Size}}} & \multirow{2}{*}{\makecell{\textbf{Warm} \\ \textbf{up}}} & \multirow{2}{*}{\makecell{\textbf{Conv} \\ \textbf{Type}}}  &  \multicolumn{5}{c}{\textbf{PSNR}} 
        \\\cline{5-9}
        &   & & & \textbf{Set5}& \textbf{Set14} & \textbf{BSD100} & \textbf{Urban100} & \textbf{Manga109}
        \\ \hline
        \myrowcolourpink $2\times$	
         &15.69	&No	&\texttt{conv1}	&38.52	&34.47	&32.56	&34.17	&39.77	\\
        \myrowcolourpink $2\times$	
        &57.60	&No	&\texttt{conv1}	&38.33	&34.17	&32.46	&33.60	&39.37	\\ 
        $2\times$	
        &57.60	&Yes	&\texttt{conv1}	&38.41	&34.33	&32.50	&33.80	&39.51	\\
        $2\times$	
        &54.23	&Yes	&\texttt{linear}	&\sotab{38.56}	&\sotaa{34.59}	&\sotaa{32.58}	&\sotab{34.32}	&\sotab{39.87}	\\
        $2\times$	
        &55.73	&Yes	&\texttt{conv3}	&\sotaa{38.65}	&\sotab{34.48}	&\sotaa{32.58}	&\sotaa{34.33}	&\sotaa{40.12}	\\
        \bottomrule[0.1em]
        \end{tabular}
    }
    \vspace{-6mm}
\end{table}

\begin{table}[!ht]
    \centering
    \caption{Investigated weight intialization and rescaling method for model scaling-up. Results reported for SR on Set5.}
    \label{tab:scaling_up_initialization_rescaling}
    \vspace{-3mm}
    \setlength{\extrarowheight}{0.7pt}
    \setlength{\tabcolsep}{2pt}
    \scalebox{0.68}{
        \begin{tabular}{llcc}
            \toprule[0.1em]
             \textbf{Method} & \textbf{Description} &$2\times$ & $3\times$ \\ \midrule[0.1em]
             Zero LayerNorm & Initialize the weight and bias of LayerNorm as 0~\citep{liu2022swin}. & 38.35 & 34.81\\ 
             \myrowcolour Residual rescale & Rescale the residual blocks by a factor of 0.01~\citep{lim2017enhanced,chen2023activating}. &38.31 & 34.79 \\ 
             Weight rescale & Rescale the weights in residual blocks by 0.1~\citep{wang2018esrgan}. & 38.36 & 34.84  \\ 
             \myrowcolour trunc\_normal\_ & Truncated normal distribution & 38.33 & 34.71 \\
             \bottomrule[0.1em]
        \end{tabular}
    }
    \vspace{-6mm}
\end{table}

\begin{table}[!ht]
\scriptsize
\setlength{\tabcolsep}{3.0pt}
\setlength{\abovecaptionskip}{0cm}
\begin{center}
    \caption{Comparison of SR results between dot production attention and cosine similarity attention for scaled-up models.}
    \label{tab:comparison_attention_type}
    \begin{tabular}{c|c|ccccc}
    \toprule[0.1em]
    Scale	&Attn. type	&Set5	&Set14	&BSD100	&Urban100	&Manga109 \\
    \midrule
    $2\times$	&cosine sim	&38.43	&34.65	&32.56	&34.13	&39.69\\	
    \myrowcolour $2\times$	&dot prod	&38.56	&34.79	&32.63	&34.49	&39.89\\
    $4\times$	&cosine sim	&33.08	&29.15	&27.96	&27.90	&31.40\\
    \myrowcolour $4\times$	&dot prod	&33.14	&29.09	&27.98	&27.96	&31.44\\
    \bottomrule[0.1em]
    \end{tabular}
\end{center}
\vspace{-6mm}
\end{table}

\noindent \textbf{Initial attempts.} 
Existing methods handle this problem with weight initialization and rescaling techniques. \citet{chen2023activating} and \citet{lim2017enhanced} reduce the influence of residual convolutional blocks by scaling those branches with a small factor (0.01). \citet{wang2018esrgan} rescale the weight parameters in the residual blocks by 0.1. \cite{liu2022swin} intialize the weight and bias of LayerNorm as 0. In addition, we also tried the truncated normal distribution to initialize the weight parameters.
However, as shown in Tab.~\ref{tab:scaling_up_initialization_rescaling}, none of the four methods improves the convergence of the scaled models, indicating that they do work for the IR transformers.
%
%

\noindent \textbf{The proposed model scaling-up solution.} 
The initial investigation indicates that the problem can be attributed to the training strategy, the initialization of the weight, and the model design.
Thus, three methods are proposed to mitigate the model scaling problem. 
\textit{First}, we warm up the training for 50k iterations at the beginning. 
As shown in Tab.~\ref{tab:scaling_up_convergence}, this mitigates the problem of degraded performance of scaled up models, but does not solve it completely. 
\textit{Secondly}, we additionally replace heavyweight $3\times3$ convolution (\texttt{conv1} in Tab.~\ref{tab:scaling_up_convergence})
with lightweight operations besides warming up the training.
Two alternatives are considered including a linear layer (\texttt{linear} in Tab.~\ref{tab:scaling_up_convergence}) and a bottleneck block with 3 lightweight convolutions ($1\times1$ conv+$3\times3$ conv+$1\times1$ conv, \texttt{conv3} in Tab.~\ref{tab:scaling_up_convergence}). 
%
Tab.~\ref{tab:scaling_up_convergence} shows that removing the large $3\times3$ convolutions leads to a much better convergence point for the large models. 
Since the bottleneck block leads to better PSNR than linear layers in most cases, it is used in the other experiments.
\textit{Thirdly}, we investigate the influence of dot product attention~\citep{liu2021swin} and cosine similarity attention~\citep{liu2022swin} on the convergence of the large models. 
Tab.~\ref{tab:comparison_attention_type} shows that dot product self-attention performs better than cosine similarity self-attention. Thus, dot product self-attention is used throughout this paper unless otherwise stated. 

\noindent \textbf{Why does replacing heavyweight convolution work?}
We hypothesize that this strategy works because of the initialization and backpropagation of the network. 
In Xavier and Kaiming weight initialization, the magnitude of the weights is inversely related to \texttt{fan\_in} / \texttt{fan\_out} of a layer, \ie
\begin{align}
    f_{in} &= c_{in} \times k ^2, \\
    f_{out} &= c_{out} \times k ^2,
\end{align}
where $f_{in}$ and $f_{out}$ denotes \texttt{fan\_in} and \texttt{fan\_out}, $c_{in}$ and $c_{out}$ denotes input and output channels, and $k$ is kernel size. 
When a dense $3\times 3$ convolution is used, $f_{in}$ and $f_{out}$ can be large, which leads to small initialized weight parameters. This in turn leads to small gradients during the backpropagation. When the network gets deeper, the vanishing gradients could lead to slow convergence. When dense $3\times3$ convolution is replaced by linear layers or bottleneck blocks, either the kernel size or the number of channels is reduced. Thus, both the two measures decreases the \texttt{fan\_in} and \texttt{fan\_out}, leading to larger initialized weights.

\noindent \textbf{Why does warmup work?}
Warmup is effective for training large models primarily because it mitigates issues related to unstable gradients and helps the optimizer gradually adapt to the model's large parameter space \citep{kalra2024warmup,goyal2017accurate}. In the early stages of training, the model's parameters are initialized randomly. A high learning rate at this stage can cause large updates, leading to unstable or divergent training due to exploding or vanishing gradients. Warmup starts with a small learning rate and gradually increases it, allowing the optimizer to find a stable path in the loss landscape before applying larger updates. 

\begin{figure}[!tb]
    \centering
    \includegraphics[width=1\linewidth]{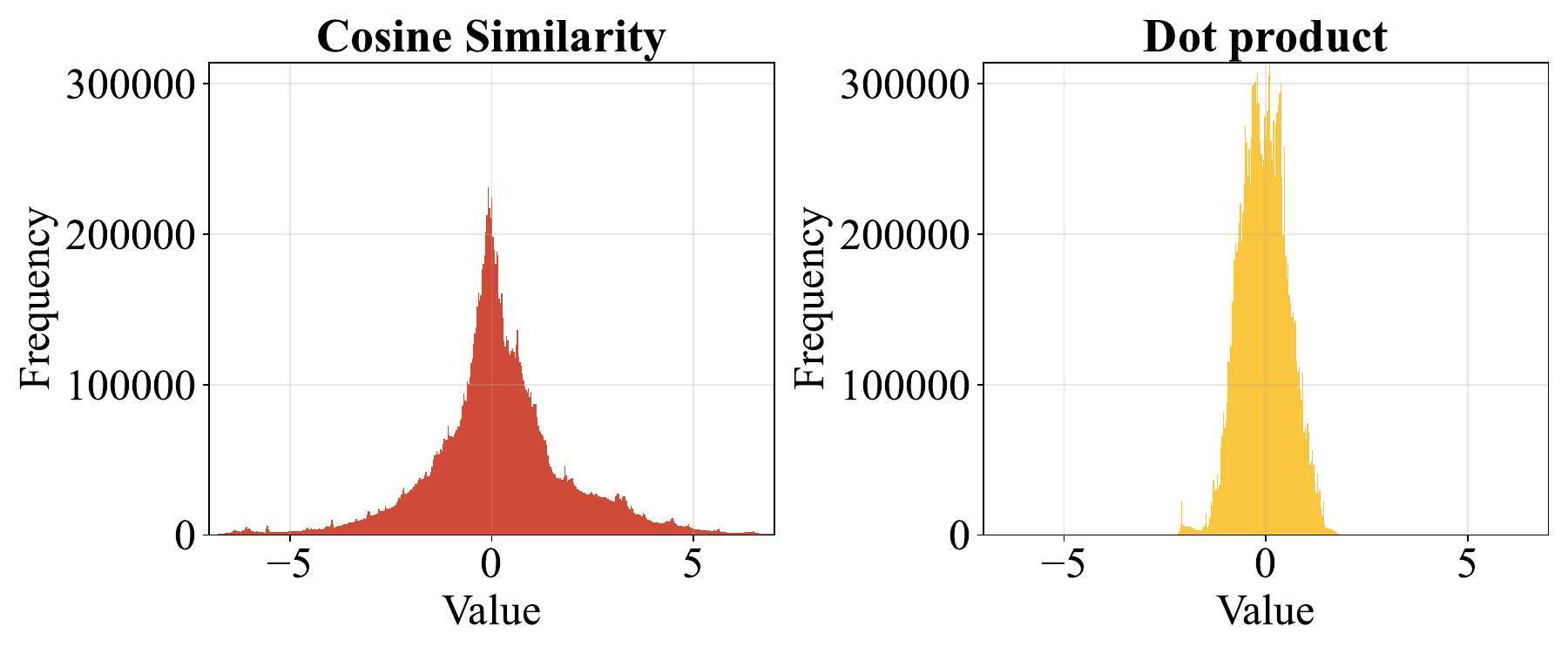}
    \vspace{-8mm}
    \caption{\revise{Gradients of cosine similarity and dot product attention.}}
    \vspace{-4mm}
    \label{fig:gradient_comparison}
\end{figure}

\noindent \textbf{\revise{Why does dot product work better?}}
We analyze the gradient of dot product and cosine similarity as follows. Suppose $\mathbf{q}$ denotes the query and $\mathbf{k}$ denotes the keys. Then dot product and cosine similarity between $\mathbf{q}$ and $\mathbf{k}$ are denoted as $\text{dot\_prod}(\mathbf{q}, \mathbf{k})$ and $\text{cos\_sim}(\mathbf{q}, \mathbf{k})$. 
The gradient of dot product and cosine similarity with respect to $\mathbf{q}$ are
\begin{align}
        \frac{\partial}{\partial \mathbf{q}} \text{dot\_prod}(\mathbf{q}, \mathbf{k}) & = \mathbf{k}, \\
        \frac{\partial}{\partial \mathbf{q}} \text{cos\_sim}(\mathbf{q}, \mathbf{k}) &  = \frac{1}{\|\mathbf{q}\|} \left(\mathbf{\hat{k}} - \text{cos\_sim}(\mathbf{q}, \mathbf{k}) \mathbf{\hat{q}}\right),
\end{align}
\revise{where $\mathbf{\hat{q}}$ and $\mathbf{\hat{k}}$ are normalized $\mathbf{q}$ and $\mathbf{k}$.}
\revise{The gradients with respect to $\mathbf{k}$ have the similar form. The gradient of cosine similarity involves more terms compared to the gradient of the dot product. This increased complexity in the gradient of cosine similarity makes it more prone to producing large or even unstable gradient values. We conducted a numerical analysis of the gradient values for the two attention methods
As shown in Fig.~\ref{fig:gradient_comparison}, the gradient of cosine similarity is indeed more prone to producing large values. This issue becomes more pronounced as the model scales up.}

\section{Experiments}
\label{sec:experiments}

In this section, 
%
we validate the effectiveness of {\ourmethod} on \textbf{7} IR tasks, \ie image SR, image Dn, JPEG image compression artifact removal (JPEG CAR), single-image motion deblurring, defocus deblurring and image demosaicking, and IR in adverse weather conditions (AWC). 
More details about the training protocols and the training/test datasets are shown in Appx.~1.
%
The best and the second-best quantitative results are reported in \sotaa{red} and \sotab{blue}. 
%

\subsection{Ablation Studies}

\noindent\textbf{Effect of $\mathscr{L}_1$ and $\mathscr{L}_2$ information flow.} 
One design choice for the $\mathscr{L}_1$/$\mathscr{L}_2$ information flow attentions is to decide whether to interleave them across transformer layers or to implement them in the same layer. To validate this choice, we develop three versions (Fig.~C, Appx.~4), including v1 where $\mathscr{L}_1$ and $\mathscr{L}_2$ attentions alternate in consecutive layers, v2 and v3 where $\mathscr{L}_1$ and $\mathscr{L}_2$ attentions are used in the same layer.
Compared with v1, v2 showed reduced performance despite increased model complexity. To address this issue, we introduce v3, where the projection layer between $\mathscr{L}_1$ and $\mathscr{L}_2$ is removed and the dimension of ${Q}$ and ${K}$ in $\mathscr{L}_1$/$\mathscr{L}_2$ attention is reduced by half to save computational complexities. 
%
Our ablation study reveals that v3 yielded the best performance, as evidenced by the results in Tab.~\ref{tab:ablation_sr_model_design}. Thus, v3 was adopted for all subsequent experiments.

\begin{table}[!tb]
\centering
\caption{Ablation study on model design with SR.}
\vspace{-3mm}
\label{tab:ablation_sr_model_design}
\setlength{\extrarowheight}{0.7pt}
\setlength{\tabcolsep}{0.7pt}
\scalebox{0.75}{

    \begin{tabular}{c|c|cc|cc}
    \toprule[0.1em]
    
    \multirow{3}{*}{\textbf{Scale}}	& \multirow{3}{*}{\textbf{\makecell{L1/L2 \\ Version}}}	& \multicolumn{4}{c}{\textbf{L3 Version}} \\ \cline{3-6}
    & & \multicolumn{2}{c|}{\textbf{Model size [M]}} & \multicolumn{2}{c}{\textbf{PSNR}} \\ \cline{3-6}
    & & \textbf{with} L3 & \textbf{w/o} L3 & \textbf{with} L3 & \textbf{w/o} L3 \\ \midrule[0.1em]
    $2\times$	&v1	&14.35	&11.87	&38.34	&38.31	\\
    $2\times$	&v2	&19.22	&16.74	&38.30	&38.22	\\
    $2\times$	&v3	&15.69	&13.21	&38.37	&38.35	\\ 
    $2\times$	&\revise{v4}	&\revise{17.19}	& -	&\revise{38.41}	&-	\\
    
    \bottomrule[0.1em]
    \end{tabular}
    }
    \vspace{-2mm}
\end{table}
\begin{table}[!tb]
\centering
    \caption{Model efficiency \vs accuracy for SR and Dn. PSNR is reported on Urban100 dataset.}
    \label{tab:params_flops_runtime}
\vspace{-3mm}
\setlength{\extrarowheight}{0.7pt}
\setlength{\tabcolsep}{0.6pt}
\scalebox{0.75}{

    \begin{tabular}{c|l|c|cccc}
        \toprule[0.1em]
        \multirow{2}{*}{\textbf{Task}}	&\multirow{2}{*}{\textbf{Network}}	&\multirow{2}{*}{\textbf{Arch.}}	&\textbf{Params}	&\textbf{FLOPs}	&\textbf{Runtime}	&\textbf{PSNR}	\\
        & & & \textbf{[M]} & \textbf{[G]} & \textbf{[ms]} & \textbf{[dB]} \\  \hline
        \multirow{4}{*}{$4\times$ SR}	&\revise{SwinIR}~\citep{liang2021swinir}	&Columnar	&11.90	&215.32	&152.24	&27.45	\\
        	&\revise{CAT}~\citep{chen2022cross}	&Columnar	&16.60	&387.86	&357.97	&27.89	\\
        	&\revise{HAT}~\citep{chen2023activating}	&Columnar	&20.77	&416.90	&368.61	&28.37	\\ 
        	&Fractal-IR (Ours)	&Columnar	&14.83	&287.20	&331.92	&28.44	\\ \midrule
        \multirow{4}{*}{Dn 50}	&\revise{SwinIR}~\citep{liang2021swinir}	&Columnar	&11.50	&804.66	&1772.84	&27.98	\\
        	&\revise{Restormer}~\citep{zamir2022restormer}	&U-shape	&26.13	&154.88	&210.44	&28.29	\\
        	&\revise{GRL}~\citep{li2023efficient}	&Columnar	&19.81	&1361.77	&3944.17	&28.59	\\
        	&Fractal-IR (Ours)	&U-shape	&22.33	&153.66	&399.05	&28.91	\\ \bottomrule[0.1em]
    \end{tabular}
    }
    \vspace{-4mm}
\end{table}

\begin{table*}[!t]
\parbox{0.6\linewidth}{
\centering
    \caption{\textbf{\textit{Classical image SR}} results. Top-2 results are highlighted in  \textcolor{red}{red} and  \textcolor{blue}{blue}.}%
    \label{tab:sr_results_main}
\vspace{-3mm}
\setlength{\extrarowheight}{0.7pt}
\setlength{\tabcolsep}{1.5pt}
\scalebox{0.65}{
        \begin{tabular}{l|c|r|cc|cc|cc|cc|cc}
        \toprule[0.1em]
        \multirow{2}{*}{\textbf{Method}} & \multirow{2}{*}{\textbf{Scale}} & {\textbf{Params}} &  \multicolumn{2}{c|}{\textbf{Set5}} &  \multicolumn{2}{c|}{\textbf{Set14}} &  \multicolumn{2}{c|}{\textbf{BSD100}} &  \multicolumn{2}{c|}{\textbf{Urban100}} &  \multicolumn{2}{c}{\textbf{Manga109}} 
        \\
        \cline{4-13}
        &  & \multicolumn{1}{c|}{\textbf{[M]}} & PSNR$\uparrow$ & SSIM$\uparrow$ & PSNR$\uparrow$ & SSIM$\uparrow$ & PSNR$\uparrow$ & SSIM$\uparrow$ & PSNR$\uparrow$ & SSIM$\uparrow$ & PSNR$\uparrow$ & SSIM$\uparrow$
        \\
        \midrule[0.1em]
        \myrowcolour EDSR~\citep{lim2017enhanced} & $2\times$ & 40.73 & 38.11 & 0.9602 & 33.92 & 0.9195 & 32.32 & 0.9013 & 32.93 & 0.9351 & 39.10 & 0.9773\\
        RCAN~\citep{zhang2018rcan}&	$2\times$&	15.44&	38.27&	0.9614&	34.12&	0.9216&	32.41&	0.9027&	33.34&	0.9384&	39.44&	0.9786\\
        \myrowcolour SAN~\citep{dai2019SAN}&	$2\times$&	15.71&	38.31&	0.9620&	34.07&	0.9213&	32.42&	0.9028&	33.10&	0.9370&	39.32&	0.9792\\

        
            
        NLSA~\citep{mei2021NLSA}&	$2\times$&	42.63&	38.34&	0.9618&	34.08&	0.9231&	32.43&	0.9027&	33.42&	0.9394&	39.59&	0.9789\\
        \myrowcolour IPT~\citep{chen2021pre}&	$2\times$&	115.48&	38.37&	-&	34.43&	-&	32.48&	-&	33.76&	-&	-&	-\\ \hline
        
        SwinIR~\citep{liang2021swinir}&	$2\times$&	11.75&	38.42&	0.9623&	34.46&	0.9250&	32.53&	0.9041&	33.81&	0.9427&	39.92&	0.9797\\  
        \myrowcolour CAT-A~\citep{chen2022cross}    & $2\times$ &16.46 & 38.51 & 0.9626 & 34.78 & 0.9265 & 32.59 & 0.9047 & 34.26 & 0.9440 & 40.10 & 0.9805 \\
        ART~\citep{zhang2022accurate}	&$2\times$ &16.40	&38.56	&0.9629	&34.59	&0.9267	&32.58	&0.9048	&34.3	&0.9452	&40.24	&0.9808	\\		
        \myrowcolour EDT~\citep{li2021efficient}&	$2\times$&	11.48&	 {38.63}&	 {0.9632}&	 {34.80}&	0.9273&	 {32.62}&	0.9052&	34.27&	0.9456&	 {40.37}&	 {0.9811}\\ 
        GRL-B~\citep{li2023efficient} &	$2\times$&	20.05&	 {38.67}&	 {0.9647}&	 {35.08}&	 \sotab{0.9303}&	 {32.67}&	 \sotab{0.9087}&	 {35.06}&	 \sotaa{0.9505}&	 {40.67}&	 {0.9818}\\ 
        \myrowcolour HAT~\citep{chen2023activating}	&$2\times$	&20.62	&38.73	&0.9637	&35.13	&0.9282	&32.69	&0.9060	&34.81	&\sotab{0.9489}	&40.71	&0.9819	\\
        {\ourmethod}-B 	&$2\times$& 14.68	&38.71	&\sotab{0.9657}	&35.16	&0.9299	&32.73	&\sotab{0.9087}	&34.94	&0.9484	&40.81	&0.9830			\\
        \myrowcolour HAT-L~\citep{chen2023activating}	&$2\times$	&40.70	&\sotaa{38.91}	&0.9646	&\sotaa{35.29}	&0.9293	&\sotab{32.74}	&0.9066	&\sotab{35.09}	&\sotaa{0.9505}	&\sotab{41.01}	&\sotab{0.9831}	\\
        {\ourmethod}-L 	&$2\times$& 39.07	&\sotab{38.87}	&\sotaa{0.9663}	&\sotab{35.27}	&\sotaa{0.9311}	&\sotaa{32.77}	&\sotaa{0.9092}	&\sotaa{35.16}	&\sotaa{0.9505}	&\sotaa{41.22}	&\sotaa{0.9846}			\\

         \bottomrule[0.1em]
        \end{tabular}
    }
}
\hspace{4pt}
\parbox{0.4\linewidth}{\centering
\caption{\textit{\textbf{Single-image motion deblurring}} on GoPro and HIDE dataset. {GoPro} dataset is used for training. 
}
\label{tab:motion_deblurring}
\vspace{-3mm}
\setlength{\extrarowheight}{0.7pt}
\setlength{\tabcolsep}{1.0pt}
\scalebox{0.65}{
    \begin{tabular}{l | c | c | c}
    \toprule[0.1em]
     & \textbf{GoPro} & \textbf{HIDE}  & Average \\
     \textbf{Method} & PSNR$\uparrow$ / SSIM$\uparrow$ & PSNR$\uparrow$ / SSIM$\uparrow$ & PSNR$\uparrow$ / SSIM$\uparrow$ \\
    \midrule[0.1em]
    DeblurGAN~\citep{deblurgan}	&28.70 / 0.858		&24.51 / 0.871		&26.61 / 0.865		\\
    \myrowcolour Nah~\etal~\citep{nah2017deep}	&29.08 / 0.914		&25.73 / 0.874		&27.41 / 0.894		\\
    DeblurGAN-v2~\citep{deblurganv2}	&29.55 / 0.934		&26.61 / 0.875		&28.08 / 0.905		\\
    \myrowcolour SRN~\citep{tao2018scale}	&30.26 / 0.934		&28.36 / 0.915		&29.31 / 0.925		\\
    SPAIR~\citep{purohit2021spatially_spair}	&32.06 / 0.953		&30.29 / 0.931		&31.18 / 0.942		\\
    \myrowcolour MIMO-UNet+~\citep{cho2021rethinking_mimo}&32.45 / 0.957		&29.99 / 0.930		&31.22 / 0.944		\\
    MPRNet~\citep{zamir2021multi}	&32.66 / 0.959		&30.96 / 0.939		&31.81 / 0.949		\\
    \myrowcolour MAXIM-3S~\citep{tu2022maxim}	&32.86 / 0.961 & 32.83 / 0.956 & 32.85 / 0.959	\\
    Restormer~\citep{zamir2022restormer}	&32.92 / {0.961}		&31.22 / 0.942	&32.07 / 0.952
    		\\
    \myrowcolour Stripformer~\citep{tsai2022stripformer} & 33.08 / {0.962} & 31.03 / 0.940 &32.06 / 0.951 \\
    ShuffleFormer~\citep{xiao2023random} &	33.38 / \sotab{0.965}	& 31.25 / 0.943 & 31.32 / 0.954 \\
    
    \myrowcolour GRL-B~\citep{li2023efficient}	&\sotab{33.93} / \sotaa{0.968}		&\sotaa{31.65} / \sotaa{0.947}		&\sotab{32.79} / \sotaa{0.958}	\\
    {\ourmethod}-L   &\sotaa{33.99} / \sotaa{0.968}		&\sotab{31.64} / \sotaa{0.947}		&\sotaa{32.82} / \sotaa{0.958}	\\
    \bottomrule[0.1em]
    \end{tabular}
    
    }
}
\vspace{-5mm}
\end{table*}

\begin{table*}[t]
\parbox{.7\linewidth}{
\centering
\caption{\textit{\textbf{Color and grayscale image denoising}} results. 
}
\label{tab:denoising}
\vspace{-3mm}
\setlength{\tabcolsep}{1.4pt}
\scalebox{0.65}{
\begin{tabular}{l | r | c c c | c c c | c c c | c c c | c c c }
\toprule[0.1em]
\multirow{3}{*}{\textbf{Method}} & \multicolumn{1}{c|}{\multirow{3}{*}{ \makecell{\textbf{Params} \\ \textbf{[M]}} }}
 & \multicolumn{9}{c||}{\textbf{Color}} & \multicolumn{6}{c}{\textbf{Grayscale}} \\ \cline{3-17}
& & \multicolumn{3}{c|}{\textbf{CBSD68}~\cite{martin2001database}} & 
\multicolumn{3}{c|}{\textbf{McMaster}~\cite{zhang2011color}} & \multicolumn{3}{c||}{\textbf{Urban100}~\cite{huang2015single}}  & \multicolumn{3}{c|}{\textbf{Set12}~\cite{zhang2017beyond}}  & \multicolumn{3}{c}{\textbf{Urban100}~\cite{huang2015single}} \\
        &  & $\sigma$$=$$15$ & $\sigma$$=$$25$ & $\sigma$$=$$50$ 
        & $\sigma$$=$$15$ & $\sigma$$=$$25$ & $\sigma$$=$$50$ & $\sigma$$=$$15$ & $\sigma$$=$$25$ & $\sigma$$=$$50$ & $\sigma$$=$$15$ & $\sigma$$=$$25$ & $\sigma$$=$$50$ & $\sigma$$=$$15$ & $\sigma$$=$$25$ & $\sigma$$=$$50$ \\ \midrule
\myrowcolour DnCNN~\cite{kiku2016beyond}	&0.56	&33.90	&31.24	&27.95	
&33.45	&31.52	&28.62	&32.98	&30.81	&27.59				&32.86	&30.44	&27.18				&32.64	&29.95	&26.26	\\		
IPT~\cite{chen2021pre}	&115.33	&-	&-	&28.39	
&-	&-	&29.98	&-	&-	&29.71				&-	&-	&-				&-	&-	&-	\\		
\myrowcolour EDT-B~\cite{li2021efficient}	&11.48	&34.39	&31.76	&28.56	
&35.61	&33.34	&30.25	&35.22	&33.07	&30.16				&-	&-	&-				&-	&-	&-	\\		
SwinIR~\cite{liang2021swinir}	&11.75	&34.42	&31.78	&28.56	
&35.61	&33.20	&30.22	&35.13	&32.90	&29.82				&33.36	&31.01	&27.91				&33.70	&31.30	&27.98	\\		
\myrowcolour Restormer~\cite{zamir2022restormer}	&26.13	&34.40	&31.79	&28.60	
&35.61	&33.34	&30.30	&35.13	&32.96	&30.02				&33.42	&31.08	&28.00				&33.79	&31.46	&28.29	\\		

Xformer~\cite{zhang2023xformer}	& 25.23	&\sotaa{34.43}	&\sotaa{31.82}	&\sotaa{28.63}	
&\sotab{35.68}	&\sotaa{33.44}	&\sotab{30.38}	&\sotab{35.29}	&\sotab{33.21}	&\sotab{30.36}	&\sotab{33.46}	&\sotab{31.16}	&\sotab{28.10}	&\sotab{33.98}	&\sotab{31.78}	&\sotab{28.71}	\\		
\myrowcolour {\ourmethod} 	& 22.33	&\sotab{34.42}	&\sotaa{31.82}	&\sotab{28.62}	
&\sotaa{35.69}	&\sotaa{33.44}	&\sotaa{30.42}	&\sotaa{35.46}	&\sotaa{33.34}	&\sotaa{30.59}	&\sotaa{33.48}	&\sotaa{31.19}	&\sotaa{28.15}	&\sotaa{34.11}	&\sotaa{31.92}	&\sotaa{28.91}	\\		
\bottomrule[0.1em]
\end{tabular}
}
\vspace{-2mm}
}
\parbox{0.3\linewidth}{
    \begin{center}
    \caption{\textit{\textbf{Dual-pixel defocus deblurring}}. 
    }
    \label{tab:defocus_deblurring}
    \vspace{-3mm}
    \setlength{\tabcolsep}{2.0pt}
    \setlength{\extrarowheight}{3pt}
    \scalebox{0.65}{
    \begin{tabular}{l | c | c | c | c}
    \toprule[0.1em]
    \multirow{2}{*}{Method} 
    & \multicolumn{4}{c}{\textbf{Combined Scenes}} \\ \cline{2-5}
        &PSNR$\uparrow$ & SSIM$\uparrow$ & MAE$\downarrow$ & LPIPS$\downarrow$		\\ \hline
    DPDNet~\citep{abuolaim2020defocus}	
    &25.13	&0.786	&0.041	&0.223	\\
    \myrowcolour RDPD~\citep{abdullah2021rdpd}	
    &25.39	&0.772	&0.040	&0.255	\\
    Uformer~\citep{wang2022uformer}	
    &25.65	&0.795	&0.039	&0.243	\\
    \myrowcolour IFAN~\citep{Lee_2021_CVPRifan}	
    &25.99	&0.804	&0.037	&0.207	\\
    Restormer~\citep{zamir2022restormer}	
    &\textcolor{blue}{26.66}	&\textcolor{blue}{0.833}	&\textcolor{blue}{0.035}	&\textcolor{blue}{0.155}	\\
    \myrowcolour {\ourmethod}-B 	
    &\textcolor{red}{27.01}	&\textcolor{red}{0.848}	&\textcolor{red}{0.034}	&\textcolor{red}{0.135}	\\ \bottomrule[0.1em]
    \end{tabular}}
    \end{center}
    \vspace{-2mm}
}
\vspace{-2mm}
\end{table*}

\begin{figure*}[!tb]
\begin{center}
\scriptsize
\begin{tabular}[b]{c@{ } c@{ } c@{ } c@{ } c@{ } c@{ }}
    

\hspace{-2mm}  
    \includegraphics[width=0.16\linewidth,valign=t]{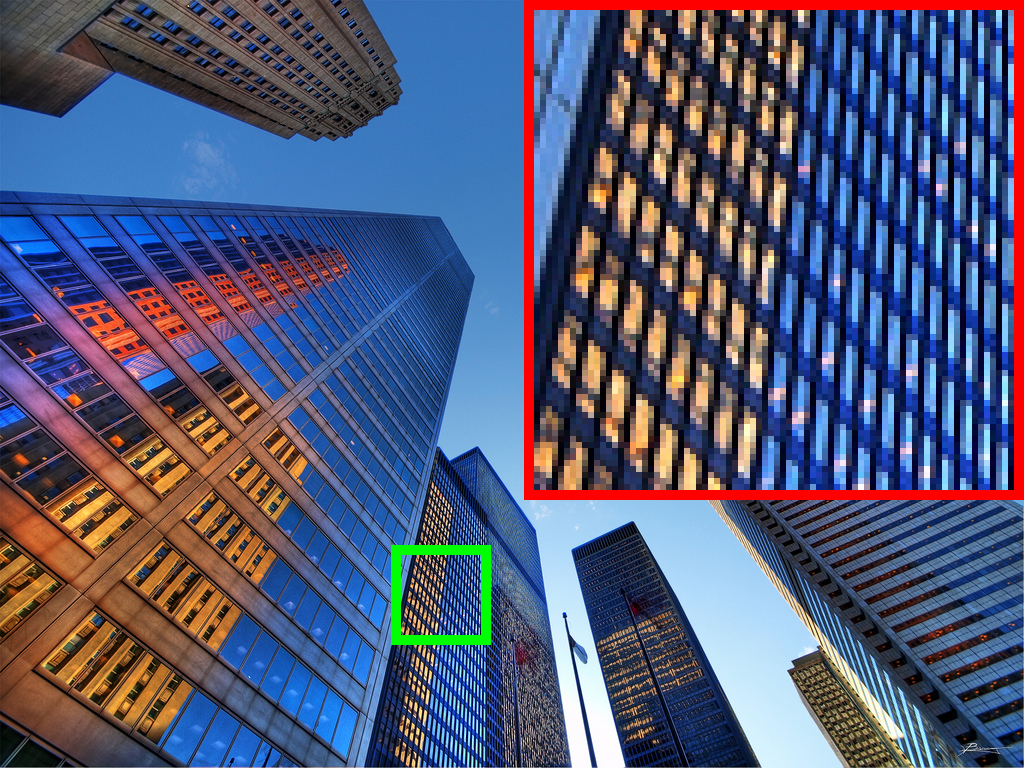} &
    \includegraphics[width=0.16\linewidth,valign=t]{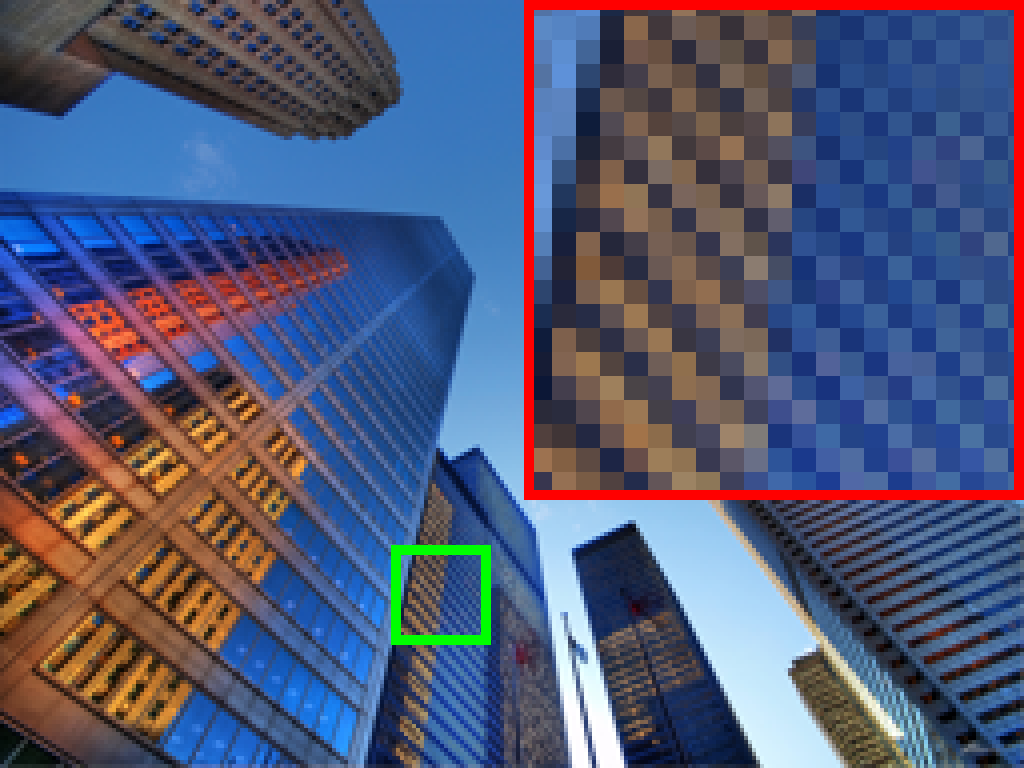} &
    \includegraphics[width=0.16\linewidth,valign=t]{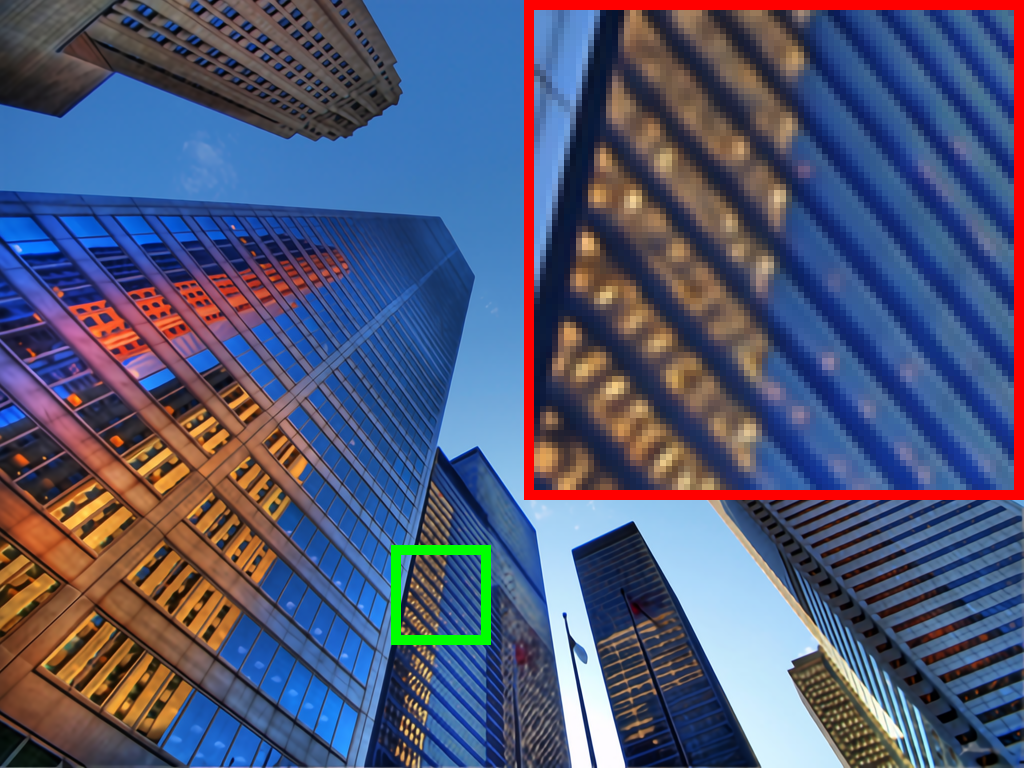} &
    \includegraphics[width=0.16\linewidth,valign=t]{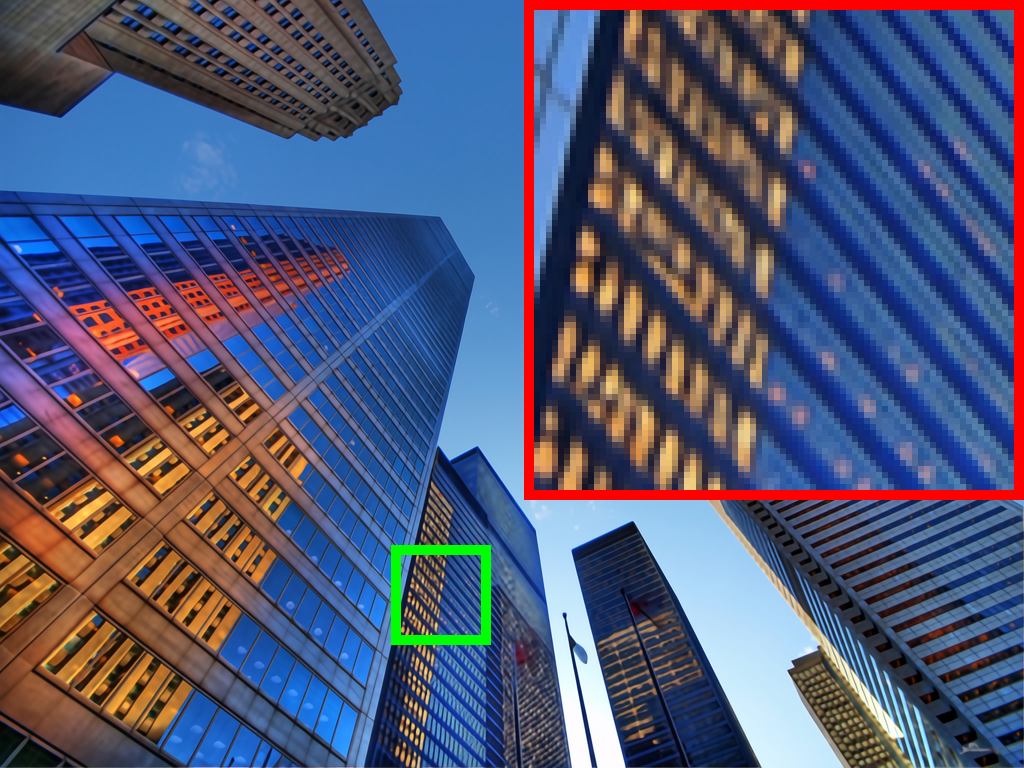} &
    \includegraphics[width=0.16\linewidth,valign=t]{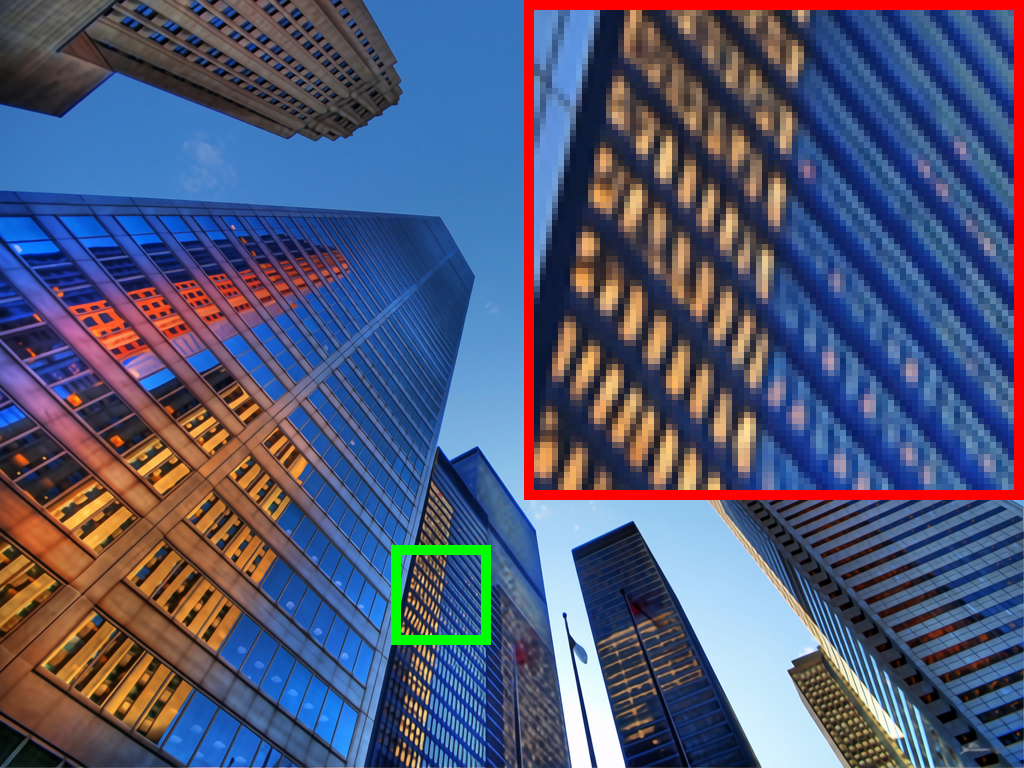} &
    \includegraphics[width=0.16\linewidth,valign=t]{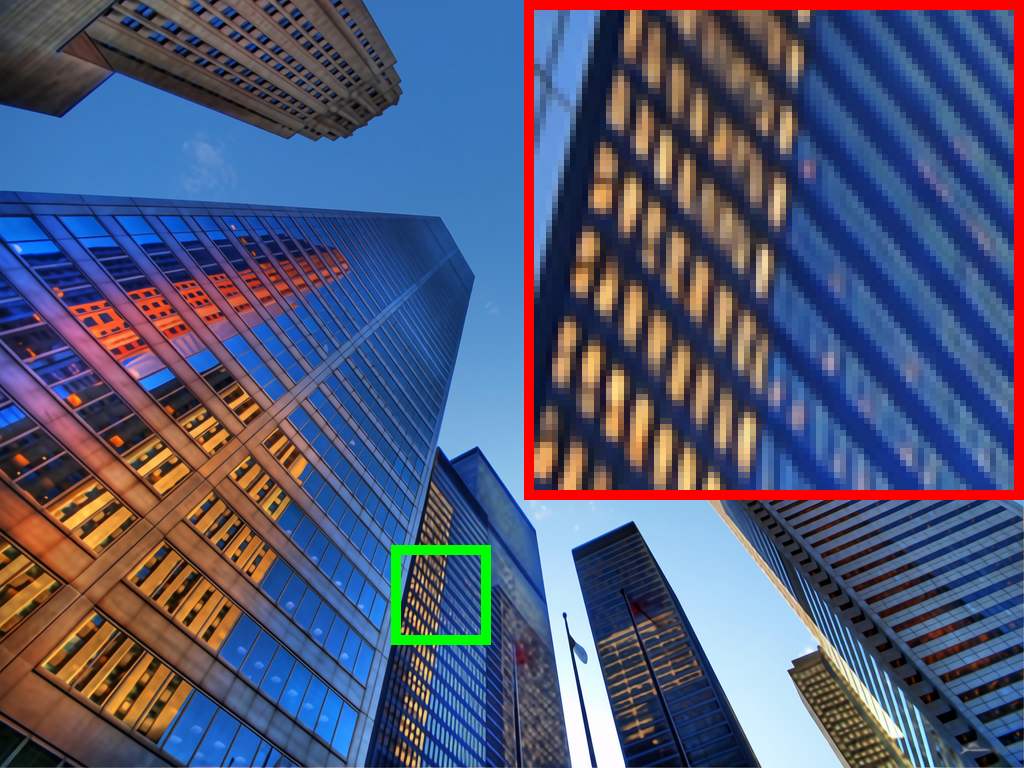} \\ 
    \revise{Groud-truth} & LR & \makecell{SwinIR \citep{liang2021swinir}} &\makecell{GRL  \citep{li2023efficient}} & \makecell{HAT  \citep{chen2023activating}} & {\ourmethod} (Ours) \\

\hspace{-2mm}  
    \includegraphics[width=0.16\linewidth,valign=t]{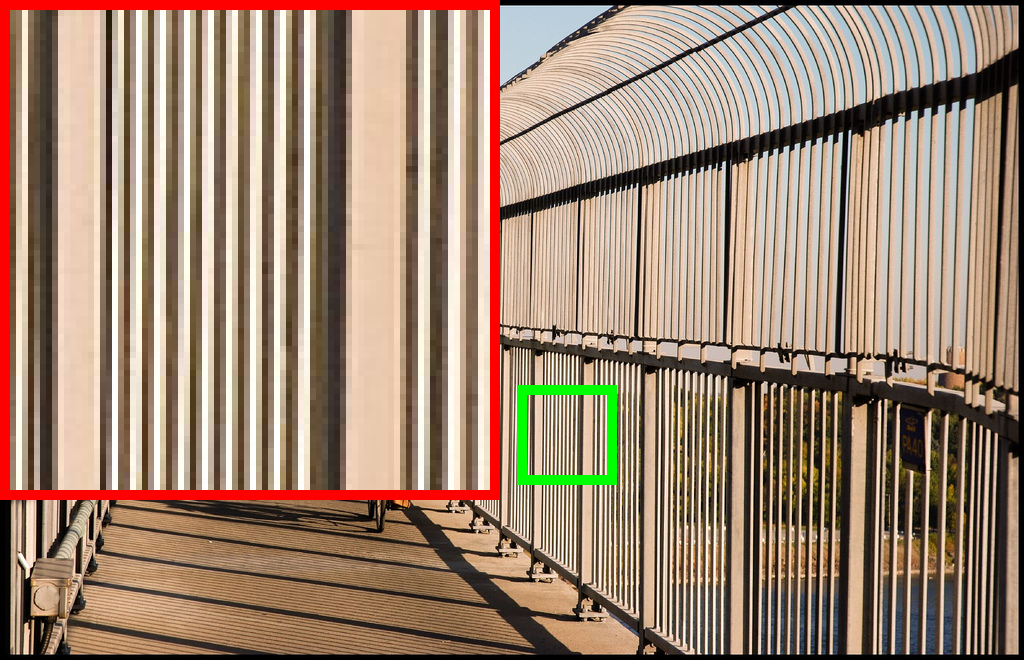} &
    \includegraphics[width=0.16\linewidth,valign=t]{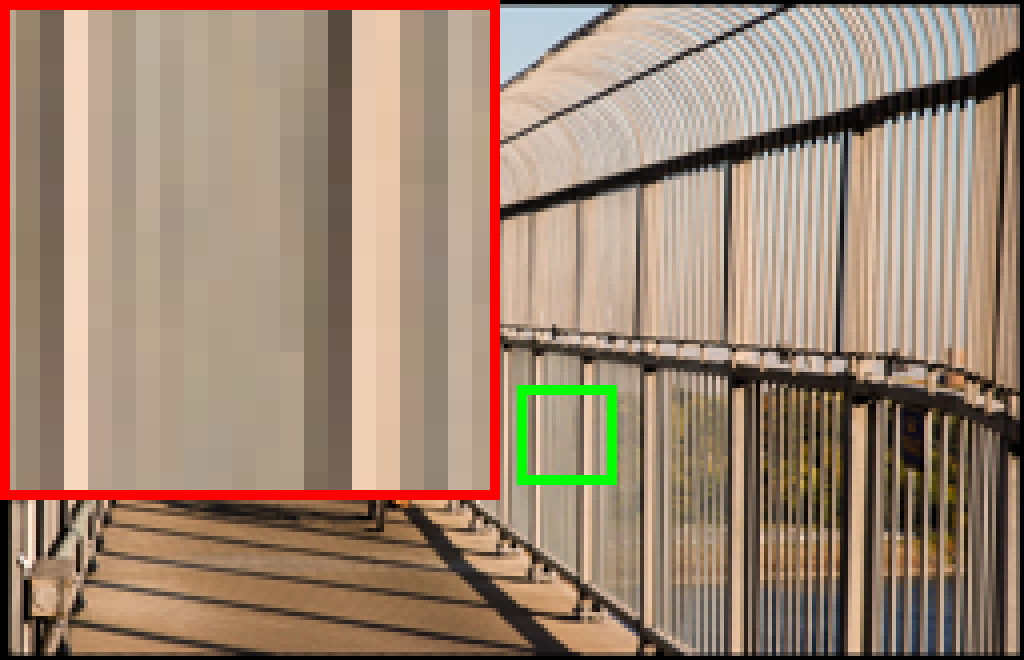} &
    \includegraphics[width=0.16\linewidth,valign=t]{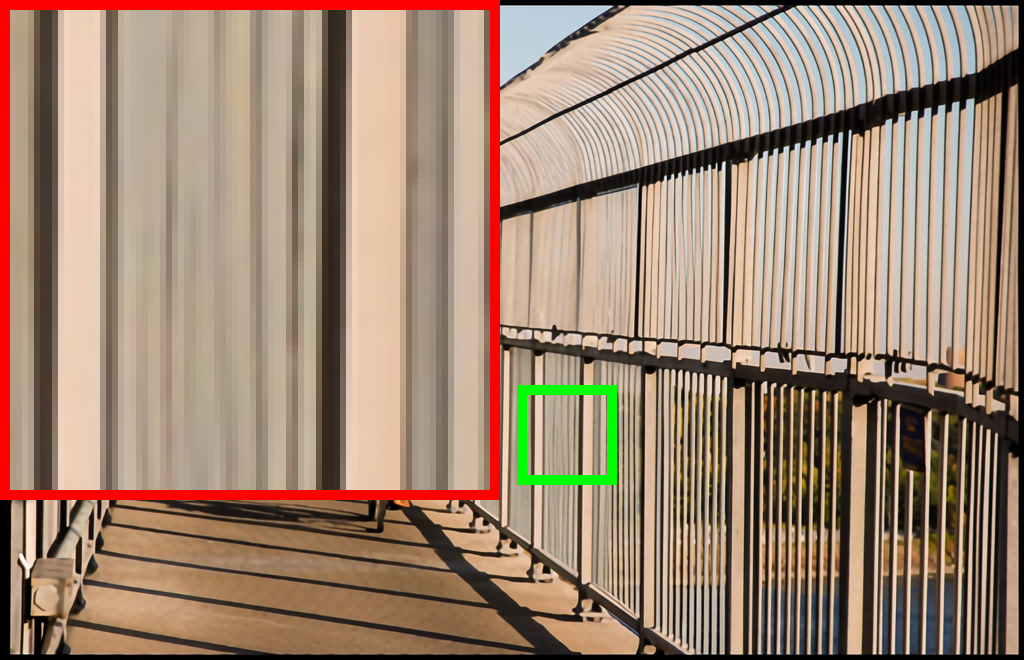} & 
    \includegraphics[width=0.16\linewidth,valign=t]{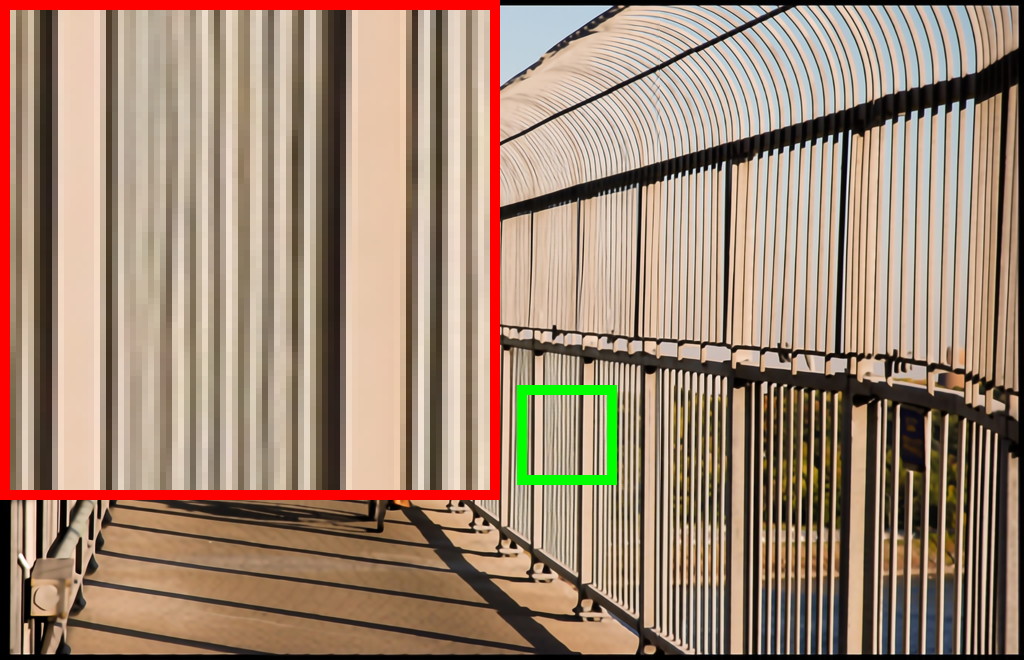} &
    \includegraphics[width=0.16\linewidth,valign=t]{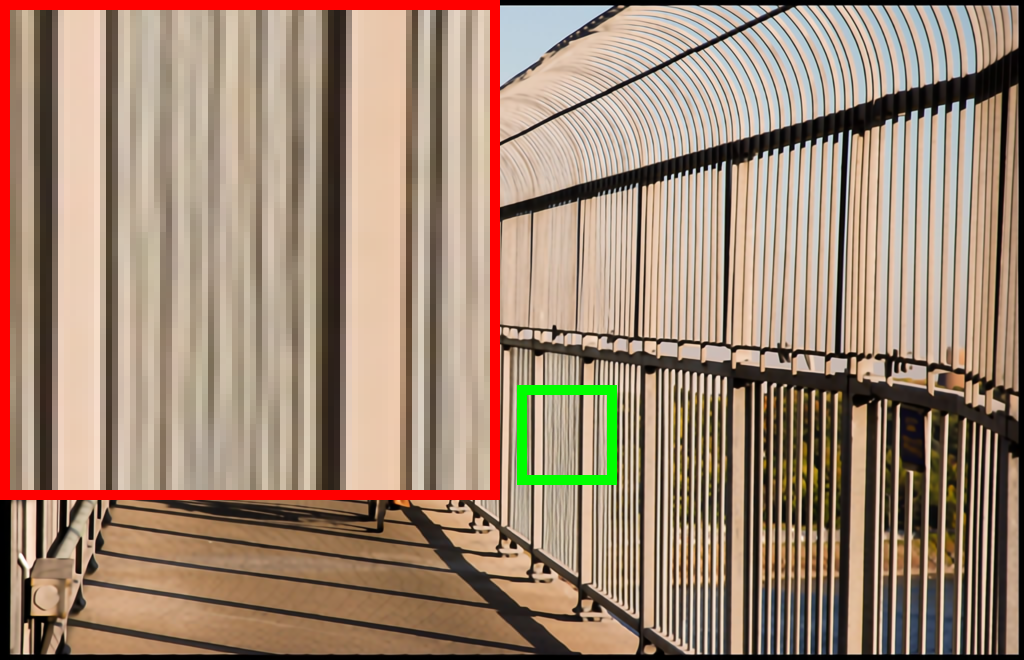} &
    \includegraphics[width=0.16\linewidth,valign=t]{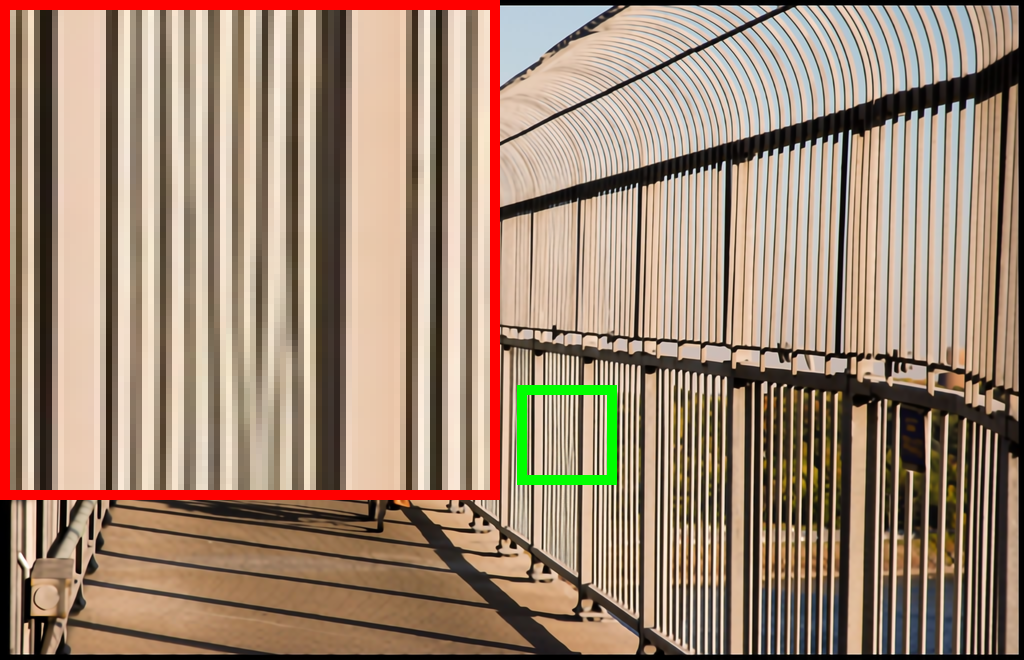} \\ 
    \revise{Groud-truth} & LR & \makecell{SwinIR  \citep{liang2021swinir}} &\makecell{GRL  \citep{li2023efficient}} & \makecell{HAT  \citep{chen2023activating}} & {\ourmethod} (Ours) \\
    
\end{tabular}
\end{center}
\vspace{-6mm}
\caption{Visual results for classical image $\times 4$ SR on Urban100 dataset.}
\label{fig:visual_sr_urban100}
\vspace{-4mm}
\end{figure*}

\noindent\textbf{\revise{Effect of the depth of the fractal structure.}} 
\revise{Ablation study was conducted to evaluate the effect of the tree structure's depth. In Tab.~\ref{tab:ablation_sr_model_design}, the depth of the tree in the v1 model is 3. Removing the $\mathscr{L}_3$ information flow reduces the depth to 2, resulting in degraded image SR performance, even on the small Set5 dataset. Additionally, a v4 model was designed by adding an information flow attention beyond $\mathscr{L}_2$ to v3 model, creating a depth-4 fractal structure. As shown in Tab.~\ref{tab:ablation_sr_model_design}, this increased complexity improves SR results. Thus, well-designed deeper tree structures lead to improved model performance but with increased model complexity.}
%

\noindent\textbf{Efficiency Analysis.} We report the efficiency comparison results on $4\times$ SR and denoising in Tab.~\ref{tab:params_flops_runtime}. 
For the columnar architecture-based SR, our {\ourmethod} achieves the best PSNR with much lower parameters (28.6\% reduction) and FLOPs (31.1\% reduction), and runtime (9.95\% reduction) compared to HAT~\citep{chen2023activating}. Similar observation can also be achieved on the denoising task.

\subsection{Evaluation of {\ourmethod} on Various IR tasks}
\noindent\textbf{Image SR.} For the classical image SR, we compared our {\ourmethod} with state-of-the-art SR models. The quantitative results are shown in Tab.~\ref{tab:sr_results_main}. More results are shown in Fig.~D of Appx.~5.
Aside from the 2nd-best PSNR on Set5 and Set14, the proposed {\ourmethod} archives the best PSNR and SSIM on all other test sets. 
In particular, significant improvements in terms of the PSNR on Urban100 (\ie 0.13 dB for $2\times$ SR of the base model) and Manga109 (\ie 0.21 dB for $2\times$ SR of the large model) compared to HAT~\citep{chen2023activating}, but with fewer trainable parameters. 
The visual results shown in Fig.~\ref{fig:visual_sr_urban100} also validate the
effectiveness of the proposed {\ourmethod} in restoring more structural content. 
%

\noindent\textbf{Image Denoising.} We provide both color and grayscale image denoising results in Tab.~\ref{tab:denoising}.
Our approach demonstrates superior performance on diverse datasets, including McMaster, and Urban100 for color images, as well as Set12 and Urban100 for grayscale images. 
For grayscale image denoising with $\sigma=50$, {\ourmethod} improves the PSNR on Set12 and Urban100 by 0.05dB and 0.20dB compared with Xformer. 
It is noteworthy that our method outperforms DRUNet, Restormer, and Xformer, despite utilizing fewer trainable parameters.
Additionally, a closer examination of more visual results is available in the appendix, further substantiates the capabilities of {\ourmethod}. 
These results illustrate its proficiency in effectively eliminating heavy noise corruption while preserving high-frequency image details. 

\begin{figure*}[!t]
    \centering
    \captionsetup[sub]{font=scriptsize} 
    \begin{subfigure}[b]{0.235\textwidth}
        \centering
        \includegraphics[width=\textwidth]{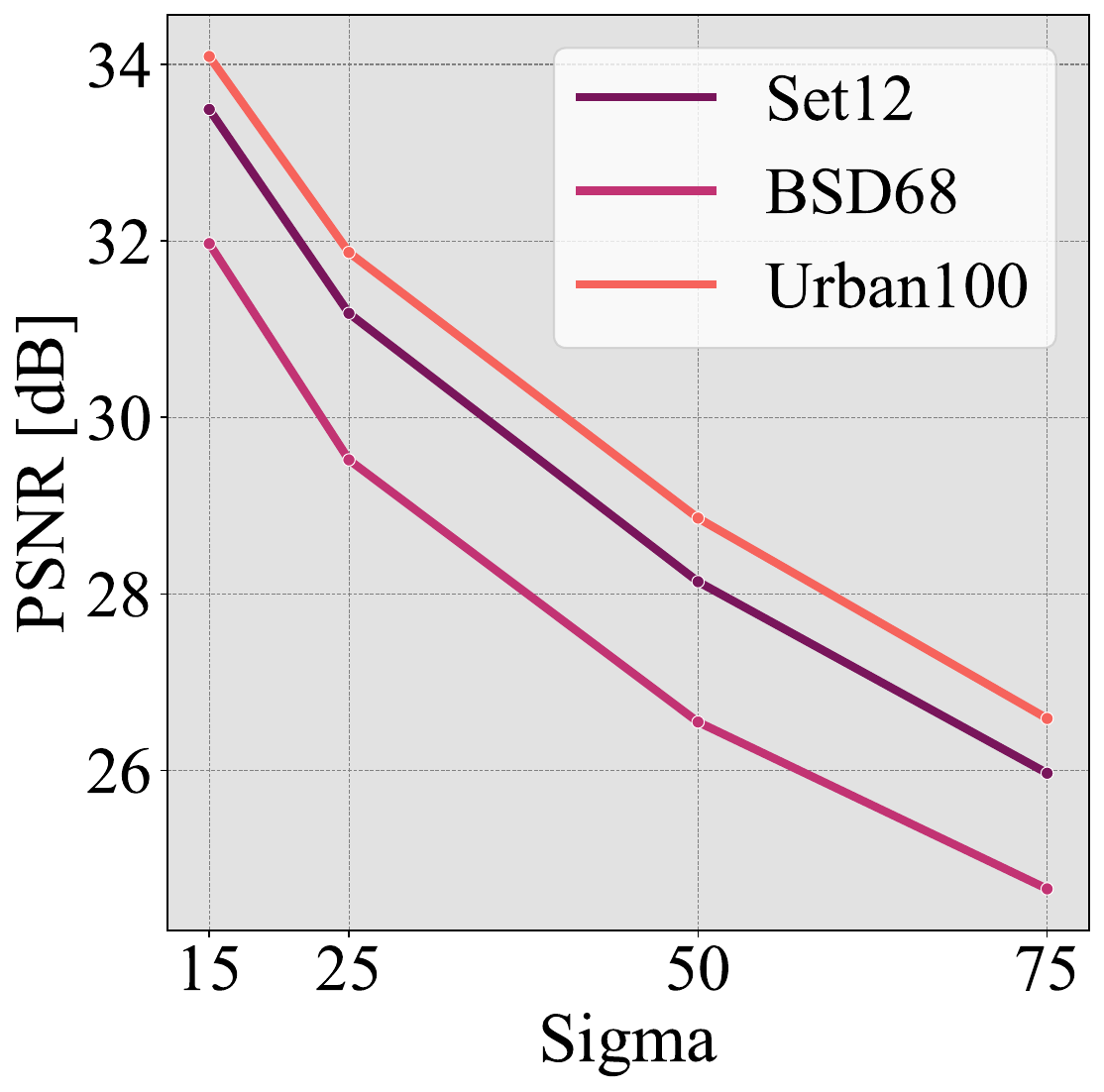}
        \caption{Grayscale image denoising}
        \label{fig:one_model1}
    \end{subfigure}
    \hfill
    \begin{subfigure}[b]{0.235\textwidth}
        \centering
        \includegraphics[width=\textwidth]{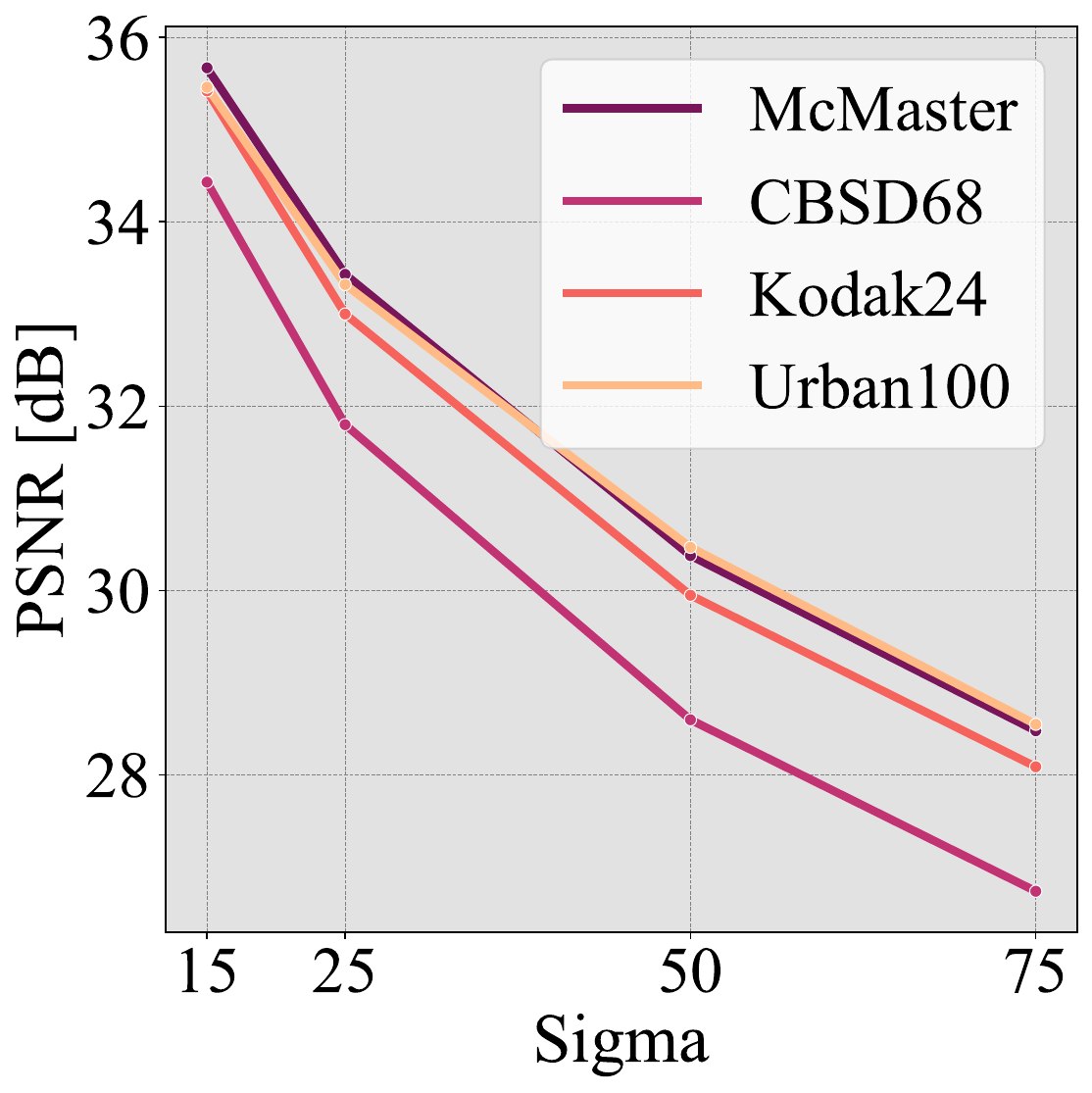}
        \caption{Color image denoising}
        \label{fig:one_model2}
    \end{subfigure}
    \hfill
    \begin{subfigure}[b]{0.235\textwidth}
        \centering
        \includegraphics[width=\textwidth]{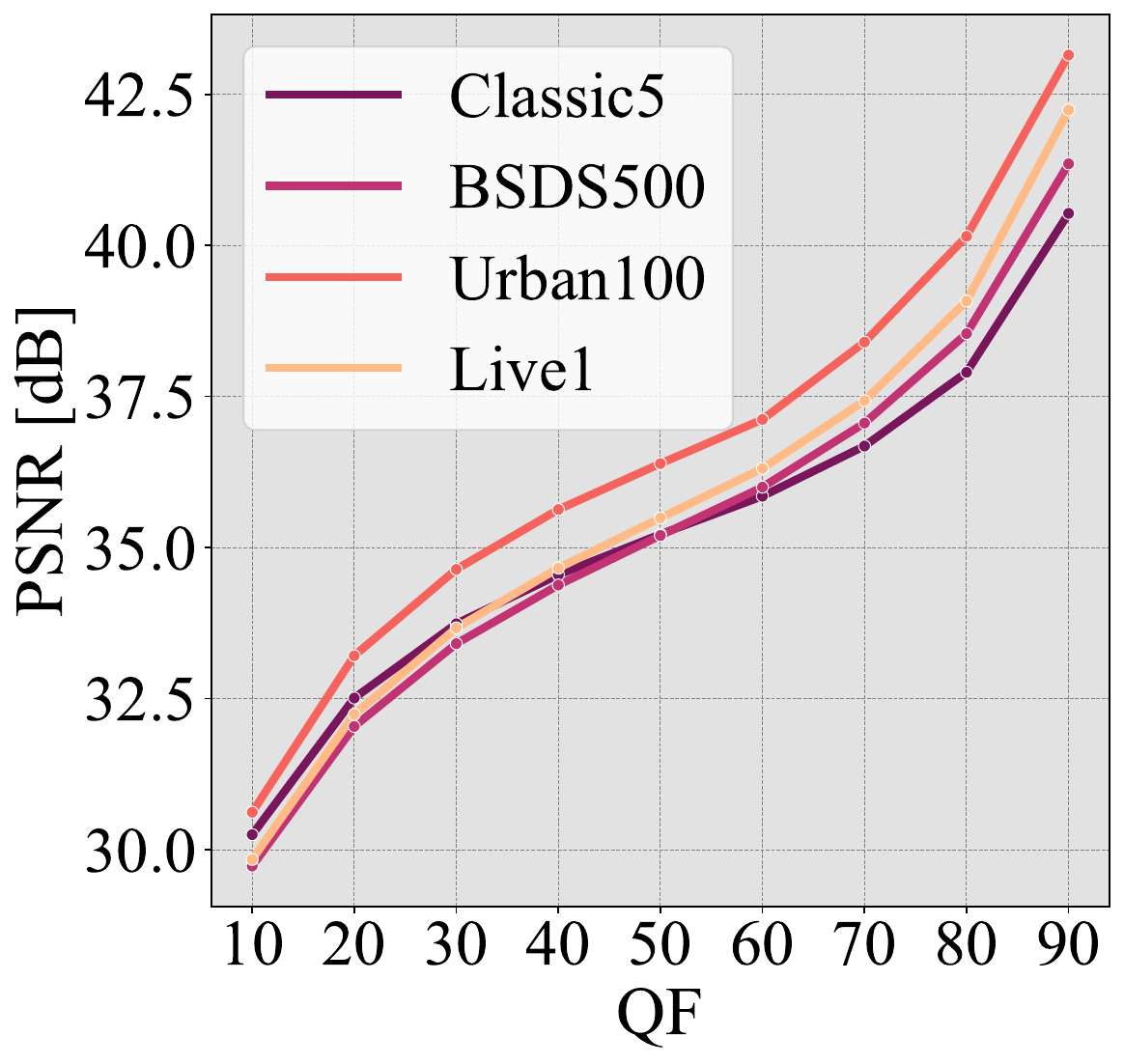}
        \caption{Grayscale image JPEG CAR}
        \label{fig:one_model3}
    \end{subfigure}
    \hfill
    \begin{subfigure}[b]{0.235\textwidth}
        \centering
        \includegraphics[width=\textwidth]{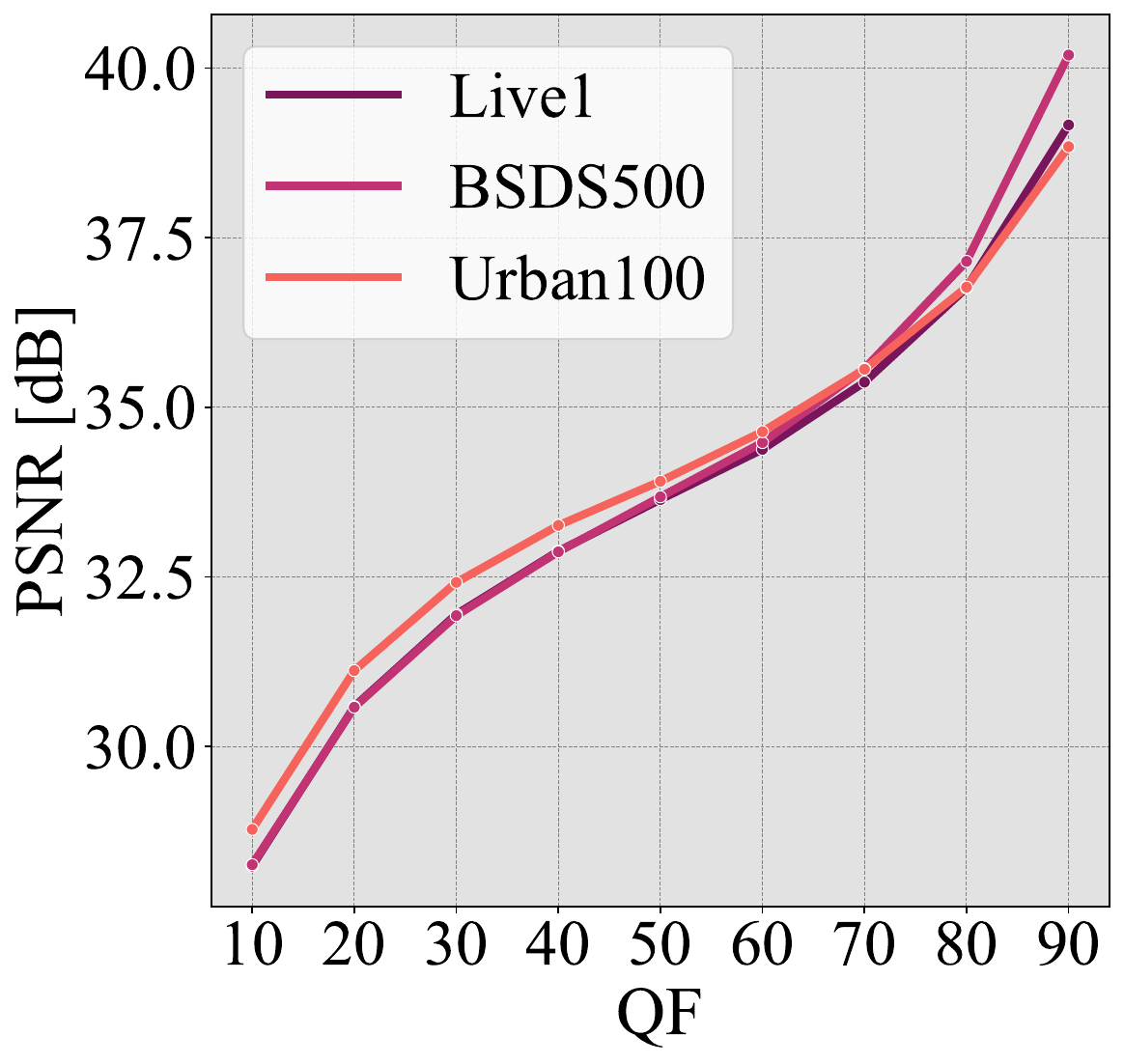}
        \caption{Color image JPEG CAR}
        \label{fig:one_model4}
    \end{subfigure}
    \vspace{-2mm}
    \caption{Training one model for multiple degradation levels of image denoising and JPEG compression artifact removal.}
    \label{fig:one_model}
\end{figure*}

\begin{table*}[!ht]
    \centering
    \setlength{\extrarowheight}{0.7pt}
    \setlength{\tabcolsep}{1.5pt}
    \begin{minipage}{0.6\linewidth} 
        \centering
        \caption{\textit{\textbf{Grayscale image JPEG compression artifact removal}} results. 
        \textcolor{magenta}{\textdagger}A single model is trained to handle multiple noise levels.} 
        \label{tab:jpeg_car_grapyscale}
        \vspace{-3mm}

        \scalebox{0.65}{ 
            \begin{tabular}{c | c| c c | c c | c c || c c | c c | c c | c c    }
            \toprule[0.1em]
            \multirow{2}{*}{Set} & \multirow{2}{*}{QF} & \multicolumn{2}{c|}{JPEG}  
            & \multicolumn{2}{c|}{\makecell{\textcolor{magenta}{\textdagger}DRUNet~\cite{zhang2021plug}}} & \multicolumn{2}{c||}{\textcolor{magenta}{\textdagger}{\ourmethod} } 
            & \multicolumn{2}{c|}{\makecell{SwinIR~\cite{liang2021swinir}}} & \multicolumn{2}{c|}{\makecell{ART~\cite{zhang2022accurate}}} & \multicolumn{2}{c|}{CAT~\cite{chen2022cross}}  & \multicolumn{2}{c}{{\ourmethod} }  \\ \cline{3-16}
            & & PSNR$\uparrow$ & SSIM$\uparrow$ & PSNR$\uparrow$ & SSIM$\uparrow$ & PSNR$\uparrow$ & SSIM$\uparrow$ & PSNR$\uparrow$ & SSIM$\uparrow$ & PSNR$\uparrow$ & SSIM$\uparrow$ & PSNR$\uparrow$ & SSIM$\uparrow$ & PSNR$\uparrow$ & SSIM$\uparrow$ \\
            \midrule[0.1em]
            {\multirow{4}{*}{\rotatebox[origin=c]{90}{\makecell{Classic5}}}}
            	&10	&27.82	&0.7600					
                &\sotab{30.16}	&\sotab{0.8234}		&\sotaa{30.25}	&\sotaa{0.8236}		
             &\sotab{30.27}	&0.8249		&\sotab{30.27}	&\sotab{0.8258}		&30.26	&\sotaa{0.8250}					&\sotaa{30.38}	&\sotaa{0.8266}		\\											
            	  &20	&30.12	&0.8340					
                  &\sotab{32.39}	&\sotab{0.8734}		&\sotaa{32.51}	&\sotaa{0.8737}		
             &32.52	&0.8748		& -	&	-	&\sotab{32.57}	&\sotaa{0.8754}					&\sotaa{32.62}	&\sotab{0.8751}		\\											
            	&30	&31.48	&0.8670					
                &\sotab{33.59}	&\sotab{0.8949}		&\sotaa{33.74}	&\sotaa{0.8954}		
             &33.73	&0.8961		&33.74	&\sotaa{0.8964}		&\sotab{33.77}	&\sotaa{0.8964}					&\sotaa{33.80}	&\sotab{0.8962}		\\											
            	  &40	&32.43	&0.8850					
                  &\sotab{34.41}	&\sotab{0.9075}		&\sotaa{34.55}	&\sotaa{0.9078}		
             &34.52	&0.9082		&34.55	&\sotab{0.9086}		&{34.58}	&\sotaa{0.9087}					&\sotaa{34.61}	&{0.9082}		\\											
             \hline									
            {\multirow{4}{*}{\rotatebox[origin=c]{90}{\makecell{LIVE1}}}}	
            	&10	&27.77	&0.7730					
                &\sotab{29.79}	&\sotab{0.8278}		&\sotaa{29.84}	&\sotaa{0.8328}		
             &29.86	&0.8287		&\sotab{29.89}	&\sotab{0.8300}		&\sotab{29.89}	&0.8295					&\sotaa{29.94}	&\sotaa{0.8359}		\\											
            	  &20	&30.07	&0.8510					
                  &\sotab{32.17}	&\sotab{0.8899}		&\sotaa{32.24}	&\sotaa{0.8926}		
             &32.25	&0.8909		&-	&	-	&\sotab{32.30}	&\sotab{0.8913}					&\sotaa{32.31}	&\sotaa{0.8938}		\\											
            	&30	&31.41	&0.8850					
                &\sotab{33.59}	&\sotab{0.9166}		&\sotaa{33.67}	&\sotaa{0.9192}		
             &33.69	&0.9174		&\sotab{33.71}	&\sotab{0.9178}		&\sotaa{33.73}	&0.9177					&\sotaa{33.73}	&\sotaa{0.9223}		\\											
            	  &40	&32.35	&0.9040					
                  &\sotab{34.58}	&\sotab{0.9312}		&\sotaa{34.66}	&\sotaa{0.9347}		
             &34.67	&0.9317		&34.70	&\sotab{0.9322}		&\sotaa{34.72}	&0.9320					&\sotab{34.71}	&\sotaa{0.9347}		\\											
            \hline											
            {\multirow{4}{*}{\rotatebox[origin=c]{90}{\makecell{Urban100}}}}	
            	&10	&26.33	&0.7816					
                &\sotab{30.31}	&\sotab{0.8745}		&\sotaa{30.62}	&\sotaa{0.8808}		
             &30.55	&0.8835		&\sotab{30.87}	&\sotab{0.8894}		&30.81	&0.8866					&\sotaa{31.07}	&\sotaa{0.8950}		\\											
            	  &20	&28.57	&0.8545					
                  &\sotab{32.81}	&\sotab{0.9241}		&\sotaa{33.21}	&\sotaa{0.9256}		
             &33.12	&0.9190		&	-&-		&33.38	&\sotaa{0.9269}					&\sotaa{33.51}	&\sotab{0.9250}		\\											
            	&30	&30.00	&0.9013					
                &\sotab{34.23}	&\sotab{0.9414}		&\sotaa{34.64}	&\sotaa{0.9478}		
             &34.58	&0.9417		&\sotab{34.81}	&0.9442		&\sotab{34.81}	&\sotab{0.9449}					&\sotaa{34.86}	&\sotaa{0.9459}		\\											
            	  &40	&31.06	&0.9215					
                  &\sotab{35.20}	&\sotab{0.9547}		&\sotaa{35.63}	&\sotaa{0.9566}		
             &35.50	&0.9515		&\sotab{35.73}	&\sotab{0.9553}		&\sotab{35.73}	&0.9511					&\sotaa{35.77}	&\sotaa{0.9561}		\\											
             \bottomrule[0.1em]
            \end{tabular}
        }
    \end{minipage}%
    \hfill
    \begin{minipage}{0.35\linewidth} 
        \centering
        \setlength{\extrarowheight}{3pt}
        \caption{\textit{\textbf{Image demosaicking}} results.}
        \vspace{-3mm}
        \label{tab:demosaicking}
        \scalebox{0.65}{ 
            \begin{tabular}{l | c c c c  c c c c }
                \toprule
                Datasets	&Matlab	
                &\makecell{RLDD \\ \cite{guo2020residual}}	 &\makecell{DRUNet \\ \cite{zhang2021plug}} &\makecell{RNAN \\ \cite{zhang2019residual}}		&\makecell{GRL-S \\ \cite{li2023efficient}} &Ours	\\ \hline
                
                Kodak	&35.78	
                &42.49	&42.68	&43.16	&\sotab{43.57}	&\sotaa{43.69}	\\
                McMaster	&34.43	
                &39.25	&39.39	&39.70	&\sotab{40.22}	&\sotaa{40.78}	\\
                \bottomrule[0.1em]
            \end{tabular}
        }
        
        \vspace{7mm} 
        \setlength{\extrarowheight}{7.5pt}
        \caption{\textit{\textbf{IR in AWC}} results.}
        \vspace{-3mm}
        \label{tab:weather}
        \scalebox{0.65}{ 
            \begin{tabular}{c|cccc}
                \toprule[0.1em]
                Dataset & \makecell{All-in-One \\ \cite{li2020all}} & \makecell{TransWeather \\ \cite{valanarasu2022transweather}} & \makecell{SemanIR \\ \cite{ren2024sharing}} & Ours \\ \midrule
                {RainDrop}~\citep{qian2018attentive}  
                & \sotaa{31.12} & 28.84 &30.82 & \sotab{30.84} \\  
                
                {Test1 (rain+fog)~\citep{li2020all}} 
                & 24.71 & {27.96} & \sotab{29.57}& \sotaa{30.93} \\
                
                {SnowTest100k-L~\citep{liu2018desnownet}} 
                & 28.33 & {28.48} & \sotab{30.76} & \sotaa{30.85} \\   
                \bottomrule[0.1em]
            \end{tabular}
        }
    \end{minipage}
\end{table*}

\noindent\textbf{Image JPEG CAR.} 
For JPEG CAR, the experiments are conducted for color and grayscale images with four quality factors (QF) (\ie 10, 20, 30, and 40).
The results for grayscale and color images are shown in Tab.~\ref{tab:jpeg_car_grapyscale} and the Tab.~E of Appx.~5. 
The quantitative results validate that the proposed {\ourmethod} outperforms most of the other comparison methods (Refer to the visual results in Fig.~\ref{fig:visual_jpeg_live1} on LIVE1 dataset). More visual comparisons are provided in the Appendix
to further support the effectiveness of {\ourmethod}.

We also trained the model to handle multiple degradation levels. The results in Tab.~\ref{tab:jpeg_car_grapyscale} show that {\ourmethod} outperforms DRUNet for grayscale image JPEG CAR by a large margin under this setting. Fig.~\ref{fig:one_model} shows the results for denoising in noise range [15, 75] and JPEG CAR in QF range [10, 90]. Despite that only one model is trained,
the model performs well on different degradation levels.

\begin{figure*}[!t]
\begin{center}
\scriptsize
\begin{tabular}[b]{c@{ } c@{ } c@{ } c@{ } c@{ } c@{ }}
    
\hspace{-1mm}  
    \includegraphics[width=0.16\linewidth,valign=t]{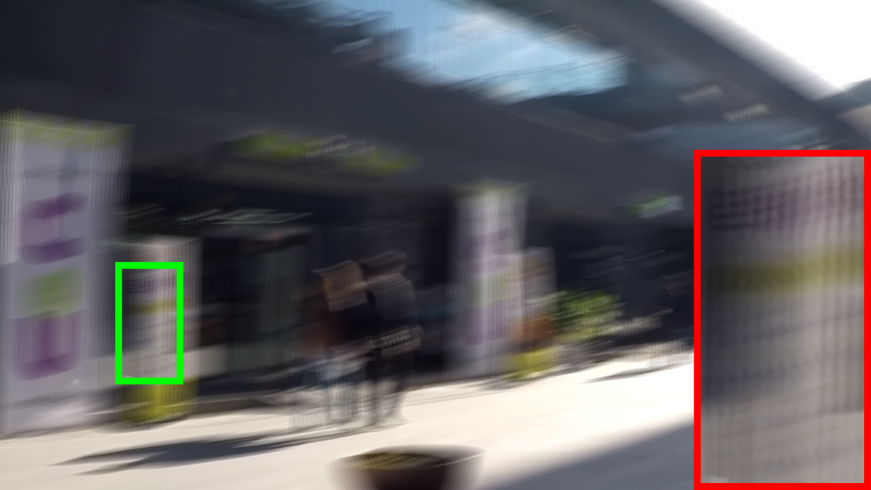} &
    \includegraphics[width=0.16\linewidth,valign=t]{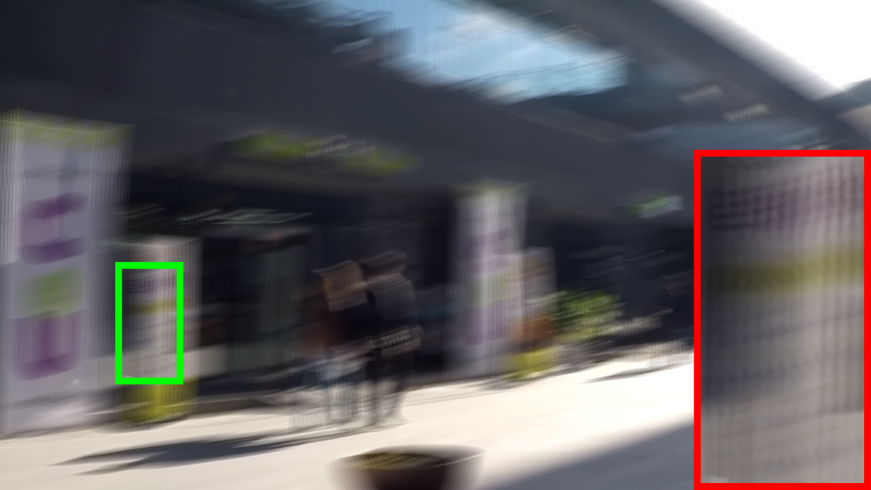} &    \includegraphics[width=0.16\linewidth,valign=t]{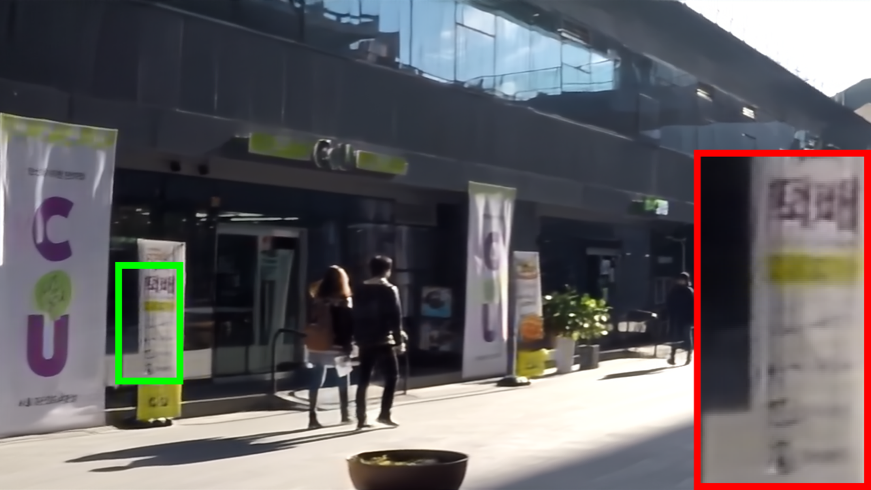} &
    \includegraphics[width=0.16\linewidth,valign=t]{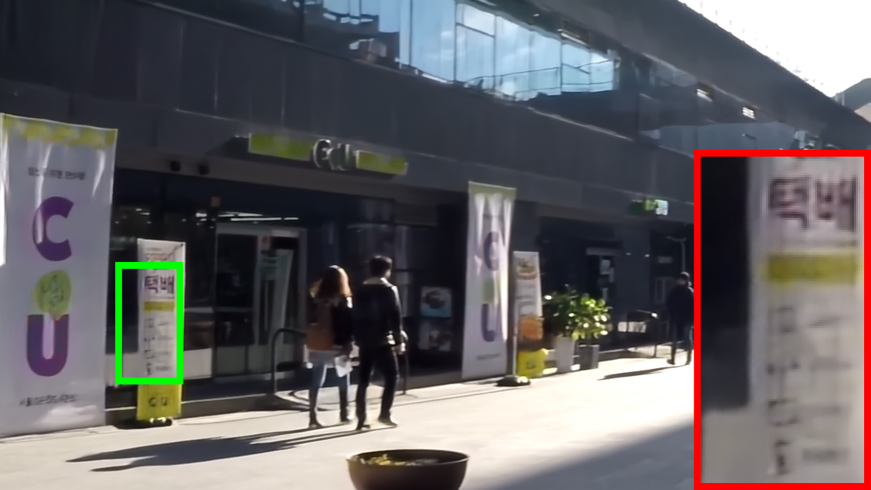} &
    \includegraphics[width=0.16\linewidth,valign=t]{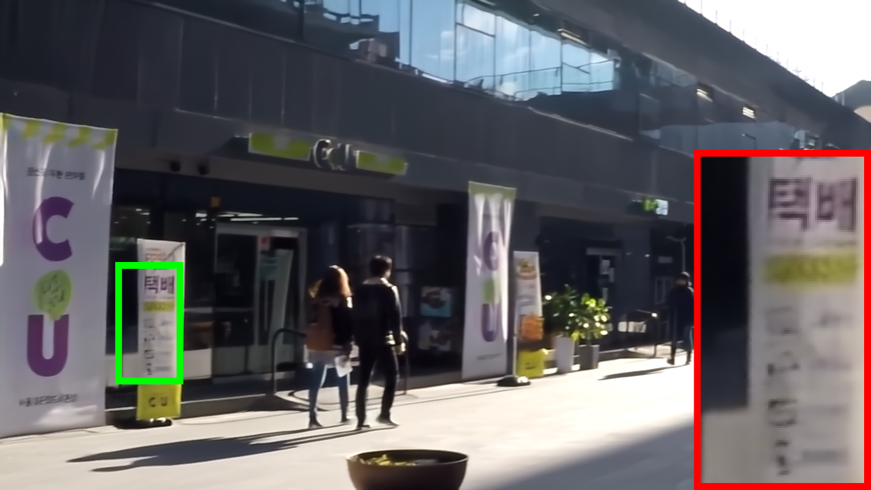} &
    \includegraphics[width=0.16\linewidth,valign=t]{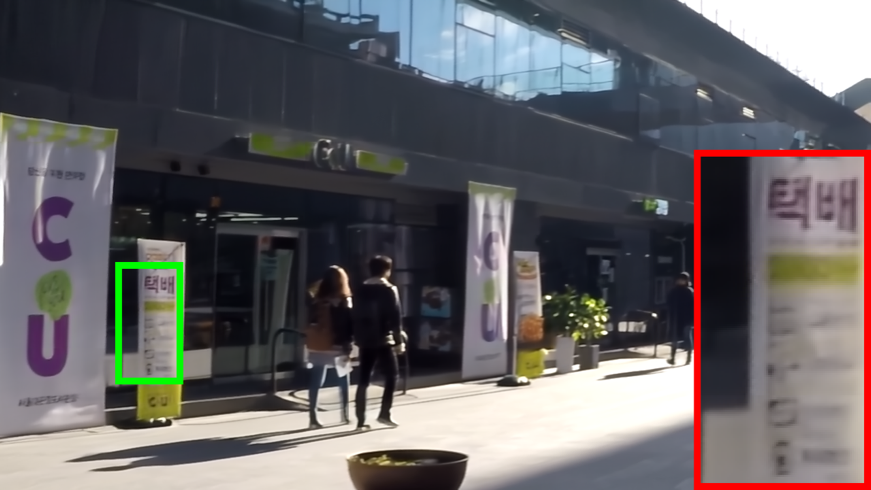} \\ 
    \revise{Blurred} & MPRNet~\citep{zamir2021multi} & \makecell{Uformer~\citep{wang2022uformer}} &\makecell{Restormer~\citep{zamir2022restormer}} & \makecell{GRL  \citep{li2023efficient}
    } & {\ourmethod} (Ours) \\
\end{tabular}
\end{center}
\vspace{-6mm}
\caption{Visual results for single image motion deblurring on GoPro dataset.}
\label{fig:visual_db_gopro}
\vspace{-4mm}
\end{figure*}
\noindent\textbf{Single-Image Motion Deblurring.} The results regarding the single-image motion deblurring are shown in Tab.~\ref{tab:motion_deblurring}. Compared with previous stat-of-the-art GRL~\citep{li2023efficient}, the proposed {\ourmethod} achieves the best results on the GoPro dataset (See visual results in Fig.~\ref{fig:visual_db_gopro}) and the second-best results on HIDE datasets. More
The results on RealBlur~\cite{rim2020real} dataset and more visual results are shown in the Appendix.

 \begin{figure}[!t]
    \small
    \centering
    \begin{minipage}{0.31\columnwidth}
        \centering
        \includegraphics[width=\columnwidth]{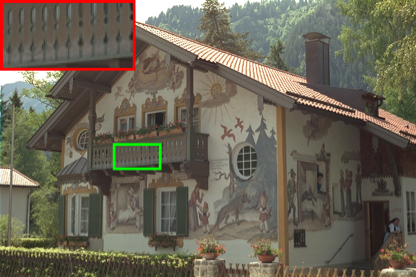}
        \vspace{-7mm} 
        \caption*{\scriptsize GT}
    \end{minipage}
    \hspace{0.01\columnwidth}
    \begin{minipage}{0.31\columnwidth}
        \centering
        \includegraphics[width=\columnwidth]{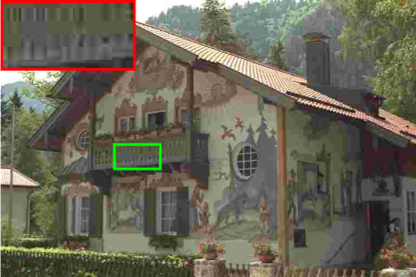}
        \vspace{-7mm} 
        \caption*{\scriptsize JPEG} 
    \end{minipage}
    \hspace{0.01\columnwidth}
    \begin{minipage}{0.31\columnwidth}
        \centering
        \includegraphics[width=\columnwidth]{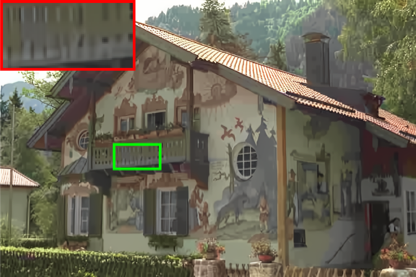}
        \vspace{-7mm} 
        \caption*{\scriptsize DRUNet~\citep{zhang2021plug}} 
    \end{minipage}

    \vspace{0.1cm} 

    \begin{minipage}{0.31\columnwidth}
        \centering
        \includegraphics[width=\columnwidth]{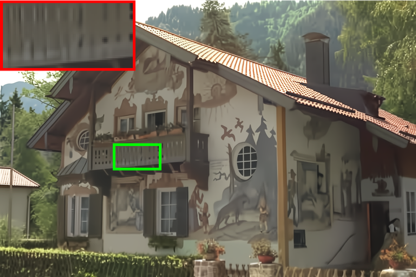}
        \vspace{-7mm} 
        \caption*{\scriptsize SwinIR~\cite{liang2021swinir}} 
    \end{minipage}
    \hspace{0.01\columnwidth}
    \begin{minipage}{0.31\columnwidth}
        \centering
        \includegraphics[width=\columnwidth]{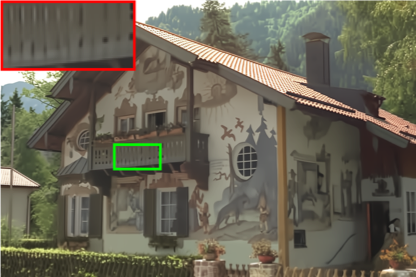}
        \vspace{-7mm} 
        \caption*{\scriptsize GRL~\cite{li2023efficient}} 
    \end{minipage}
    \hspace{0.01\columnwidth}
    \begin{minipage}{0.31\columnwidth}
        \centering
        \includegraphics[width=\columnwidth]{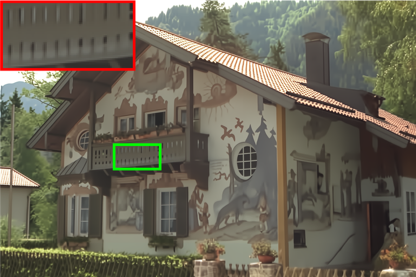}
        \vspace{-7mm}
        \caption*{\scriptsize {\ourmethod} (Ours)} 
    \end{minipage}
    \vspace{-2mm}
    \caption{Visual results for JPEG CAR on LIVE1 dataset} 
    \label{fig:visual_jpeg_live1}
    \vspace{-5mm}
\end{figure}

\noindent\textbf{Defocus Deblurring.} We also validate the effectiveness of our {\ourmethod} for dual-pixel defocus deblurring. The results in Tab.~\ref{tab:defocus_deblurring} show that {\ourmethod} outperforms the previous methods. 
Compared with Restormer on the combined scenes, our {\ourmethod} achieves a decent performance boost of 0.35 dB for dual-pixel defocus deblurring.

\noindent\textbf{Image Demosaicking.} We compare {\ourmethod} with the other methods
for demosaicking in Tab.~\ref{tab:demosaicking}. It shows that the proposed {\ourmethod} archives the best performance on both the Kodak and MaMaster test datasets, especially, \revise{0.12 dB} and \revise{0.56} dB absolute improvement compared to GRL.

\noindent\textbf{IR in AWC.} We validate {\ourmethod} in adverse weather conditions like rain, fog, and snow.
We compare {\ourmethod} with 
three methods in Tab.~\ref{tab:weather}. Our method achieves the best performance on Test1 (\ie \revise{4.6\%} improvement) and SnowTest100k-L (\ie \revise{0.09} dB improvement), while the second-best PSNR on RainDrop compared to all other methods. The visual result in the Appendix
shows that {\ourmethod} can remove more adverse weather artifacts.


\vspace{2mm}
\section{Conclusion}
\label{sec:conclusion}
In this paper, we introduced a fractal information flow principle for image restoration. Leveraging this concept, we devised a new model called {\ourmethod}, which progressively propagates information within local regions, facilitates information exchange in non-local ranges, and mitigates information isolation in the global context. We investigated how to scale up an IR model.
The effectiveness and generalizability of {\ourmethod} were validated through comprehensive experiments across various IR tasks.

{
    \small
    \bibliographystyle{ieeenat_fullname}
    \bibliography{main}
}

\clearpage
\clearpage

\maketitlesupplementary
\setcounter{section}{0}
\setcounter{figure}{0}    
\setcounter{table}{0}    

\renewcommand{\thetable}{\Alph{table}}
\renewcommand{\thefigure}{\Alph{figure}}



\section{Experimental Settings}
\label{sec:supp:training}
\subsection{Architecture Details}
\label{subsec:supp:architecture_details}
We choose two commonly used basic architectures for IR tasks including the U-shape hierarchical architecture and the columnar architecture.
The columnar architecture is used for image SR while the U-shape architecture is used for other IR tasks including image denoising, JPEG CAR, image deblurring, IR in adverse weather conditions, image deblurring, and image demosaicking. 
We included details on the structure of the {\ourmethod} in Tab.~\ref{table:model_details}. This table outlines the number of {\ourmethod} stages and the distribution of {\ourmethod} layers within each stage for a thorough understanding of our model's architecture.

\begin{figure}[!h]
    \centering
    \includegraphics[width=1\linewidth]{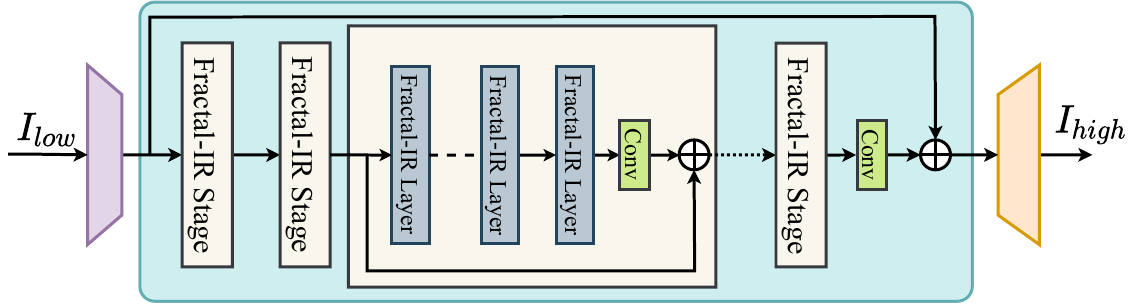}
    \vspace{-2mm}
    \caption{The columnar {\ourmethod} architecture.}
    \label{fig:architecture_columnar}
    \vspace{-3mm}
\end{figure}

\begin{table*}[!ht]
    \centering
    \small
    \vspace{-3mm}
    \caption{The details of the {\ourmethod} stages and {\ourmethod} layers per stage for both architectures.}
    \setlength{\extrarowheight}{1pt}
    \setlength\tabcolsep{3pt} 
    \label{table:model_details}
    \vspace{-3mm}
    \scalebox{0.86}{
    \begin{tabular}{c|ccc|cc}
    \toprule[0.1em]
    & \multicolumn{3}{c|}{U-shaped architecture}         & \multicolumn{2}{c}{Columnar architecture}                    \\ \hline
    & Down Stages & Upstages & Latent Stage & \multicolumn{1}{c}{{\ourmethod}-Base} & {\ourmethod}-Large \\ \hline
    Num. of {\ourmethod} Stages      & 3           & 3        & 1            & \multicolumn{1}{c}{6}           & 8            \\
    Num. of {\ourmethod} Layer/Stage & 6           & 6        & 6            & \multicolumn{1}{c}{6}           & 8            \\ \bottomrule[0.1em]
    \end{tabular}}
    \vspace{-4mm}
\end{table*}

\subsection{Training Details}
The proposed {\ourmethod} explores \textbf{7} different IR tasks, and the training settings vary slightly for each task. These differences encompass the architecture of the proposed {\ourmethod}, variations in training phases, choice of the optimizer, employed loss functions, warm-up settings, learning rate schedules, batch sizes, and patch sizes. We have provided a comprehensive overview of these details.

In addition, there are several points about the training details we want to make further explanation. 1) 
For image SR, the network is pre-trained on ImageNet~\citep{deng2009imagenet}.
This is inspired by previous works~\citep{dong2014learning,chen2021pre,li2021efficient,chen2023activating}. 
2) The optimizer used for IR in AWC is Adam~\citep{kingma2014adam}, while AdamW~\citep{loshchilov2018decoupled} is used for the rest IR tasks. 
3) The training losses for IR in AWC are the smooth L1 and the Perception VGG loss~\citep{johnson2016perceptual,simonyan2015very}. For image deblurring, the training loss is the Charbonnier loss. For the rest IR task, the L1 loss is commonly used during the training. 4) For IR in AWC, we adopted similar training settings as Transweather~\citep{valanarasu2022transweather}, the model is trained for a total of 750K iterations.


\subsection{Data and Evaluation}
The training dataset and test datasets for different IR tasks are described in this section.
For IR in AWC, we used a similar training pipeline as Transweather with only one phase. Additionally, for tasks such as image super-resolution (SR), JPEG CAR, image denoising, and demosaicking, how the corresponding low-quality images are generated is also briefly introduced below.

\noindent\textbf{Image SR.}
For image SR, the LR image is synthesized by \texttt{Matlab} bicubic downsampling function before the training. We investigated the upscalingg factors $\times2$, $\times3$, and $\times4$. 

\begin{itemize}[leftmargin=*]
    \item  The training datasets: DIV2K~\citep{agustsson2017ntire} and Flickr2K~\citep{lim2017enhanced}. 

    \item  The test datasets: 
    Set5~\citep{bevilacqua2012low}, Set14~\citep{zeyde2010single}, BSD100~\citep{martin2001database}, Urban100~\citep{huang2015single}, and Manga109~\citep{matsui2017sketch}.
\end{itemize}

\noindent\textbf{Image Denoising.}
For image denoising, we conduct experiments on both color and grayscale image denoising. During training and testing, noisy images are generated by adding independent additive white Gaussian noise (AWGN) to the original images. The noise levels are set to $\sigma = 15, 25, 50$. We train individual networks at different noise levels. The network takes the noisy images as input and tries to predict noise-free images. Additionally, we also tried to train one model for all noise levels.

\begin{itemize}[leftmargin=*]
    \item  The training datasets: DIV2K~\citep{agustsson2017ntire}, Flickr2K~\citep{lim2017enhanced}, WED~\citep{ma2016waterloo}, and BSD400~\citep{martin2001database}. 

    \item  The test datasets for color image: CBSD68~\citep{martin2001database}, Kodak24~\citep{franzen1999kodak}, McMaster~\citep{zhang2011color}, and Urban100~\citep{huang2015single}.
    
    \item  The test datasets for grayscale image: Set12~\citep{zhang2017beyond}, BSD68~\citep{martin2001database}, and Urban100~\citep{huang2015single}.
\end{itemize}

\noindent\textbf{JPEG compression artifact removal.}
For JPEG compression artifact removal, the JPEG image is compressed by the \texttt{cv2} JPEG compression function. The compression function is characterized by the quality factor. We investigated four compression quality factors including 10, 20, 30, and 40. The smaller the quality factor, the more the image is compressed, meaning a lower quality. We also trained one model to deal with different quality factors.

\begin{itemize}[leftmargin=*]
    \item  The training datasets: DIV2K~\citep{agustsson2017ntire}, Flickr2K~\citep{lim2017enhanced}, and WED~\citep{ma2016waterloo}. 

    \item  The test datasets: Classic5~\citep{foi2007Classic5}, LIVE1~\citep{sheikh2005live}, Urban100~\citep{huang2015single}, BSD500~\citep{arbelaez2010contour}. 
\end{itemize}

\begin{table*}[!htb]
    \scriptsize
    \setlength{\extrarowheight}{0.7pt}
    \setlength{\tabcolsep}{4pt}
    \centering
    \caption{Space and time complexity of classical attention mechanisms.}
    \label{tab:space_time_complexity}
    \scalebox{1.0}{
            \begin{tabular}{c|ccc}
                \toprule[0.1em]
            \textbf{Attn. method} &\textbf{Time complexity} &\textbf{Space complexity} & \makecell{\textbf{Max receptive field of} \\ \textbf{two transformer layers}} \\ \hline
            \myrowcolour Global Attn.                  & $\mathcal{O}\left((4+2\gamma)BHWC^2 + {2}B(HW)^2C\right)$               & $\mathcal{O}\left(4BHWC + B(HW)^2h\right)$                & $H \times W$      \\
            Window Attn. ($p \times p$)   & $\mathcal{O}\left((4+2\gamma)BHWC^2 + {2}BHWp^2C\right)$                & $\mathcal{O}\left(4BHWC + BHWhp^2\right)$                 & $2p \times 2p$    \\
            \myrowcolour Window Attn. ($8P \times 8P$) & $\mathcal{O}\left((4+2\gamma)BHWC^2 + {128}BHWp^2s^2C\right)$           & $\mathcal{O}\left(4BHWC + 64BHWhp^2s^2\right)$            & $16P \times 16P$  \\
            The proposed                  & $\mathcal{O}\left((5+2\gamma)BHWC^2 + \frac{3}{2}BHW(p^2+s^2)C\right)$  & $\mathcal{O}\left(3BHWC + BHWh\max{(p^2, s^2)}\right)$    & $16P \times 16P$  \\
                \bottomrule[0.1em]
                \end{tabular}
        
        }
\end{table*}

\noindent\textbf{IR in Adverse Weather Conditions.} 
For IR in adverse weather conditions, the model is trained on a combination of images degraded by a variety of adverse weather conditions. The same training and test dataset is used as in Transweather~\citep{valanarasu2022transweather}. The training data comprises 9,000 images sampled from Snow100K \citep{liu2018desnownet}, 1,069 images from Raindrop \citep{qian2018attentive}, and 9,000 images from Outdoor-Rain \citep{li2019heavy}. Snow100K includes synthetic images degraded by snow, Raindrop consists of real raindrop images, and Outdoor-Rain contains synthetic images degraded by both fog and rain streaks. The proposed method is tested on both synthetic and real-world datasets.

\begin{itemize}[leftmargin=*]
    \item  The test datasets: test1 dataset~\citep{li2020all, li2019heavy}, the RainDrop test dataset~\citep{qian2018attentive}, and the Snow100k-L test. 
\end{itemize}

\noindent\textbf{Image Deblurring.}
For single-image motion deblurring, 

\begin{itemize}[leftmargin=*]
    \item  The training datasets: GoPro~\citep{nah2017deep} dataset. 

    \item  The test datasets: 
    GoPro~\citep{nah2017deep}, HIDE~\citep{shen2019human}, RealBlur-R~\citep{rim2020real}, and RealBlur-J~\citep{rim2020real} datasets.
\end{itemize}

\noindent\textbf{Defocus Deblurring.}
The task contains two modes including single-image defocus deblurring and dual-pixel defocus deblurring. For single-image defocus deblurring, only the blurred central-view image is available. For dual-pixel defocus deblurring, both the blurred left-view and right-view images are available. The dual-pixel images could provide additional information for defocus deblurring and thus could lead to better results. PSNR, SSIM, and mean absolute error (MAE) on the RGB channels are reported. Additionally, the image perceptual quality score LPIPS is also reported.

\begin{itemize}[leftmargin=*]
    \item  The training datasets: DPDD~\citep{abuolaim2020defocus} training dataset. The training subset contains 350 scenes.  

    \item  The test datasets: 
    DPDD~\citep{abuolaim2020defocus} test dataset. The test set contains 37 indoor scenes and 39 outdoor scenes
\end{itemize}

\noindent\textbf{Image Demosaicking.}
For image demosaicking, the mosaic image is generated by applying a Bayer filter on the ground-truth image. Then the network try to restore high-quality image. The mosaic image is first processed by the default \texttt{Matlab} demosaic function and then passed to the network as input. 

\begin{itemize}[leftmargin=*]
    \item  The training datasets: DIV2K~\citep{agustsson2017ntire} and Flickr2K~\citep{lim2017enhanced}. 

    \item  The test datasets: Kodak~\citep{franzen1999kodak}, McMaster~\citep{zhang2011color}. 
\end{itemize}

\begin{figure*}[!th]
    \centering
    \includegraphics[width=1\linewidth]{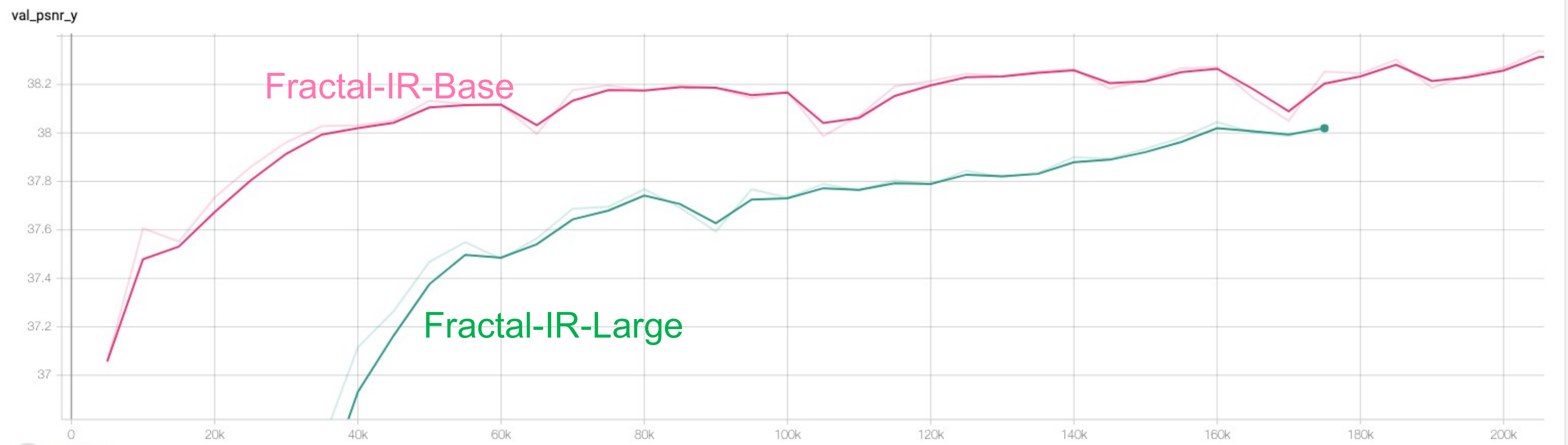}
    \caption{When the SR model is scale-up from {\ourmethod}-B to {\ourmethod}-L, the model {\ourmethod}-L converges slower than {\ourmethod}-B.}
    \label{fig:convergence}
\end{figure*}

\section{Space and Time Complexity}
\label{sec:space_time_complexity}

We compare the space and time complexity and the effective receptive field of the proposed method with a couple of other self-attention methods including global attention and window attention. Suppose the input feature has the dimension $B \times C \times H \times W$, the window size of window attention is $p$, the number of attention heads is $h$, larger patch size of the proposed L2 information flow is $P=s \times p$, the expansion ratio of the MLP in transformer layer is $\gamma$. For time complexity, both self-attention and the feed-forward network are considered. For space complexity, we consider the tensors that have to appear in the memory at the same time, which include the input tensor, the query tensor, the key tensor, the value tensor, and the attention map.

The time complexity of the proposed transformer layer is
\begin{align}
    \mathcal{O}\left((5+2\gamma)BHWC^2  + \frac{3}{2}BHWp^2C \right. \\ \nonumber
    \left. +\frac{3}{2}BHWs^2C   + 9\gamma BHWC\right).
\end{align}
The last term is very small compared with the former two terms and can be omitted. Thus, the time complexity is simplified as \begin{equation}
    \mathcal{O}\left((5+2\gamma)BHWC^2 + \frac{3}{2}BHWp^2C+\frac{3}{2}BHWs^2C\right).
\end{equation}

The space complexity of the proposed transformer layer is
\begin{equation}
    \mathcal{O}\left(3BHWC + BHWh\max{(p^2, s^2)}\right).
\end{equation} 
The maximum receptive field of two consecutive transformer layers is $16P$.

In Tab.~\ref{tab:space_time_complexity}, we list the space and time complexity, and maximum receptive field of global attention, window attention, and the proposed method. As shown in this table, window attention is much more efficient than global attention but with the cost of reduced receptive field. The proposed fractal information flow mechanism is more efficient than window attention in propagating information to the global range. As shown in the third row, to achieve the same receptive field as the proposed method, the space and time complexity of window attention is much higher than that of the proposed method.

\section{Model Scaling-up}
\label{sec:supp:model_scaling_up}

\begin{figure*}[!htb]
    \centering
    \includegraphics[width=0.99\linewidth]{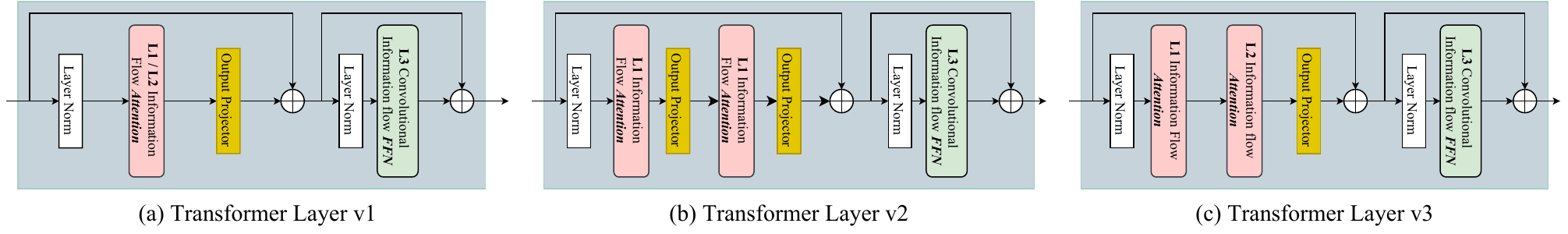}
    \vspace{-2mm}
    \caption{Comparison of three types of transformer layers designed in this paper.}
    \label{fig:attention_version}
    \vspace{-3mm}
\end{figure*}

\begin{table}[!ht]
\caption{Model scaling-up exploration with SR.}
\label{tab:scaling_up_convergence_full}
\setlength{\extrarowheight}{0.7pt}
\setlength{\tabcolsep}{2pt}
\centering
\scalebox{0.65}{
    \begin{tabular}{c|c|cc|ccccc}
    \toprule[0.1em]
    \multirow{2}{*}{\textbf{Scale}}  &  \multirow{2}{*}{\makecell{\textbf{Model} \\ \textbf{Size}}} & \multirow{2}{*}{\makecell{\textbf{Warm} \\ \textbf{up}}} & \multirow{2}{*}{\makecell{\textbf{Conv} \\ \textbf{Type}}}  &  \multicolumn{5}{c}{\textbf{PSNR}} 
    \\\cline{5-9}
    &   & & & \textbf{Set5}& \textbf{Set14} & \textbf{BSD100} & \textbf{Urban100} & \textbf{Manga109}
    \\ \hline
    \myrowcolourpink $2\times$	
     &15.69	&No	&\texttt{conv1}	&38.52	&34.47	&32.56	&34.17	&39.77	\\
    \myrowcolourpink $2\times$	
    &57.60	&No	&\texttt{conv1}	&38.33	&34.17	&32.46	&33.60	&39.37	\\ 
    $2\times$	
    &57.60	&Yes	&\texttt{conv1}	&38.41	&34.33	&32.50	&33.80	&39.51	\\
    $2\times$	
    &54.23	&Yes	&\texttt{linear}	&\sotab{38.56}	&\sotaa{34.59}	&\sotaa{32.58}	&\sotab{34.32}	&\sotab{39.87}	\\
    $2\times$	
    &55.73	&Yes	&\texttt{conv3}	&\sotaa{38.65}	&\sotab{34.48}	&\sotaa{32.58}	&\sotaa{34.33}	&\sotaa{40.12}	\\ \midrule
    \myrowcolourpink $3\times$	
    &15.87	&No	&\texttt{conv1}	&35.06	&30.91	&29.48	&30.02	&34.41	\\
    \myrowcolourpink $3\times$	
    &57.78	&No	&\texttt{conv1}	&34.70	&30.62	&29.33	&29.11	&33.96	\\ 
    $3\times$	
    &57.78	&Yes	&\texttt{conv1}	&34.91	&30.77	&29.39	&29.53	&34.12	\\
    $3\times$	
    &54.41	&Yes	&\texttt{linear}	&\sotab{35.13}	&\sotaa{31.04}	&\sotaa{29.52}	&\sotab{30.20}	&\sotab{34.54}	\\
    $3\times$	
    &55.91	&Yes	&\texttt{conv3}	&\sotaa{35.14}	&\sotab{31.03}	&\sotab{29.51}	&\sotaa{30.22}	&\sotaa{34.76}	\\ 
    \midrule
    \myrowcolourpink $4\times$	&15.84	&No	&\texttt{conv1}	&33.00	&29.11	&27.94	&27.67	&31.41	\\
    \myrowcolourpink $4\times$	&57.74	&No	&\texttt{conv1}	&33.08	&29.19	&27.97	&27.83	&31.56	\\ 
    $4\times$	&57.74	&Yes	&\texttt{conv1}	&32.67	&28.93	&27.83	&27.11	&30.97	\\
    $4\times$	&54.37	&Yes	&\texttt{linear}	&\sotaa{33.06}	&\sotaa{29.16}	&\sotaa{27.99}	&\sotaa{27.93}	&\sotaa{31.66}	\\
    $4\times$	&55.88	&Yes	&\texttt{conv3}	&\sotaa{33.06}	&\sotaa{29.16}	&\sotab{27.97}	&\sotab{27.87}	&\sotab{31.54}	\\
    \bottomrule[0.1em]
    \end{tabular}
    }
\end{table}

As mentioned in the main paper, when the initially designed SR model is scaled up from about 10M parameters to about 50M parameters, the performance of the large SR model becomes worse. The effect is shown in Fig.~\ref{fig:convergence}. 
The PSNR curve on the Set5 dataset for the first 200k iterations is shown in this figure. The scale-up model {\ourmethod}-L converges slower than the smaller model {\ourmethod}-B.
The same phenomenon could be observed by comparing the first two rows for each upscaling factor in Tab.~\ref{tab:scaling_up_convergence_full}, where scaled-up models converge to worse local minima. A similar problem occurs in previous works~\citep{chen2023activating,lim2017enhanced}.

\begin{table*}[!ht]
    \scriptsize
    \setlength{\extrarowheight}{0.7pt}
    \setlength{\tabcolsep}{2pt}
    \setlength{\abovecaptionskip}{0cm}
    \begin{center}
    \caption{\textbf{\textit{Classical image SR}} results. Top-2 results are highlighted in  \textcolor{red}{red} and  \textcolor{blue}{blue}.}%
    \label{tab:sr_results_full}
    \scalebox{0.9}{
        \begin{tabular}{l|c|r|cc|cc|cc|cc|cc}
        \toprule[0.1em]
        \multirow{2}{*}{\textbf{Method}} & \multirow{2}{*}{\textbf{Scale}} & {\textbf{Params}} &  \multicolumn{2}{c|}{\textbf{Set5}} &  \multicolumn{2}{c|}{\textbf{Set14}} &  \multicolumn{2}{c|}{\textbf{BSD100}} &  \multicolumn{2}{c|}{\textbf{Urban100}} &  \multicolumn{2}{c}{\textbf{Manga109}} 
        \\
        \cline{4-13}
        &  & \multicolumn{1}{c|}{\textbf{[M]}} & PSNR$\uparrow$ & SSIM$\uparrow$ & PSNR$\uparrow$ & SSIM$\uparrow$ & PSNR$\uparrow$ & SSIM$\uparrow$ & PSNR$\uparrow$ & SSIM$\uparrow$ & PSNR$\uparrow$ & SSIM$\uparrow$
        \\
        \midrule[0.1em]
        \myrowcolour EDSR~\citep{lim2017enhanced} & $2\times$ & 40.73 & 38.11 & 0.9602 & 33.92 & 0.9195 & 32.32 & 0.9013 & 32.93 & 0.9351 & 39.10 & 0.9773\\
        SRFBN~\citep{li2019feedback} & $2\times$ & 2.14 & 38.11 & 0.9609 & 33.82 & 0.9196 & 32.29 & 0.9010 & 32.62 & 0.9328 & 39.08 & 0.9779\\
        \myrowcolour RCAN~\citep{zhang2018rcan}&	$2\times$&	15.44&	38.27&	0.9614&	34.12&	0.9216&	32.41&	0.9027&	33.34&	0.9384&	39.44&	0.9786\\
        SAN~\citep{dai2019SAN}&	$2\times$&	15.71&	38.31&	0.9620&	34.07&	0.9213&	32.42&	0.9028&	33.10&	0.9370&	39.32&	0.9792\\

        \myrowcolour HAN~\citep{niu2020HAN}&	$2\times$&	63.61&	38.27&	0.9614&	34.16&	0.9217&	32.41&	0.9027&	33.35&	0.9385&	39.46&	0.9785\\
        
            
        NLSA~\citep{mei2021NLSA}&	$2\times$&	42.63&	38.34&	0.9618&	34.08&	0.9231&	32.43&	0.9027&	33.42&	0.9394&	39.59&	0.9789\\
        \myrowcolour IPT~\citep{chen2021pre}&	$2\times$&	115.48&	38.37&	-&	34.43&	-&	32.48&	-&	33.76&	-&	-&	-\\ \hline
        
        SwinIR~\citep{liang2021swinir}&	$2\times$&	11.75&	38.42&	0.9623&	34.46&	0.9250&	32.53&	0.9041&	33.81&	0.9427&	39.92&	0.9797\\  
        \myrowcolour CAT-A~\citep{chen2022cross}    & $2\times$ &16.46 & 38.51 & 0.9626 & 34.78 & 0.9265 & 32.59 & 0.9047 & 34.26 & 0.9440 & 40.10 & 0.9805 \\
        ART~\citep{zhang2022accurate}	&$2\times$ &16.40	&38.56	&0.9629	&34.59	&0.9267	&32.58	&0.9048	&34.3	&0.9452	&40.24	&0.9808	\\		
        \myrowcolour EDT~\citep{li2021efficient}&	$2\times$&	11.48&	 {38.63}&	 {0.9632}&	 {34.80}&	0.9273&	 {32.62}&	0.9052&	34.27&	0.9456&	 {40.37}&	 {0.9811}\\ 
        GRL-B~\citep{li2023efficient} &	$2\times$&	20.05&	 {38.67}&	 {0.9647}&	 {35.08}&	 \sotab{0.9303}&	 {32.67}&	 \sotab{0.9087}&	 {35.06}&	 \sotaa{0.9505}&	 {40.67}&	 {0.9818}\\ 
        \myrowcolour HAT~\citep{chen2023activating}	&$2\times$	&20.62	&38.73	&0.9637	&35.13	&0.9282	&32.69	&0.9060	&34.81	&\sotab{0.9489}	&40.71	&0.9819	\\
        {\ourmethod}-B (Ours)	&$2\times$& 14.68	&38.71	&\sotab{0.9657}	&35.16	&0.9299	&32.73	&\sotab{0.9087}	&34.94	&0.9484	&40.81	&0.9830			\\
        \myrowcolour HAT-L~\citep{chen2023activating}	&$2\times$	&40.70	&\sotaa{38.91}	&0.9646	&\sotaa{35.29}	&0.9293	&\sotab{32.74}	&0.9066	&\sotab{35.09}	&\sotaa{0.9505}	&\sotab{41.01}	&\sotab{0.9831}	\\
        {\ourmethod}-L (Ours)	&$2\times$& 39.07	&\sotab{38.87}	&\sotaa{0.9663}	&\sotab{35.27}	&\sotaa{0.9311}	&\sotaa{32.77}	&\sotaa{0.9092}	&\sotaa{35.16}	&\sotaa{0.9505}	&\sotaa{41.22}	&\sotaa{0.9846}			\\
        \midrule[0.1em]
        
        \myrowcolour EDSR~\citep{lim2017enhanced} & $3\times$ & 43.68 & 34.65 & 0.9280 & 30.52 & 0.8462 & 29.25 & 0.8093 & 28.80 & 0.8653 & 34.17 & 0.9476\\
        SRFBN~\citep{li2019feedback} & $3\times$ &2.83 & 34.70 & 0.9292 & 30.51 & 0.8461 & 29.24 & 0.8084 & 28.73 & 0.8641 & 34.18 & 0.9481\\    
        \myrowcolour RCAN~\citep{zhang2018rcan}&	$3\times$&	15.63&	34.74&	0.9299&	30.65&	0.8482&	29.32&	0.8111&	29.09&	0.8702&	34.44&	0.9499\\
        SAN~\citep{dai2019SAN}&	$3\times$&	15.90&	34.75&	0.9300&	30.59&	0.8476&	29.33&	0.8112&	28.93&	0.8671&	34.30&	0.9494\\

        \myrowcolour HAN~\citep{niu2020HAN}&	$3\times$&	64.35&	34.75&	0.9299&	30.67&	0.8483&	29.32&	0.8110&	29.10&	0.8705&	34.48&	0.9500\\
        
            
        NLSA~\citep{mei2021NLSA}&	$3\times$&	45.58&	34.85&	0.9306&	30.70&	0.8485&	29.34&	0.8117&	29.25&	0.8726&	34.57&	0.9508\\
        \myrowcolour IPT~\citep{chen2021pre}&	$3\times$&	115.67&	34.81&	-&	30.85&	-&	29.38&	-&	29.49&	-&	-&	-\\ \hline
        
        SwinIR~\citep{liang2021swinir}&	$3\times$&	11.94&	34.97&	0.9318&	30.93&	0.8534&	29.46&	0.8145&	29.75&	0.8826&	35.12&	0.9537\\  
        \myrowcolour CAT-A~\citep{chen2022cross}      & $3\times$ &16.64 & 35.06 & 0.9326 & 31.04 & 0.8538 & 29.52 & 0.8160 & 30.12 & 0.8862 & 35.38 & 0.9546 \\
        ART~\citep{zhang2022accurate}		&$3\times$ &16.58	&35.07	&0.9325	&31.02	&0.8541	&29.51	&0.8159	&30.1	&0.8871	&35.39	&0.9548	\\		
        \myrowcolour EDT~\citep{li2021efficient}&	$3\times$&	11.66&	35.13&	0.9328&	31.09&	0.8553&	29.53&	0.8165&	30.07&	0.8863&	35.47&	0.9550\\ 
        GRL-B~\citep{li2023efficient} &	$3\times$&	20.24&	35.12&	0.9353&	31.27&	\sotab{0.8611}&	29.56&	0.8235&	30.92&	\sotab{0.8990}&	35.76&	0.9566\\ 
        \myrowcolour HAT~\citep{chen2023activating}	&$3\times$	&20.81	&35.16	&0.9335	&31.33	&0.8576	&29.59	&0.8177	&30.7	&0.8949	&35.84	&0.9567	\\
        {\ourmethod}-B (Ours)	&$3\times$& 14.87	&35.11	&\sotab{0.9372}	&31.37	&0.8598	&29.60	&\sotab{0.8240}	&30.79	&0.8977	&35.92	&\sotab{0.9583}			\\
        \myrowcolour HAT-L~\citep{chen2023activating}	&$3\times$	&40.88	&\sotaa{35.28}	&0.9345	&\sotab{31.47}	&0.8584	&\sotab{29.63}	&0.8191	&\sotab{30.92}	&0.8981	&\sotab{36.02}	&0.9576	\\
        {\ourmethod}-L (Ours)	&$3\times$& 39.26	&\sotab{35.20}	&\sotaa{0.9380}	&\sotaa{31.55}	&\sotaa{0.8616}	&\sotaa{29.67}	&\sotaa{0.8256}	&\sotaa{31.07}	&\sotaa{0.9020}	&\sotaa{36.12}	&\sotaa{0.9588}			\\ \midrule[0.1em]
        
        \myrowcolour EDSR~\citep{lim2017enhanced} & $4\times$ & 43.09 & 32.46 & 0.8968 & 28.80 & 0.7876 & 27.71 & 0.7420 & 26.64 & 0.8033 & 31.02 & 0.9148\\
        SRFBN~\citep{li2019feedback} & $4\times$ &3.63 & 32.47 & 0.8983 & 28.81 & 0.7868 & 27.72 & 0.7409 & 26.60 & 0.8015 & 31.15 & 0.9160\\

        \myrowcolour RCAN~\citep{zhang2018rcan}&	$4\times$&	15.59&	32.63&	0.9002&	28.87&	0.7889&	27.77&	0.7436&	26.82&	0.8087&	31.22&	0.9173\\
        SAN~\citep{dai2019SAN}&	$4\times$&	15.86&	32.64&	0.9003&	28.92&	0.7888&	27.78&	0.7436&	26.79&	0.8068&	31.18&	0.9169\\

        \myrowcolour HAN~\citep{niu2020HAN}&	$4\times$&	64.20&	32.64&	0.9002&	28.90&	0.7890&	27.80&	0.7442&	26.85&	0.8094&	31.42&	0.9177\\
        
            
        NLSA~\citep{mei2021NLSA}&	$4\times$&	44.99&	32.59&	0.9000&	28.87&	0.7891&	27.78&	0.7444&	26.96&	0.8109&	31.27&	0.9184\\
        \myrowcolour IPT~\citep{chen2021pre}&	$4\times$&	115.63&	32.64&	-&	29.01&	-&	27.82&	-&	27.26&	-&	-&	-\\ \hline
        
        SwinIR~\citep{liang2021swinir}&	$4\times$&	11.90&	32.92&	0.9044&	29.09&	0.7950&	27.92&	0.7489&	27.45&	0.8254&	32.03&	0.9260\\  
        \myrowcolour CAT-A~\citep{chen2022cross}      & $4\times$ &16.60 & {33.08} & 0.9052 & 29.18 & 0.7960 & 27.99 & 0.7510 & 27.89 & 0.8339 & 32.39 & 0.9285 \\
        ART~\citep{zhang2022accurate}	&$4\times$	& 16.55 &33.04	&0.9051	&29.16	&0.7958	&27.97	&0.7510	&27.77	&0.8321	&32.31	&0.9283	\\		
        \myrowcolour EDT~\citep{li2021efficient}&	$4\times$&	11.63&	 {33.06}&	0.9055&	 {29.23}&	0.7971&	 {27.99}&	0.7510&	27.75&	0.8317&	 {32.39}&	 {0.9283}\\ 
        GRL-B~\citep{li2023efficient} &	$4\times$&	20.20&	 {33.10}&	 {0.9094}&	 {29.37}&	 {0.8058}&	 {28.01}&	 \sotab{0.7611}&	 {28.53}&	 \sotab{0.8504}&	 {32.77}&	 {0.9325}\\ 
        \myrowcolour HAT~\citep{chen2023activating}	&$4\times$	&20.77	&33.18	&0.9073	&29.38	&0.8001	&28.05	&0.7534	&28.37	&0.8447	&32.87	&0.9319	\\
        {\ourmethod}-B (Ours)	&$4\times$& 14.83	&33.14	&\sotab{0.9095}	&29.40	&\sotab{0.8029}	&28.08	&\sotab{0.7611}	&28.44	&0.8448	&32.90	&0.9323			\\
        \myrowcolour HAT-L~\citep{chen2023activating}	&$4\times$	&40.85	&\sotaa{33.30}	&0.9083	&\sotab{29.47}	&0.8015	&\sotab{28.09}	&0.7551	&\sotab{28.60}	&0.8498	&\sotab{33.09}	&\sotab{0.9335}	\\
        {\ourmethod}-L (Ours)	&$4\times$& 39.22	&\sotab{33.22}	&\sotaa{0.9103}	&\sotaa{29.49}	&\sotaa{0.8041}	&\sotaa{28.13}	&\sotaa{0.7622}	&\sotaa{28.72}	&\sotaa{0.8514}	&\sotaa{33.13}	&\sotaa{0.9366}				\\\bottomrule[0.1em]
        \end{tabular}}
\end{center}
\end{table*}

\section{Ablation study}
As mentioned in the main paper, to investigate the effect of $\mathscr{L}_1$ and $\mathscr{L}_2$ information flow, we designed three versions of the {\ourmethod} layers. The architecture of the three transformer layers is shown in Fig.~\ref{fig:attention_version}.

\section{More Quantitative Experimental Results}
\label{sec:supp:more_experimental_results}
Due to the limited space in the main manuscript, we only report a part of the experimental result. In this section, we show the full quantitative experimental results for each IR task in the following.



\subsection{Single-image defocus deblurring}
In addition to the dual-pixel defocus deblurring results, we also shown single-image defocus deblurring results in Tab.~\ref{tab:defocus_deblurring_full}


\noindent\textbf{One model for multiple degradation levels.}
For image denoising and JPEG CAR, we trained a single model to handle multiple degradation levels. This setup makes it possible to apply one model to deal with images that have been degraded under different conditions, making the model more flexible and generalizable. 
During training, the noise level is randomly sampled from the range $[15, 75]$ while the JPEG compression quality factor is randomly sampled from the range $[10, 90]$. The degraded images are generated online. During the test phase, the degradation level is fixed to a certain value. 
The experimental results are summarized in Fig.~6 of our main manuscript. The numerical results for grayscale JPEG CAR are presented in Tab.~12 of our main manuscript. 
These results show that in the one-model-multiple-degradation setting \textcolor{magenta}{\textdagger}, the proposed {\ourmethod} achieves the best performance. 

\subsection{Generalizing one model to more types degradations}

\revise{To validate the generalization capability of the proposed method to different types of degradation, we conducted the following experiments. First, we used the same model for both denoising and JPEG compression artifact removal tasks. Notably, a single model was trained to handle varying levels of degradation. The experimental results for denoising are shown in Tab.~\ref{tab:denoising_one_for_all} while the results for JPEG compression artifact removal are shown in Tab.~\ref{tab:jpeg_car_color} and Tab. 12 of our main manuscript. 
Second, we performed experiments on image restoration under adverse weather conditions, including rain, fog, and snow.
The results are shown in Tab. 14 of our main manuscript. These three sets of experiments collectively highlight that the proposed fractal information flow mechanism enables training a single model that generalizes effectively to various types and levels of degradation.}

\begin{table*}[t]
    \centering
    \caption{\textit{\textbf{Color image JPEG compression artifact removal} results.}}
    \label{tab:jpeg_car_color}
    \setlength{\extrarowheight}{0.7pt}
    \setlength{\tabcolsep}{2.4pt}
    \scalebox{0.66}{
    \begin{tabular}{c | c | c c | c c | c c | c c | c c || c c | c c | c c}
    \toprule[0.1em]
    \multirow{2}{*}{Set} & \multirow{2}{*}{QF} & \multicolumn{2}{c|}{JPEG}  & \multicolumn{2}{c|}{\makecell{\textcolor{magenta}{\textdagger}QGAC}} & \multicolumn{2}{c|}{\makecell{\textcolor{magenta}{\textdagger}FBCNN}} & \multicolumn{2}{c|}{\textcolor{magenta}{\textdagger}DRUNet} & \multicolumn{2}{c||}{\textcolor{magenta}{\textdagger}{\ourmethod} (Ours)} & \multicolumn{2}{c|}{SwinIR} & \multicolumn{2}{c|}{GRL-S} & \multicolumn{2}{c}{{\ourmethod} (Ours)} \\ \cline{3-18}
    & & PSNR & SSIM & PSNR & SSIM & PSNR & SSIM & PSNR & SSIM & PSNR & SSIM & PSNR & SSIM & PSNR & SSIM & PSNR & SSIM  \\
    \midrule[0.1em]
    {\multirow{4}{*}{\rotatebox[origin=c]{90}{\makecell{LIVE1}}}}
    &10	&25.69	&0.7430		&27.62	&0.8040		&27.77	&0.8030		&\sotab{27.47}	&\sotab{0.8045}		&\sotaa{28.24}	&\sotaa{0.8149}		&28.06	&0.8129		&\sotab{28.13}	&\sotab{0.8139}		&\sotaa{28.36}	&\sotaa{0.8180}		\\
    &20	&28.06	&0.8260		&29.88	&0.8680		&30.11	&0.8680		&\sotab{30.29}	&\sotab{0.8743}		&\sotaa{30.59}	&\sotaa{0.8786}		&30.44	&0.8768		&\sotab{30.49}	&\sotab{0.8776}		&\sotaa{30.66}	&\sotaa{0.8797}		\\
    &30	&29.37	&0.8610		&31.17	&0.8960		&31.43	&0.8970		&\sotab{31.64}	&\sotab{0.9020}		&\sotaa{31.95}	&\sotaa{0.9055}		&31.81	&0.9040		&\sotab{31.85}	&\sotab{0.9045}		&\sotaa{32.02}	&\sotaa{0.9063}		\\
    &40	&30.28	&0.8820		&32.05	&0.9120		&32.34	&0.9130		&\sotab{32.56}	&\sotab{0.9174}		&\sotaa{32.88}	&\sotaa{0.9205}		&32.75	&0.9193		&\sotab{32.79}	&\sotab{0.9195}		&\sotaa{32.94}	&\sotaa{0.9210}		\\
    \midrule[0.1em]
    {\multirow{4}{*}{\rotatebox[origin=c]{90}{\makecell{BSD500}}}}
    &10	&25.84	&0.7410		&27.74	&\sotab{0.8020}		&\sotab{27.85}	&0.7990		&27.62	&0.8001		&\sotaa{28.26}	&\sotaa{0.8070}		&28.22	&0.8075		&\sotab{28.26}	&\sotab{0.8083}		&\sotaa{28.35}	&\sotaa{0.8092}		\\
    &20	&28.21	&0.8270		&30.01	&0.8690		&30.14	&0.8670		&\sotab{30.39}	&\sotab{0.8711}		&\sotaa{30.58}	&\sotaa{0.8741}		&30.54	&0.8739		&\sotab{30.57}	&\sotab{0.8746}		&\sotaa{30.61}	&\sotaa{0.8740}		\\
    &30	&29.57	&0.8650		&31.330	&0.8980		&31.45	&0.8970		&\sotab{31.73}	&\sotab{0.9003}		&\sotaa{31.93}	&\sotaa{0.9029}		&31.90	&0.9025		&\sotab{31.92}	&\sotab{0.9030}		&\sotaa{31.99}	&\sotaa{0.9035}		\\
    &40	&30.52	&0.8870		&32.25	&0.9150		&32.36	&0.9130		&\sotab{32.66}	&\sotab{0.9168}		&\sotaa{32.87}	&\sotaa{0.9193}		&32.84	&0.9189		&\sotab{32.86}	&\sotab{0.9192}		&\sotaa{32.92}	&\sotaa{0.9195}		\\
    \hline
    {\multirow{4}{*}{\rotatebox[origin=c]{90}{\makecell{Urban100}}}}
    &10	&24.46	&0.7612		&-	&-		&-	&-		&\sotab{27.10}	&\sotab{0.8400}		&\sotaa{28.78}	&\sotaa{0.8666}		&28.18	&0.8586		&\sotab{28.54}	&\sotab{0.8635}		&\sotaa{29.11}	&\sotaa{0.8727}		\\
    &20	&26.63	&0.8310		&-	&-		&-	&-		&\sotab{30.17}	&\sotab{0.8991}		&\sotaa{31.12}	&\sotaa{0.9087}		&30.53	&0.9030		&\sotab{30.93}	&\sotab{0.9067}		&\sotaa{31.36}	&\sotaa{0.9115}		\\
    &30	&27.96	&0.8640		&-	&-		&-	&-		&\sotab{31.49}	&\sotab{0.9189}		&\sotaa{32.42}	&\sotaa{0.9265}		&31.87	&0.9219		&\sotab{32.24}	&\sotab{0.9247}		&\sotaa{32.57}	&\sotaa{0.9279}		\\
    &40	&28.93	&0.8825		&-	&-		&-	&-		&\sotab{32.36}	&\sotab{0.9301}		&\sotaa{33.26}	&\sotaa{0.9363}		&32.75	&0.9329		&\sotab{33.09}	&\sotab{0.9348}		&\sotaa{33.37}	&\sotaa{0.9373}		\\
    \bottomrule[0.1em]
    \end{tabular}}
\end{table*}

\begin{table}[!ht]
\centering
\caption{\textit{\textbf{Single image motion deblurring} on RealBlur dataset.} \textcolor{violet}{\textdagger}: Methods trained on RealBlur.}
\label{tab:motion_deblurring_realblur}
\setlength{\extrarowheight}{0.7pt}
\setlength{\tabcolsep}{0.7pt}
\scalebox{0.65}{
    \begin{tabular}{l | c | c  |c}
    \toprule[0.1em]
     & {\textbf{RealBlur-R}} & {\textbf{RealBlur-J}} & Average\\
     \textbf{Method} & PSNR$\uparrow$ / SSIM$\uparrow$ & PSNR$\uparrow$ / SSIM$\uparrow$ & PSNR$\uparrow$ / SSIM$\uparrow$ \\
    \midrule[0.1em]

    

    						
    						

    \textcolor{violet}{\textdagger}DeblurGAN-v2
    &36.44 / 0.935		&29.69 / 0.870		&33.07 / 0.903	\\
    \myrowcolour \textcolor{violet}{\textdagger}SRN~\citep{tao2018scale}&38.65 / 0.965		&31.38 / 0.909		&35.02 / 0.937	\\
    \textcolor{violet}{\textdagger}MPRNet~\citep{zamir2021multi}	&39.31 / 0.972		&31.76 / 0.922		&35.54 / 0.947	\\
    \myrowcolour \textcolor{violet}{\textdagger}MIMO-UNet+~\citep{cho2021rethinking_mimo}	&- / -		&32.05 / 0.921		& - / -	\\
    \textcolor{violet}{\textdagger}MAXIM-3S~\citep{tu2022maxim}	&39.45 / 0.962		&\sotab{32.84} / \sotaa{0.935}		&36.15 / 0.949	\\
    \myrowcolour \textcolor{violet}{\textdagger}BANet~\citep{tsai2022banet}	&39.55 / 0.971		&32.00 / 0.923		&35.78 / 0.947	\\
    \textcolor{violet}{\textdagger}MSSNet~\citep{kim2022mssnet}	&39.76 / 0.972		&32.10 / 0.928		&35.93 / 0.950	\\
    DeepRFT+~\citep{mao2023intriguing} & 39.84 / 0.972 & 32.19 / 0.931 & 36.02 / 0.952 \\
    \myrowcolour \textcolor{violet}{\textdagger}Stripformer~\citep{tsai2022stripformer} & 39.84 / \sotab{0.974}  & 32.48 / 0.929 &36.16 / 0.952 \\
    \textcolor{violet}{\textdagger}GRL-B~\citep{li2023efficient} & \sotab{40.20} / \sotab{0.974}		&32.82 / 0.932	 &\sotab{36.51} / \sotab{0.953} \\
    \myrowcolour \textcolor{violet}{\textdagger}{\ourmethod}-L	(Ours) &\sotaa{40.40} / \sotaa{0.976}		&\sotaa{32.92} / \sotab{0.933}		&\sotaa{36.66} / \sotaa{0.954}	\\
    \bottomrule[0.1em]
    \end{tabular}}

\end{table}
\begin{table*}[!t]
    \begin{center}
    \caption{\textit{\textbf{Defocus deblurring}} results. 
    \textbf{S:} single-image defocus deblurring. 
    \textbf{D:} dual-pixel defocus deblurring.}
    \label{tab:defocus_deblurring_full}
    \setlength{\tabcolsep}{2.0pt}
    \setlength{\extrarowheight}{0.7pt}
    \scalebox{0.65}{
    \begin{tabular}{l | c | c | c | c | c | c | c | c | c | c | c | c}
    \toprule[0.1em]
    \multirow{2}{*}{Method} & \multicolumn{4}{c|}{\textbf{Indoor Scenes}} & \multicolumn{4}{c|}{\textbf{Outdoor Scenes}} & \multicolumn{4}{c}{\textbf{Combined}} \\ \cline{2-13}
    	&PSNR$\uparrow$ & SSIM$\uparrow$ & MAE$\downarrow$ & LPIPS$\downarrow$			&PSNR$\uparrow$ & SSIM$\uparrow$ & MAE$\downarrow$ & LPIPS$\downarrow$ &PSNR$\uparrow$ & SSIM$\uparrow$ & MAE$\downarrow$ & LPIPS$\downarrow$		\\ \hline
    EBDB$_S$~\citep{karaali2017edge_EBDB}	&25.77	&0.772	&0.040	&0.297	&21.25	&0.599	&0.058	&0.373	&23.45	&0.683	&0.049	&0.336	\\
    DMENet$_S$~\citep{lee2019deep_dmenet}	&25.50	&0.788	&0.038	&0.298	&21.43	&0.644	&0.063	&0.397	&23.41	&0.714	&0.051	&0.349	\\
    JNB$_S$~\citep{shi2015just_jnb}	&26.73	&0.828	&0.031	&0.273	&21.10	&0.608	&0.064	&0.355	&23.84	&0.715	&0.048	&0.315	\\
    DPDNet$_S$~\citep{abuolaim2020defocus}	&26.54	&0.816	&0.031	&0.239	&22.25	&0.682	&0.056	&0.313	&24.34	&0.747	&0.044	&0.277	\\
    KPAC$_S$~\citep{son2021single_kpac}	&27.97	&0.852	&0.026	&0.182	&22.62	&0.701	&0.053	&0.269	&25.22	&0.774	&0.040	&0.227	\\
    IFAN$_S$~\citep{Lee_2021_CVPRifan}	&28.11	&0.861	&0.026	&0.179	&22.76	&0.720	&0.052	&0.254	&25.37	&0.789	&0.039	&0.217	\\
    Restormer$_S$~\citep{zamir2022restormer}	&\textcolor{red}{28.87}	&\textcolor{blue}{0.882}	&\textcolor{blue}{0.025}	&\textcolor{blue}{0.145}	&\textcolor{blue}{23.24}	&\textcolor{blue}{0.743}	&\textcolor{blue}{0.050}	&\textcolor{blue}{0.209}	&\textcolor{blue}{25.98}	&\textcolor{blue}{0.811}	&\textcolor{blue}{0.038}	&\textcolor{blue}{0.178}	\\
    Fractal-IR$_S$-B (Ours)	&\textcolor{blue}{28.73}	&\textcolor{red}{0.885}	&\textcolor{red}{0.025}	&\textcolor{red}{0.140}	&\textcolor{red}{23.66}	&\textcolor{red}{0.766}	&\textcolor{red}{0.048}	&\textcolor{red}{0.196}	&\textcolor{red}{26.13}	&\textcolor{red}{0.824}	&\textcolor{red}{0.037}	&\textcolor{red}{0.169}		\\											\midrule
    DPDNet$_D$~\citep{abuolaim2020defocus}	&27.48	&0.849	&0.029	&0.189	&22.90	&0.726	&0.052	&0.255	&25.13	&0.786	&0.041	&0.223	\\
    \myrowcolour RDPD$_D$~\citep{abdullah2021rdpd}	&28.10	&0.843	&0.027	&0.210	&22.82	&0.704	&0.053	&0.298	&25.39	&0.772	&0.040	&0.255	\\
    Uformer$_D$~\citep{wang2022uformer}	&28.23	&0.860	&0.026	&0.199	&23.10	&0.728	&0.051	&0.285	&25.65	&0.795	&0.039	&0.243	\\
    \myrowcolour IFAN$_D$~\citep{Lee_2021_CVPRifan}	&28.66	&0.868	&\textcolor{blue}{0.025}	&0.172	&23.46	&0.743	&0.049	&0.240	&25.99	&0.804	&0.037	&0.207	\\
    Restormer$_D$~\citep{zamir2022restormer}	&\textcolor{blue}{29.48}	&\textcolor{blue}{0.895}	&\textcolor{red}{0.023}	&\textcolor{blue}{0.134}	&\textcolor{blue}{23.97}	&\textcolor{blue}{0.773}	&\textcolor{blue}{0.047}	&\textcolor{blue}{0.175}	&\textcolor{blue}{26.66}	&\textcolor{blue}{0.833}	&\textcolor{blue}{0.035}	&\textcolor{blue}{0.155}	\\
    \myrowcolour Fractal-IR$_D$-B (Ours)	&\textcolor{red}{29.70}	&\textcolor{red}{0.902}	&\textcolor{red}{0.023}	&\textcolor{red}{0.116}	&\textcolor{red}{24.46}	&\textcolor{red}{0.798}	&\textcolor{red}{0.045}	&\textcolor{red}{0.154}	&\textcolor{red}{27.01}	&\textcolor{red}{0.848}	&\textcolor{red}{0.034}	&\textcolor{red}{0.135}	\\ \bottomrule[0.1em]
    \end{tabular}}
    \end{center}
    \vspace{-2mm}
\end{table*}

\begin{table*}[t]
\centering
\caption{\textit{\textbf{Color and grayscale image denoising}} results. A single model is trained to handle multiple noise levels.
}
\label{tab:denoising_one_for_all}
\setlength{\extrarowheight}{0.7pt}
\setlength{\tabcolsep}{1.2pt}
\scalebox{0.625}{
    \begin{tabular}{l | c | c c c | c c c | c c c | c c c || c c c | c c c }
    \toprule[0.1em]
    \multirow{3}{*}{\textbf{Method}} & \multicolumn{1}{c|}{\multirow{3}{*}{ \makecell{\textbf{Params} \\ \textbf{[M]}} }}
     & \multicolumn{12}{c||}{\textbf{Color}} & \multicolumn{6}{c}{\textbf{Grayscale}} \\ \cline{3-20}
    & & \multicolumn{3}{c|}{\textbf{CBSD68}} & \multicolumn{3}{c|}{\textbf{Kodak24}} & \multicolumn{3}{c|}{\textbf{McMaster}} & \multicolumn{3}{c||}{\textbf{Urban100}}  & \multicolumn{3}{c|}{\textbf{Set12}}  & \multicolumn{3}{c}{\textbf{Urban100}} \\
            &  & $\sigma$$=$$15$ & $\sigma$$=$$25$ & $\sigma$$=$$50$ & $\sigma$$=$$15$ & $\sigma$$=$$25$ & $\sigma$$=$$50$ & $\sigma$$=$$15$ & $\sigma$$=$$25$ & $\sigma$$=$$50$ & $\sigma$$=$$15$ & $\sigma$$=$$25$ & $\sigma$$=$$50$ & $\sigma$$=$$15$ & $\sigma$$=$$25$ & $\sigma$$=$$50$ & $\sigma$$=$$15$ & $\sigma$$=$$25$ & $\sigma$$=$$50$ \\ \midrule
    \myrowcolour DnCNN~\citep{kiku2016beyond}	& 0.56	&33.90	&31.24	&27.95	&34.60	&32.14	&28.95	&33.45	&31.52	&28.62	&32.98	&30.81	&27.59	&32.67	&30.35	&27.18				&32.28	&29.80	&26.35	\\
    FFDNet~\citep{zhang2018ffdnet}	& 0.49	&33.87	&31.21	&27.96	&34.63	&32.13	&28.98	&34.66	&32.35	&29.18	&33.83	&31.40	&28.05	&32.75	&30.43	&27.32				&32.40	&29.90	&26.50	\\
    \myrowcolour IRCNN~\citep{zhang2017learning}	& 0.19	&33.86	&31.16	&27.86	&34.69	&32.18	&28.93	&34.58	&32.18	&28.91	&33.78	&31.20	&27.70	&32.76	&30.37	&27.12				&32.46	&29.80	&26.22	\\
    DRUNet~\citep{zhang2021plug}	& 32.64	&34.30	&31.69	&28.51	&35.31	&32.89	&29.86	&35.40	&33.14	&30.08	&34.81	&32.60	&29.61	&33.25	&30.94	&27.90				&33.44	&31.11	&27.96	\\
    \myrowcolour Restormer~\citep{zamir2022restormer}	& 26.13	&\sotab{34.39}	&\sotab{31.78}	&\sotab{28.59}	&\sotab{35.44}	&\sotaa{33.02}	&\sotaa{30.00}	&\sotab{35.55}	&\sotab{33.31}	&\sotab{30.29}	&\sotab{35.06}	&\sotab{32.91}	&\sotab{30.02}	&\sotab{33.35}	&\sotab{31.04}	&\sotab{28.01}		&\sotab{33.67}	&\sotab{31.39}	&\sotab{28.33}	\\
    {\ourmethod} (Ours)	& 22.33	&\sotaa{34.43}	&\sotaa{31.80}	&\sotaa{28.60}	&\sotaa{35.42}	&\sotab{33.00}	&\sotab{29.95}	&\sotaa{35.67}	&\sotaa{33.43}	&\sotaa{30.38}	&\sotaa{35.46}	&\sotaa{33.32}	&\sotaa{30.47}	&\sotaa{33.49}	&\sotaa{31.18}	&\sotaa{28.14}		&\sotaa{34.09}	&\sotaa{31.87}	&\sotaa{28.86}	\\
    \bottomrule[0.1em]
    \end{tabular}
}
\end{table*}

\section{\revise{Comparison with ShuffleFormer and Shuffle Transformer}}
\revise{We compare with Random shuffle transformer (ShuffleFormer)~\citep{xiao2023random} and Shuffle transformer~\citep{huang2021shuffle}. Both methods use spatial shuffle operations to facilitate non-local information exchange, with one being random and the other deterministic.}

\revise{Random Shuffle Transformer (ShuffleFormer)~\citep{xiao2023random} applies random shuffling on the spatial dimension, which increases the probability of global information existing within a local window. While this operation extends the receptive field globally in a single step, it compromises the relevance of pixels within the window. In contrast, the fractial information flow proposed in this paper progressively propagates information from local to global while preserving the relevance of attended pixels. A comparison with ShuffleFormer on image deblurring is presented in Tab. 9 of our main manuscript. {\ourmethod} outperforms ShuffleFormer by a significant margin while using 55.5\% fewer parameters. This demonstrates the effectiveness of the fractal information flow method introduced in this work.}

\revise{Shuffle Transformer~\citep{huang2021shuffle} employs a spatial shuffle operation to aggregate information from distant pixels or tokens. However, it differs from the proposed {\ourmethod} in several key aspects. First, Shuffle Transformer does not enable progressive information propagation within a fractal tree structure. Second, its shuffle operation is based on a fixed grid size of $g = 8$. The distance between pixels in the shuffled window is $H/g$ and $W/g$ along the two axes, which directly depends on the image size. For large images (e.g., 1024 pixels), this design forces distant pixels to attend to one another, often introducing irrelevant information. Consequently, this operation is unsuitable for image restoration tasks, where image sizes can become extremely large. In contrast, the L2 information flow attention proposed in this paper limits the maximum patch size, thereby constraining the maximum distance between pixels at this stage. This restriction enhances the relevance of pixel interactions, making it more effective for image restoration tasks.}

\section{More Visual Results}
\label{sec:supp:more_visual_results}
\begin{figure*}[!t]
    \centering
    \includegraphics[width=0.9\linewidth]{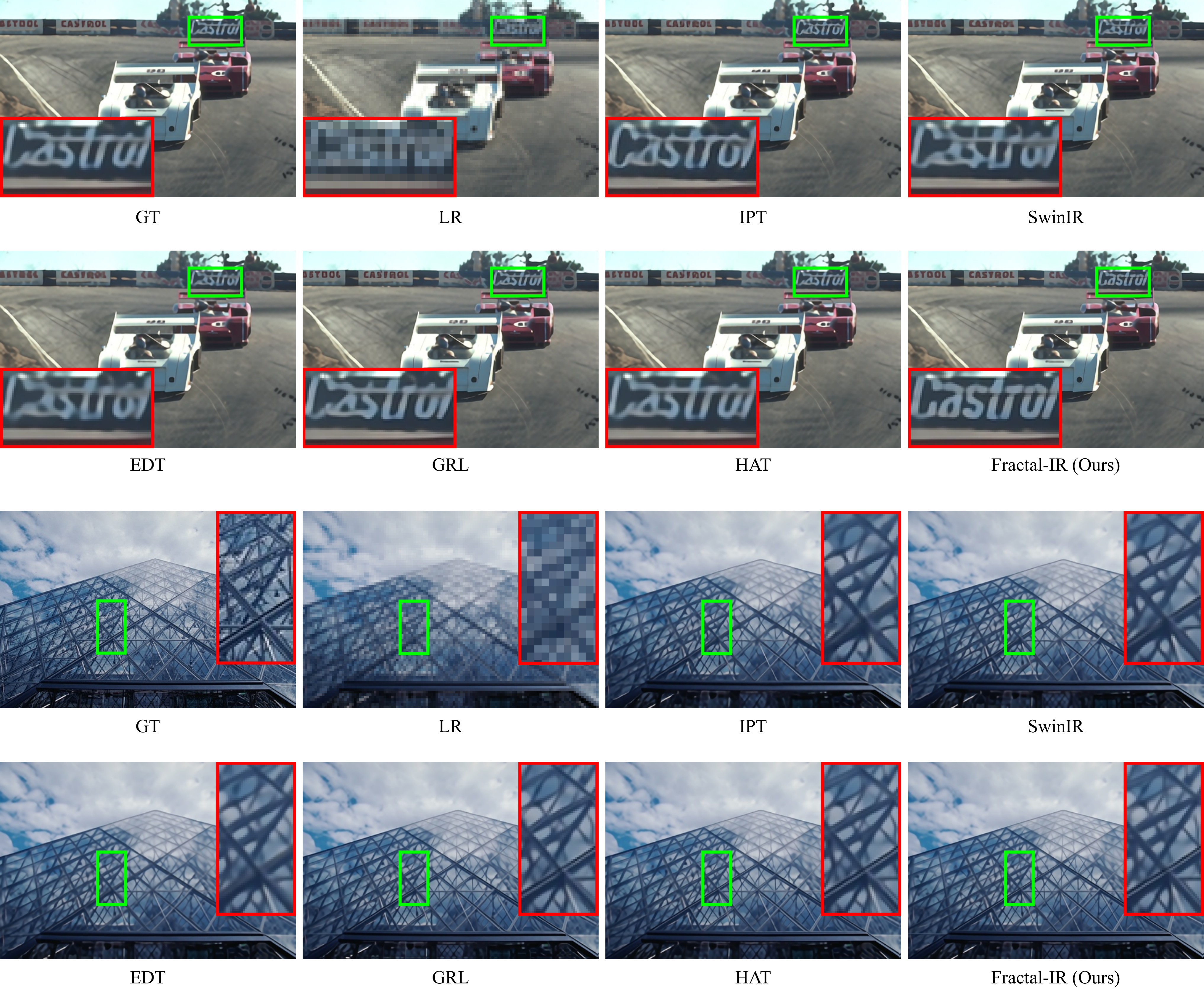}
    \vspace{-3mm}
    \caption{Visual results for classical image $\times 4$ SR on B100 dataset.}
    \label{fig:supp_visual_sr_b100_part1}
\end{figure*}

To further support the effectiveness and generalizability of the proposed {\ourmethod} intuitively. We provide more visual comparison in terms of image SR (Fig.~\ref{fig:supp_visual_sr_b100_part1}, 
),
JPEG compression artifact removal (Fig.~\ref{fig:supp_visual_jpeg_color_bsd500}
), image restoration in adverse weather conditions(Fig.~\ref{fig:supp_weather_fig})
blow. As shown in those figures, the visual results of the proposed {\ourmethod} are improved compared with the other methods.

\section{Limitations}
\label{sec:supp:limitation}
Despite the state-of-the-art performance of {\ourmethod}, our explorations towards scaling up the model for IR in this paper are still incomplete. Scaling up the IR model is intricate, involving considerations like model design, data collection, and computing resources. We hope our work can catalyze positive impacts on future research, encouraging more comprehensive scaling-up explorations and propelling IR into the domain of large-scale models.

\begin{figure*}[!t]
    \centering
    \includegraphics[width=1.0\linewidth]{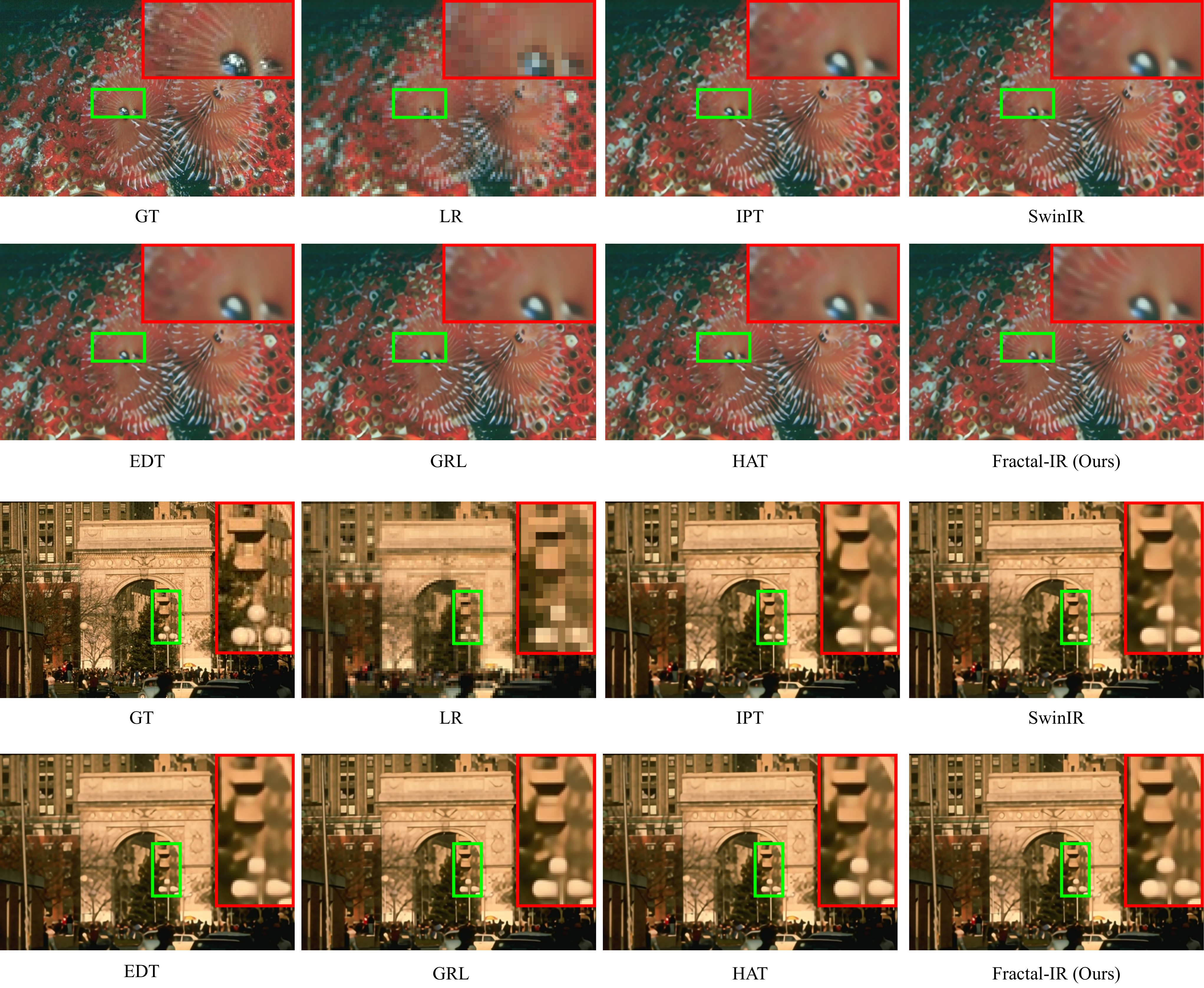}
    \vspace{-6mm}
    \caption{Visual results for classical image SR on B100 dataset. The upscaling factor is $\times 4$.}
    \label{fig:supp_visual_sr_b100_part2}
\end{figure*}





\begin{figure*}[!t]
    \centering
    \includegraphics[width=1.0\linewidth]{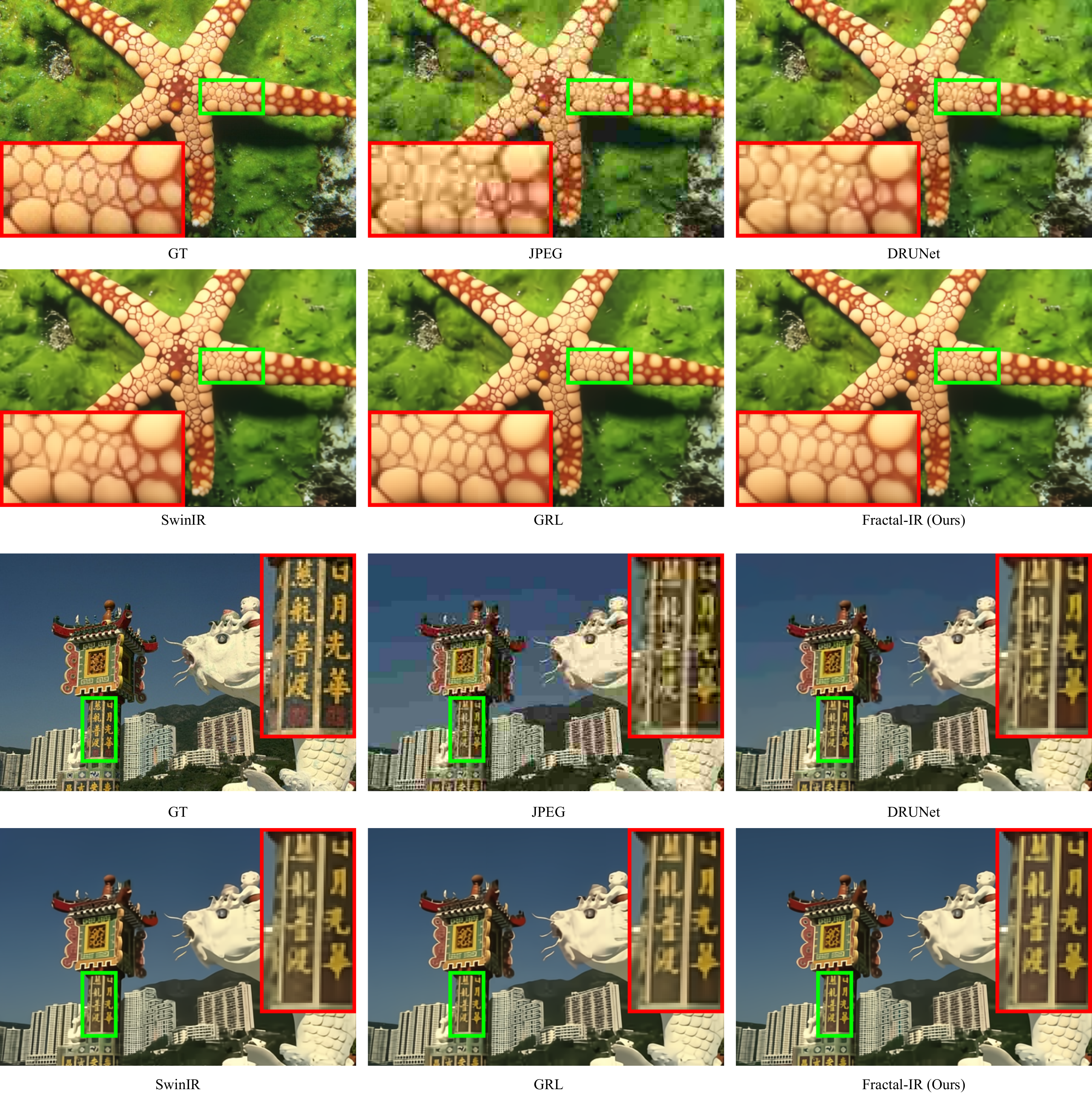}
    \vspace{-6mm}
    \caption{Visual results for color image JPEG compression artifact removal on BSD500 dataset. The quality factor of JPEG image compression is $10$.}
    \label{fig:supp_visual_jpeg_color_bsd500}
\end{figure*}

\begin{figure*}[!t]
    \centering
    \includegraphics[width=1.0\linewidth]{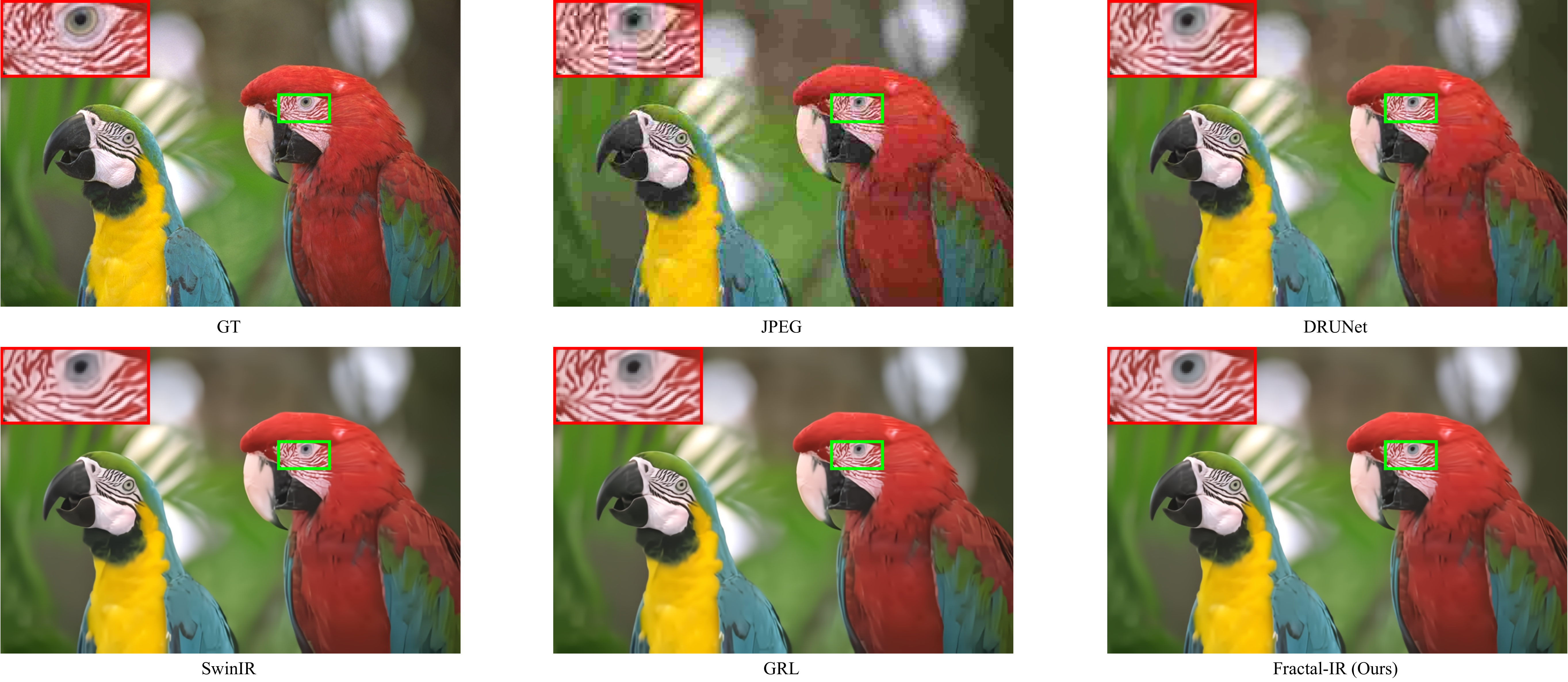}
    \vspace{-6mm}
    \caption{Visual results for color image JPEG compression artifact removal on LIVE1 dataset. The quality factor of JPEG image compression is $10$.}
    \label{fig:supp_visual_jpeg_color_live1}
\end{figure*}

\begin{figure*}[!t]
    \centering
    \includegraphics[width=1.0\linewidth]{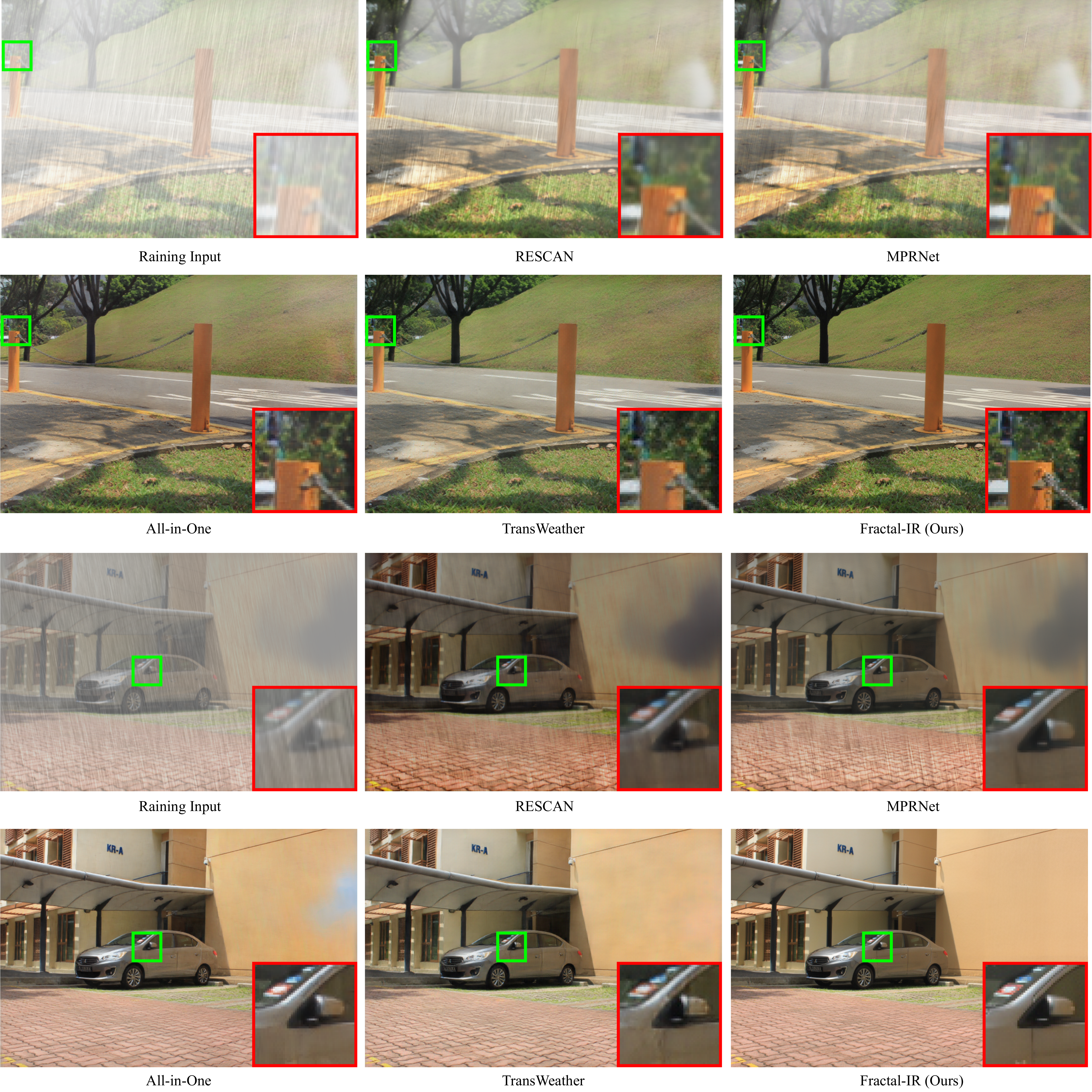}
    \vspace{-6mm}
    \caption{Visual results for restoring images in adverse weather conditions.}
    \label{fig:supp_weather_fig}
\end{figure*}



\section{Impact Statement}
\label{sec:supp:impact}
The proposed {\ourmethod} framework significantly advances image restoration by addressing critical challenges in model scalability, efficiency, and generalization across diverse degradation types and resolutions. By leveraging a fractal-based design, {\ourmethod} effectively balances local and global information processing, reducing the computational overhead typically associated with self-attention mechanisms in transformers. Furthermore, our holistic approach to model scaling ensures that {\ourmethod} can be trained effectively at larger capacities, enabling it to capture richer image patterns and deliver state-of-the-art performance across a broad range of restoration tasks. This work opens up new possibilities for deploying scalable, efficient, and generalizable models in real-world image restoration.

\end{document}